\documentclass{article}
\usepackage{sfa_preprint,times}

\usepackage{amsmath,amsfonts,bm}

\def\eqref#1{equation~\ref{#1}}

\def\1{\bm{1}}

\def\rj{{\textnormal{j}}}

\def\rn{{\textnormal{n}}}

\def\ru{{\textnormal{u}}}

\def\rx{{\textnormal{x}}}
\def\ry{{\textnormal{y}}}
\def\rz{{\textnormal{z}}}

\def\rvepsilon{{\mathbf{\epsilon}}}

\def\rvf{{\mathbf{f}}}

\def\rvn{{\mathbf{n}}}

\def\vzero{{\bm{0}}}

\def\va{{\bm{a}}}
\def\vb{{\bm{b}}}
\def\vc{{\bm{c}}}
\def\vd{{\bm{d}}}

\def\vf{{\bm{f}}}
\def\vg{{\bm{g}}}
\def\vh{{\bm{h}}}

\def\vk{{\bm{k}}}

\def\vp{{\bm{p}}}
\def\vq{{\bm{q}}}

\def\vs{{\bm{s}}}

\def\vu{{\bm{u}}}
\def\vv{{\bm{v}}}
\def\vw{{\bm{w}}}
\def\vx{{\bm{x}}}
\def\vy{{\bm{y}}}
\def\vz{{\bm{z}}}

\def\mA{{\bm{A}}}

\def\mC{{\bm{C}}}

\def\mK{{\bm{K}}}

\def\mV{{\bm{V}}}

\def\mSigma{{\bm{\Sigma}}}

\DeclareMathAlphabet{\mathsfit}{\encodingdefault}{\sfdefault}{m}{sl}
\SetMathAlphabet{\mathsfit}{bold}{\encodingdefault}{\sfdefault}{bx}{n}
\newcommand{\tens}[1]{\bm{\mathsfit{#1}}}

\def\tK{{\tens{K}}}

\def\tQ{{\tens{Q}}}

\def\tV{{\tens{V}}}

\def\tY{{\tens{Y}}}

\newcommand{\E}{\mathbb{E}}

\newcommand{\R}{\mathbb{R}}

\newcommand{\softmax}{\mathrm{softmax}}

\newcommand{\KL}{D_{\mathrm{KL}}}
\newcommand{\Var}{\mathrm{Var}}

\newcommand{\Cov}{\mathrm{Cov}}

\DeclareMathOperator*{\argmin}{arg\,min}

\newcommand{\mPi}{\bm{\Pi}}
\newcommand{\vpi}{\bm{\pi}}
\newcommand{\vomega}{\bm{\omega}}
\renewcommand{\rvepsilon}{{\bm{\epsilon}}}
\newcommand{\diff}{\mathrm{d}}

\usepackage{float}
\usepackage{needspace}
\usepackage{hyperref}
\usepackage{url}
\usepackage{xurl}
\usepackage[section]{placeins}
\usepackage{graphicx}
\usepackage{booktabs}
\usepackage{multirow}
\usepackage{tabularx}
\usepackage{subcaption}
\usepackage{amsthm}
\usepackage{tikz}
\usetikzlibrary{arrows.meta,positioning,calc}

\hypersetup{
    hidelinks,
    pdftitle={SFA: A Statistical Physics-Inspired Diffusion Transformer for Efficient and High-Quality Image Generation},
    pdfauthor={Zixuan Huang, Tianhao You, and Guozhang Chen},
    pdfsubject={Image generation with Scale-Free Attention},
    pdfkeywords={Scale-Free Attention, diffusion transformers, image generation, efficient attention}
}

\renewcommand{\eqref}[1]{(\ref{#1})}

\newtheorem{theorem}{Theorem}[section]
\newtheorem{proposition}[theorem]{Proposition}
\newtheorem{lemma}[theorem]{Lemma}

\theoremstyle{remark}

\title{SFA: A Statistical Physics-Inspired
    Diffusion Transformer for Efficient
    and High-Quality Image Generation}

\author{%
  Zixuan Huang$^{1,2}$ \qquad Tianhao You$^{1}$ \qquad Guozhang Chen$^{1,3,4}\thanks{Corresponding author: \texttt{guozhang.chen@pku.edu.cn}}$ \\
  $^{1}$School of Computer Science \\
  $^{2}$School of Physics \\
  $^{3}$State Key Laboratory of Multimedia Information Processing\\
  $^{4}$National Engineering Research Center of Visual Technology, \\
  Peking University, Beijing 100871, China
}

\begin{document}

\maketitle

\begin{abstract}
    Diffusion transformers have achieved remarkable success in image generation, but accelerating attention while preserving generation quality remains a major challenge at high resolutions. Inspired by statistical physics and the widespread occurrence of scale-free laws in nature, we introduce Scale-Free Attention (SFA), which preserves fine local interactions while summarizing distant regions at progressively coarser scales. For $N$ tokens, mean pooling, token-count correction, and a shared softmax over a scale-free spatial hierarchy give SFA $O(N\log N)$ complexity in both forward and backward passes. Our Triton implementation achieves $357.3\times$ forward and $265.6\times$ backward attention speedups over FlashAttention at 262,144 tokens. After 40 epochs of PixelFlow training from scratch on ImageNet at $256\times256$ resolution, SFA achieves the best generation quality among five attention mechanisms, including full attention. Fr\'echet inception distance evaluated on 50,000 generated samples (FID-50k) improves from 13.25 for full attention to 12.51, and the advantage over full attention persists through 80 epochs. SFA also achieves the best generation quality among the compared efficient-attention methods on Flickr-Faces-HQ (FFHQ). Beyond training from scratch, Stable Diffusion 3.5 Medium with attention replaced by SFA retains the generation quality of its full-attention counterpart under matched fine-tuning budgets. We obtain similar results when fine-tuning pretrained PixelFlow on ImageNet. SFA establishes scale-free structure as an effective design principle for efficient, high-quality diffusion transformers.

\end{abstract}

\section{Introduction}
\label{sec:introduction}

Generating coherent visual content at high resolution requires models that capture both fine detail and long-range structure. Diffusion transformers (DiTs) have achieved strong image-generation performance by combining scalable transformer backbones with iterative denoising \citep{peebles2023dit,sd35article}. However, the quadratic cost of all-to-all attention becomes a major constraint as spatial resolution grows, particularly for direct pixel-space generation with PixelFlow~\citep{pixelflow}. Substantially reducing this cost while matching or exceeding the generation quality of full attention remains an open problem.

Existing approaches address different parts of this challenge. FlashAttention improves exact attention through memory-efficient tiling and parallel execution \citep{dao2022flashattention,dao2023flashattention2}. Its arithmetic cost remains quadratic in token count. Linear attention uses feature-map aggregation to reduce this cost to linear complexity \citep{katharopoulos2020linear,choromanski2021performer}. However, the finite feature representations underlying this efficiency trade computational cost against approximation fidelity, making preservation of pretrained softmax interactions a separate challenge. Sparse methods such as Log-linear Sparse Attention (LLSA) \citep{zhou2026llsa} reduce interactions through hierarchical selection. Selection adds runtime overhead and omits fine interactions outside the selected regions, which can limit generation quality. Hierarchical attention offers structured global approximations, with H-Transformer-1D \citep{zhu2021htransformer} developed for one-dimensional sequences. Fast Multipole Attention (FMA) \citep{kang2025fma} extends multilevel computation to image classification tasks using learned pooling and separate softmax at each scale, which complicates efficient kernel implementation and precludes direct resolution extrapolation and drop-in fine-tuning from full-attention checkpoints. These limitations restrict its applicability to long sequences and large pretrained generative models. The remaining challenge is to combine subquadratic cost, efficient execution, and generation quality competitive with full attention in both training from scratch and adaptation.

Scale-free laws, widely studied in statistical physics, appear in natural images as approximate scale invariance and power spectra close to an inverse-square law \citep{field1987relations,ruderman1994statistics}. Consistent with this spatial prior, we observe an approximate power law in the aggregate radial attention of pretrained PixelFlow, whereas random initialization gives a nearly flat profile (Figure~\ref{fig:main-overview}(C)). Motivated by this observation, we introduce \textbf{Scale-Free Attention (SFA)}, which preserves exact local interactions and partitions the remaining image into progressively larger blocks, each represented by its mean key and value. SFA adds the logarithm of each block's token count to its logit and applies a single softmax to all local tokens and block summaries. Our contributions are threefold.

\paragraph{Scale-free attention.}
By combining mean pooling, a $\log|B|$ correction for spatial blocks $B$ containing $|B|$ tokens, and a shared softmax over a two-dimensional scale-free hierarchy, SFA retains the original attention projections without additional parameters or hyperparameters. For $N$ tokens, both forward and backward passes have $O(N\log N)$ complexity. In our experiments, SFA achieves higher generation quality than the three efficient-attention baselines and matches or exceeds the performance of full attention (Figure~\ref{fig:overview-results}(C,D)).

\paragraph{Fast Triton kernels.}
We implement SFA in Triton~\citep{tillet2019triton}, using recursive pooling and kernels that fuse attention across spatial scales. Figures~\ref{fig:kernel-forward} and~\ref{fig:kernel-backward} in Appendix~\ref{app:implementation} illustrate the forward and backward kernel architectures, respectively. At 262,144 tokens, SFA achieves $357.3\times$ forward and $265.6\times$ backward speedups over FlashAttention on an RTX 5090 (Figure~\ref{fig:overview-results}(A,B)).

\paragraph{Approximation guarantees.}
For a fixed spatial partition, we prove that SFA uniquely minimizes reverse Kullback--Leibler (KL) divergence to full attention among blockwise uniform distributions, and bound attention-mass and output errors under logarithmic spatial logits.

\section{Related Work}
\label{sec:related_work}

\paragraph{Generative modeling.}
Denoising diffusion models learn to reverse gradual corruption \citep{ho2020ddpm}, while score-based models describe generation through reverse-time stochastic dynamics \citep{song2021score}. Flow matching learns vector fields along prescribed probability paths \citep{lipman2023flowmatching}, and rectified flow uses straight interpolations between noise and data \citep{liu2023rectifiedflow}. These objectives can operate in different representations: latent diffusion reduces spatial computation through learned compression \citep{rombach2022ldm}, whereas PixelFlow generates directly in pixel space through a resolution cascade \citep{pixelflow}. Diffusion transformers apply scalable transformer backbones to latent patches \citep{vaswani2017attention,peebles2023dit}, improving generation as model capacity and token count increase, and multimodal rectified flow transformers extend attention to joint image--text interactions \citep{sd35article}. Across these settings, repeated evaluation during sampling makes the cost of spatial interactions central to generation efficiency.

\begingroup
\makeatletter
\setlength{\@fptop}{0pt}
\makeatother

\begin{figure}[!t]
    \centering
    \includegraphics[width=\linewidth]{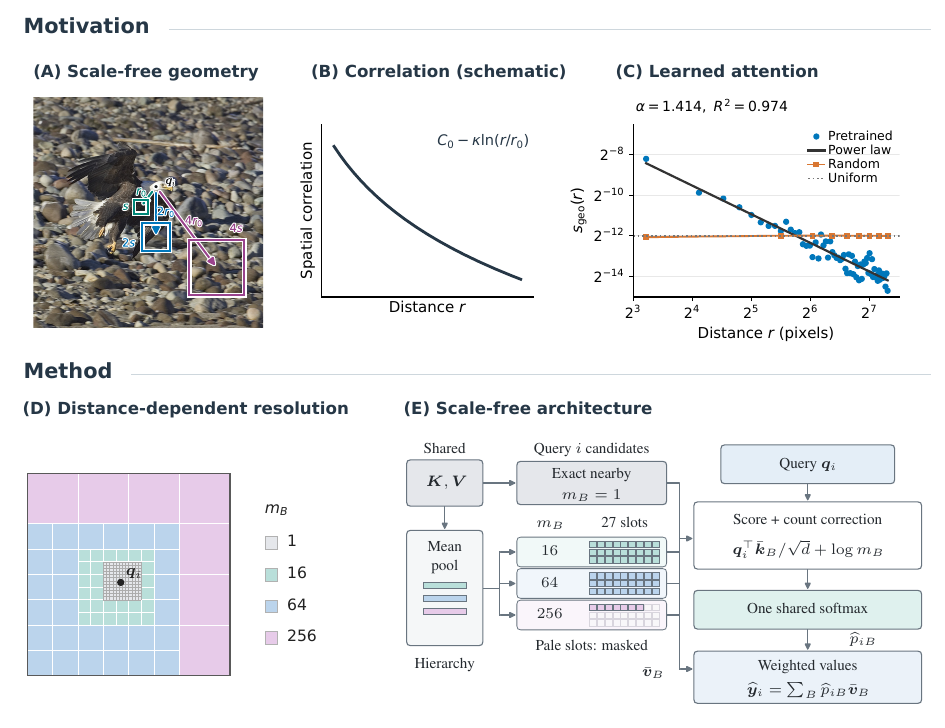}
    \caption{Motivation and construction of SFA (\textbf{ours}). \textbf{(A)} Image regions at center distances $r_0,2r_0,4r_0$ from query $\vq_i$ have side lengths $s,2s,4s$. The smallest block lies on the bird, the intermediate block crosses its outline, and the largest contains only background. \textbf{(B)} Schematic logarithmic spatial correlation $C_0-\kappa\ln(r/r_0)$, where $r$ is separation, $r_0>0$ is the reference distance, $C_0$ is an offset, and $\kappa>0$ controls decay. \textbf{(C)} The geometric mean $s_{\mathrm{geo}}(r)$ of layer--head radial profiles follows an approximate power law in pretrained PixelFlow-XL, with decay exponent $\alpha$ and coefficient of determination $R^2$; random initialization lies near the uniform reference. The horizontal axis shows distance $r$ in pixels on a base-2 logarithmic scale (Appendix~\ref{app:powerlaw-details}). \textbf{(D)} The partition covers a $64\times64$ token grid with exact nearby interactions and progressively coarser distant blocks, preserving each original query. \textbf{(E)} Key/value matrices $\mK,\mV$ supply exact local tokens and a shared mean-pooling hierarchy. Each query uses the same 27-slot layout per pooled scale, with inactive slots masked; active counts are 27, 27, and 7 for (D). Scoring each representative with query $\vq_i$, adding $\log m_B$, and applying one softmax across all scales yields probabilities $\widehat p_{iB}$ for weighted-value aggregation. Here $m_B=|B|$ counts original tokens in block $B$, and $d$ is the head dimension.}
    \label{fig:main-overview}
\end{figure}

\begin{figure}[!t]
    \centering
    \includegraphics[width=\linewidth]{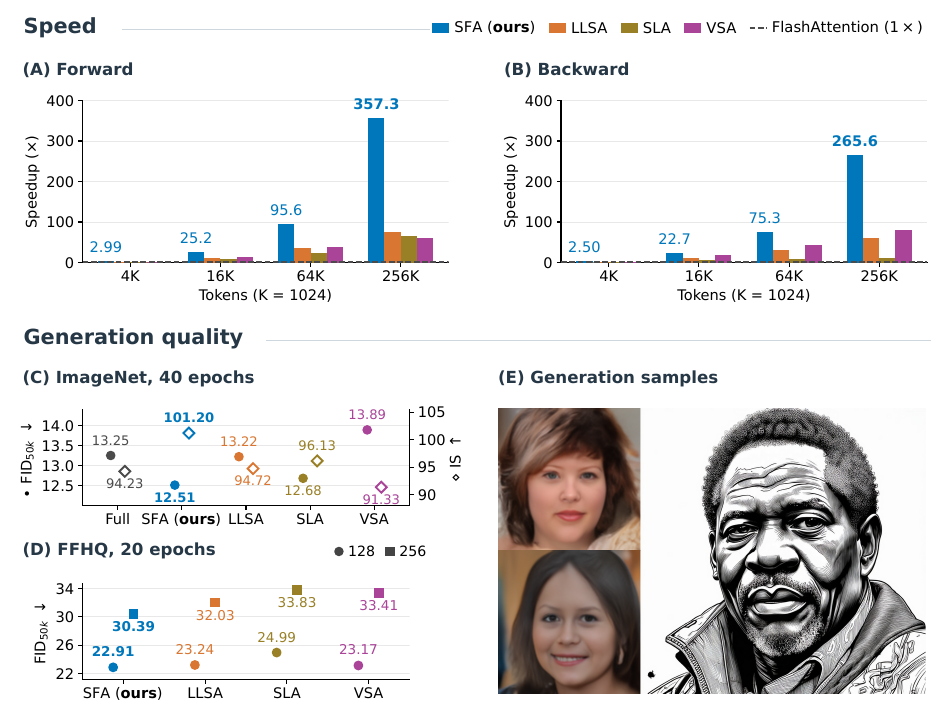}
    \caption{Attention efficiency and generation quality of SFA (\textbf{ours}). \textbf{(A,B)} Forward and backward attention speedups over FlashAttention for 4,096--262,144 tokens on an RTX 5090. \textbf{(C)} PixelFlow trained from scratch on ImageNet~\citep{deng2009imagenet} at $256\times256$ resolution for 40 epochs. Circles show Fr\'echet inception distance (FID; \citealp{heusel2017fid}) on the left axis (lower is better), and diamonds show inception score (IS; \citealp{salimans2016improved}) on the right axis (higher is better), both evaluated on 50,000 generated samples. \textbf{(D)} FFHQ~\citep{karras2019stylegan} FID-50k after 20 epochs at each resolution: circles and squares denote $128\times128$ and $256\times256$, respectively. Protocols and full results are in Appendices~\ref{app:kernel-performance}--\ref{app:pixelflow-full-results}. \textbf{(E)} Generated face samples at $128\times128$ (left) and a text-to-image sample from Stable Diffusion 3.5 Medium~\citep{sd35medium} at $1024\times1024$ (right). The text prompt is reproduced in Appendix~\ref{app:qualitative}.}
    \label{fig:overview-results}
\end{figure}

\paragraph{Efficient diffusion attention.}
Trainable efficient attention reduces interaction cost while allowing the generator to adapt to the resulting operator. For pretrained generators, compatibility with existing projections and output resolution further supports adaptation through fine-tuning. Recent approaches differ in how they allocate computation: Combiner \citep{combiner} extends sparse attention patterns with indirect attention to regional summaries, providing a global receptive field. FMMformer \citep{FMMformer} computes near- and far-field interactions separately, using an approach similar to linear attention for the latter. Hourglass Diffusion Transformer (HDiT) \citep{hdit} uses low-resolution global self-attention for coherence. Piecewise Sparse Attention (PISA) \citep{PISA} uses dynamic block selection and Taylor approximation of weakly correlated blocks. Video Sparse Attention (VSA) \citep{vsa} selects important blocks using pooled features, and Sparse-Linear Attention (SLA) \citep{sla} combines sparse interactions with a linear component. Log-linear Sparse Attention (LLSA) \citep{zhou2026llsa} performs hierarchical selection, enriching fine tokens with pooled keys and values from multiple scales. SFA likewise uses recursive means and shared attention computation. Its fixed spatial interaction lists maintain global coverage through a content-independent allocation of resolution. Appendix~\ref{app:linear-hardware-attention} further compares SFA with Sana~\citep{xie2025sana} and Native Sparse Attention (NSA)~\citep{yuan2025nsa}.

\paragraph{Multiscale approximation.}
This fixed spatial construction connects SFA to multiscale attention approximations. FMA \citep{kang2025fma} uses learned pooling kernels and separate softmax at each scale. H-Transformer-1D \citep{zhu2021htransformer} operates on one-dimensional sequences and also coarse-grains queries. Multi Resolution Analysis (MRA) \citep{zeng2022mra} coarsens queries and uses data-driven criteria to determine where coarsening is appropriate. Focal Attention \citep{yang2021focal} uses multiscale overlapping partitions and introduces learnable pooling, whereas Radial Attention \citep{radial} reduces interactions with spatiotemporally distant video patches. Token Merging \citep{bolya2023tome} combines similar tokens in vision transformers, reducing the sequence length passed to subsequent layers and adding a log token-count bias to the attention logits. Further comparisons appear in Appendix~\ref{app:extended-related-work}.

\endgroup

\begingroup
    \let\FloatBarrier\relax
    \section{Scale-Free Attention}
\label{sec:method}

Deterministic scalars are italic, random scalars upright, vectors bold, and matrices bold capitals; plain indexed letters denote coordinates. Logarithms are natural unless specified. Consider one head on an $L\times L$ grid, with power-of-two side length $L\geq8$ and $N=L^2$ tokens. Indices $i,j\in\{1,\ldots,N\}$ identify positions $\vx_i,\vx_j\in\R^2$, queries and keys $\vq_i,\vk_j\in\R^d$ after positional encoding, and values $\vv_j\in\R^{d_v}$. Thus, $d$ and $d_v$ are the query/key and value dimensions.

Write $z_{ij}=\vq_i^\top\vk_j/\sqrt d$ for the logits and $a_{ij}=e^{z_{ij}}/\sum_{t=1}^N e^{z_{it}}$ for the full-attention weights, collected in $\va_i=(a_{i1},\ldots,a_{iN})^\top$. Here $\top$ denotes transpose and $\|\cdot\|_2$ is the Euclidean norm.

We call our method Scale-Free Attention (SFA) because it uses the same budget of representative tokens at each doubling of distance and normalizes their count-corrected logits in a single softmax across scales. We retain nearby tokens individually and group distant tokens into blocks whose side lengths grow approximately in proportion to distance. These blocks form a complete, disjoint partition $\mathcal P_i$. With a constant number of blocks at each of $O(\log L)=O(\log N)$ scales, the partition contains $O(\log N)$ blocks per query. Section~\ref{sec:partition} specifies this construction. For a block $B$, let $m_B=|B|$ and let $\E_B$ denote uniform averaging over its tokens. We can then express the full-attention output and its block approximation as
\begin{equation}
    \begin{aligned}
        \vy_i
        &=\frac{\sum_{j=1}^N e^{z_{ij}}\vv_j}{\sum_{j=1}^N e^{z_{ij}}}
        =\frac{\sum_{B\in\mathcal P_i}\sum_{j\in B}e^{z_{ij}}\vv_j}
        {\sum_{B\in\mathcal P_i}\sum_{j\in B}e^{z_{ij}}}\\[4pt]
        &=\frac{\sum_{B\in\mathcal P_i}m_B\E_B[e^{z_{ij}}\vv_j]}
        {\sum_{B\in\mathcal P_i}m_B\E_B[e^{z_{ij}}]}\\[4pt]
        &\approx\frac{\sum_{B\in\mathcal P_i}m_Be^{\E_B z_{ij}}\E_B\vv_j}
        {\sum_{B\in\mathcal P_i}m_Be^{\E_B z_{ij}}}\\[4pt]
        &=\frac{\sum_{B\in\mathcal P_i}e^{\bar z_{iB}+\log m_B}\bar{\vv}_B}
        {\sum_{B\in\mathcal P_i}e^{\bar z_{iB}+\log m_B}}
        =:\widehat{\vy}_i.
    \end{aligned}
    \label{eq:sfa}
\end{equation}
Here $\bar{\vk}_B=\E_B\vk_j$, $\bar{\vv}_B=\E_B\vv_j$, and $\bar z_{iB}=\E_Bz_{ij}$ are the block means. For the first three equalities, we regroup the sums and express the inner sum as an expectation, which is therefore exact. Next, the approximation replaces each block's logits by their mean, yielding a single softmax over block representatives. The factor $m_B$ becomes the additive correction $\log m_B$ in the block logit, preserving the population represented by each pooled block. This approximation leverages the logarithmic decay property of correlations between distant tokens, which we analyze in greater detail in Section~\ref{sec:theory} and Appendix~\ref{app:theory}, including reverse-KL optimality and approximation-error bounds. And we denote the normalized coefficient of $\bar{\vv}_B$ in the final expression by $\widehat p_{iB}$, the block probability mass. This heuristic derivation explains SFA's aggregation rule.

\subsection{Spatial scale as a guide to attention}
\label{sec:motivation}

To choose the spatial resolution of attention, we use the scale structure of natural images illustrated in Figure~\ref{fig:main-overview}(A,B). An isotropic finite-band approximation to their inverse-square power spectra \mbox{\citep{ruderman1994statistics}} gives logarithmic spatial correlation over intermediate distances in two dimensions (Appendix~\ref{app:spatial-correlation}). Define the distance between distinct tokens as $r_{ij}=\|\vx_i-\vx_j\|_2>0$ and fix a reference distance $r_*>0$. If query--key interactions inherit this dependence, their logits admit a radial model with query-dependent offset $b_i$, decay exponent $\alpha_i\geq0$, and content residual $u_{ij}$:
\begin{equation}
    z_{ij}=b_i-\alpha_i\log(r_{ij}/r_*)+u_{ij}, \qquad \alpha_i\geq 0.
    \label{eq:log-distance}
\end{equation}
The radial weights $\exp(z_{ij}-u_{ij})\propto r_{ij}^{-\alpha_i}$ vary by the same factor over any fixed distance ratio. Section~\ref{sec:spatial-statistics} and Appendix~\ref{app:powerlaw-details} present the corresponding empirical measurements. This power law motivates the growth of block sizes in proportion to distance. The pretrained attention profile in Figure~\ref{fig:main-overview}(C) supports this spatial allocation, within which content-dependent dot products allow the model to learn interaction strengths.

\subsection{A complete multiscale partition}
\label{sec:partition}

SFA realizes this allocation through a scale-free two-dimensional hierarchical partition similar to that in FMA \citep{kang2025fma} and H-Transformer \citep{zhu2021htransformer}. Each query belongs to a $4\times4$ base block whose $3\times3$ block neighborhood, clipped to the image boundary, supplies at most 144 exact singleton tokens. Distant regions are represented through a quadtree. At each pooled scale, we take the children of the $3\times3$ neighborhood around the query ancestor's parent and exclude the ancestor's own $3\times3$ neighborhood. This leaves at most 27 blocks per scale, with side lengths doubling at successive scales.

This construction covers the entire image. Together with the exact local tokens, the pooled blocks form a complete, disjoint partition $\mathcal P_i$ for query $i$: every original token belongs to exactly one represented block. The 16 queries in each base block share this partition while retaining their original resolution and computing their own attention weights. Figure~\ref{fig:main-overview}(D) illustrates this coverage; Appendix~\ref{app:interaction-lists} specifies the interaction lists and boundary handling.

\subsection{Pooling and normalization across scales}
\label{sec:aggregation}

Mean keys and values implement the approximation in Eq.~\eqref{eq:sfa}. Pooling follows positional encoding, including rotary position embeddings (RoPE; \citealp{su2024roformer}), so linearity preserves the mean logit: $\bar z_{iB}=\vq_i^\top\bar{\vk}_B/\sqrt d=\E_Bz_{ij}$. For singleton blocks, the means recover the original key and value and $\log m_B=0$, allowing exact local interactions and distant summaries to use the same aggregation rule. A shared softmax jointly normalizes their count-corrected logits across all scales. Related token-count corrections appear in token merging \citep{bolya2023tome}.

\subsection{Linear preprocessing and log-linear attention}
\label{sec:computation}

After pooling base blocks, each coarser mean averages four child means, allowing summaries to be constructed once and shared across queries. The geometric decrease in summary counts gives $O(N(d+d_v))$ preprocessing, while constant representatives per scale give $O(N(d+d_v)\log N)$ forward and backward cost.

Our Triton kernels reuse shared data and fuse attention across scales, with forward and backward architectures detailed in Appendix~\ref{app:implementation} (Figures~\ref{fig:kernel-forward} and~\ref{fig:kernel-backward}). SFA retains the original projections, head dimensions, and output shape, enabling direct reuse of pretrained weights.

\endgroup
\section{Approximation and Scale Structure}
\label{sec:theory}

We characterize SFA's approximation to full attention and relate its error to the relative spatial resolution of the partition. For probability vectors $\vp$ and $\vq$ with coordinates $p_j$ and $q_j$, write their Kullback--Leibler (KL) divergence as $\KL(\vp\|\vq)=\sum_jp_j\ln(p_j/q_j)$.

\paragraph{Optimality for a fixed partition.}
For a query $i$, expand each SFA block probability over its constituent tokens: $\widehat a_{ij}=\widehat p_{iB}/m_B$ for $j\in B$. Let $\widehat{\va}_i=(\widehat a_{i1},\ldots,\widehat a_{iN})^\top$ and let $\mathcal U_i$ be the family of probability distributions uniform within each block of $\mathcal P_i$.

\begin{theorem}
    \label{thm:kl_projection}
    For any finite logits and fixed partition $\mathcal P_i$, SFA uniquely minimizes reverse KL within $\mathcal U_i$:
    \begin{equation}
        \widehat{\va}_i=\argmin_{\vb\in\mathcal U_i}\KL(\vb\|\va_i), \qquad \widehat{\vy}_i=\sum_j\widehat a_{ij}\vv_j.
        \label{eq:kl_projection}
    \end{equation}
\end{theorem}

Mean pooling, token-count correction, and shared normalization jointly realize this optimum. If logits are constant within each block, SFA recovers full attention exactly for arbitrary values. More generally, it approximates the distribution induced by the original queries and keys, providing a basis for reusing pretrained weights. Partition refinement enlarges $\mathcal U_i$ and therefore cannot increase the minimum divergence, providing control over the tradeoff between accuracy and computation. Appendix~\ref{app:projection_proof} proves these projection and refinement properties.

Approximation quality depends on the key variation seen by each query. For $j\in B$, define the projected residual $\epsilon_{ij}=\vq_i^\top(\vk_j-\bar{\vk}_B)/\sqrt d$. Variations orthogonal to the query leave its logits unchanged, so substantial within-block differences in key features can coexist with accurate attention. For nonzero residuals, output error also depends on their association with the values. Appendix~\ref{app:feature_statistics} bounds these contributions, connecting the approximation to learned content-dependent interactions beyond the radial model.

\paragraph{Accuracy across spatial scales.}
For a pooled block $B$, let $\ell_B$ be its diameter and $r^-_{iB}=\min_{j\in B}r_{ij}>0$ its minimum distance from the query. Their ratio measures the relative spatial resolution. Write $p_{iB}=\sum_{j\in B}a_{ij}$ for its full-attention probability mass. Define the global and within-block value ranges as $R_V=\max_{j,t}\|\vv_j-\vv_t\|_\infty$ and $R_{V,B}=\max_{j,t\in B}\|\vv_j-\vv_t\|_\infty$, where $\|\cdot\|_\infty$ denotes the maximum absolute coordinate.

\Needspace{11\baselineskip}
\begin{theorem}
    \label{thm:scale_accuracy}
    Suppose every pooled block follows Eq.~\eqref{eq:log-distance} with $u_{ij}=0$ and satisfies $\ell_B\le\eta r^-_{iB}$ for $\eta\ge0$. Let $\delta_i=\alpha_i\ln(1+\eta)$. Then
    \begin{equation}
        \begin{aligned}
            \KL(\widehat{\va}_i\|\va_i)&\le\frac{\delta_i^2}{8},\\
            \|\vy_i-\widehat{\vy}_i\|_\infty
            &\le R_V\tanh\!\left(\frac{\delta_i^2}{32}\right)
            +\tanh\!\left(\frac{\delta_i}{4}\right)\sum_Bp_{iB}R_{V,B}.
        \end{aligned}
        \label{eq:scale_bounds}
    \end{equation}
\end{theorem}

The two output terms separate changes in probability mass across blocks from changes in value weighting within blocks. Their dependence on relative block size allows larger distant blocks to retain the same accuracy control, independently of absolute distance and token count when $\alpha_i$ and $R_V$ are uniformly bounded. If $R_{V,B}=O(\eta)$ uniformly across blocks, both terms are $O(\eta^2)$. Appendix~\ref{app:scale_proof} proves these bounds and extends them to logits with within-block content variation.

For $r>0$ in the continuous two-dimensional radial model, a complete annulus between $r$ and $2r$ has token population proportional to $r^2$ under uniform density. Its unnormalized attention mass therefore scales as $r^{2-\alpha_i}$, so distant tokens can collectively remain important despite weak individual interactions. The $\log m_B$ correction retains this population factor when the region is pooled (Appendix~\ref{app:radial-mass}).

\paragraph{A preference for coherent far-field signals.}
The same aggregation also shapes the information available during training. At fixed block masses and with weights independent of the value noise, uniform aggregation minimizes output noise variance under independent, equal-variance noise. This favors coherent far-field signals while exact local interactions retain fine detail (Appendix~\ref{app:noise_variance}). Pooling after RoPE further acts as a low-pass filter on position-modulated keys, with bandwidth decreasing as block size grows (Appendix~\ref{app:rope_pooling}). These properties give SFA a spatial inductive bias for image generation alongside its controlled approximation of pretrained attention.

\section{Experiments}
\label{sec:experiments}

\subsection{Experimental setup}

Our baselines are LLSA~\citep{zhou2026llsa}, SLA~\citep{sla}, and VSA~\citep{vsa}, using their official Triton implementations and matching the selected-block budgets of SLA and VSA to LLSA. Full attention provides an additional reference for ImageNet generation and efficiency measurements. Figure~\ref{fig:overview-results}(E) shows representative generated images.

We evaluate PixelFlow~\citep{pixelflow} on ImageNet at $256\times256$ resolution~\citep{deng2009imagenet}, replacing attention in its highest-resolution stage of 4096 tokens and retaining full attention in the other three stages. For longer spatial sequences, we use the small DiT variant (DiT-S; \citealp{peebles2023dit}) in pixel space on FFHQ~\citep{karras2019stylegan} at resolutions 128, 256, and 512, following the resolution cascade of LLSA in which each resolution continues from the preceding one (Table~\ref{tab:ffhq-schedule}). FID~\citep{heusel2017fid} and inception score (IS; \citealp{salimans2016improved}) use 50,000 generated samples, and all throughput measurements use one NVIDIA RTX 5090 GPU. Appendix~\ref{app:experiment-details} gives the detailed configurations, evaluation protocols, and fixed-rate baseline comparisons.

\subsection{Spatial statistics of pretrained attention}
\label{sec:spatial-statistics}

In pretrained PixelFlow-XL, geometric-mean radial attention $s_{\mathrm{geo}}(r)$ at distance $r$ follows $s_{\mathrm{geo}}(r)\propto r^{-\alpha}$ with $\alpha=1.414$ over 8--160 pixels (Figure~\ref{fig:main-overview}(C)). The power-law fit achieves $R^2=0.974$, compared with 0.589 for logarithmic decay, both evaluated on the original attention scale. By contrast, random initialization produces a nearly uniform radial profile, indicating that the spatial scaling emerges through training and supporting SFA's prior. Appendix~\ref{app:powerlaw-details} gives the measurement procedure and aggregate and individual fit comparisons.

\subsection{Attention and end-to-end efficiency}

Compared with FlashAttention, SFA's forward and backward speedups increase from $2.99\times$ and $2.50\times$ at 4096 tokens to $357.3\times$ and $265.6\times$ at 262,144 tokens (Figure~\ref{fig:overview-results}(A,B)). At the largest scale, SFA's forward and backward speedups over LLSA are $4.70\times$ and $4.45\times$, respectively.

End-to-end gains are bounded by projections, feed-forward networks, and other computations outside attention. The kernel gains above yield $25.75\times$ faster training and $22.54\times$ faster generation on FFHQ at $512\times512$ resolution, relative to full attention. In Stable Diffusion 3.5 Medium at $2048\times2048$, replacing both image self-attention and image-to-image interactions in joint attention (preserving all other interactions) gives $2.10\times$ training and $1.97\times$ generation speedups over full attention.

It should be emphasized that on FFHQ, SFA achieves nearly constant token throughput at a fixed training token batch, demonstrating that attention is no longer the computational bottleneck even at 262,144 tokens per sequence. Tables~\ref{tab:kernel-full}, \ref{tab:ffhq-runtime-full}, \ref{tab:pixelflow-runtime-full}, and~\ref{tab:sd35-runtime} provide the full kernel, FFHQ, PixelFlow, and Stable Diffusion 3.5 Medium measurements, respectively. Across these model-level comparisons, SFA's memory consumption remains close to the FlashAttention baseline.

\FloatBarrier

\subsection{Generation from scratch}

To assess generation quality alongside these efficiency gains, we train all five methods on ImageNet for 40 epochs. Figure~\ref{fig:overview-results}(C) summarizes their final FID and IS, while Table~\ref{tab:pixelflow-main} also reports the 10-epoch results and model throughput. We additionally continue full attention and SFA to 80 epochs to examine whether the quality advantage persists with longer training.

\begin{table}[!ht]
    \centering
    \caption{PixelFlow on ImageNet-$256$, with runtime measured for the entire model using image batch sizes 4 for training and 16 for generation. Both FID and IS use 50,000 generated samples. Quality is reported at epochs 10 and 40. Bold marks the best value in each column.}
    \label{tab:pixelflow-main}
    \setlength{\tabcolsep}{4pt}
    \begin{tabular}{lrrrrrr}
        \toprule
        & Train & Generate & \multicolumn{2}{c}{FID$_{50\mathrm{k}}$ $\downarrow$} & \multicolumn{2}{c}{IS $\uparrow$} \\
        \cmidrule(lr){4-5}\cmidrule(lr){6-7}
        Method & (ms/step) & (images/s) & 10 epochs & 40 epochs & 10 epochs & 40 epochs \\
        \midrule
        Full & 308.6 & 0.178 & 41.88 & 13.25 & 30.58 & 94.23 \\
        LLSA & 300.5 & 0.179 & 41.66 & 13.22 & 31.81 & 94.72 \\
        SLA & 316.7 & 0.171 & 41.10 & 12.68 & 31.38 & 96.13 \\
        VSA & 313.0 & 0.175 & 42.62 & 13.89 & 30.40 & 91.33 \\
        SFA (\textbf{ours}) & \textbf{286.8} & \textbf{0.194} & \textbf{40.66} & \textbf{12.51} & \textbf{32.90} & \textbf{101.20} \\
        \bottomrule
    \end{tabular}
\end{table}

\begin{figure}[!ht]
    \centering
    \includegraphics[width=\linewidth]{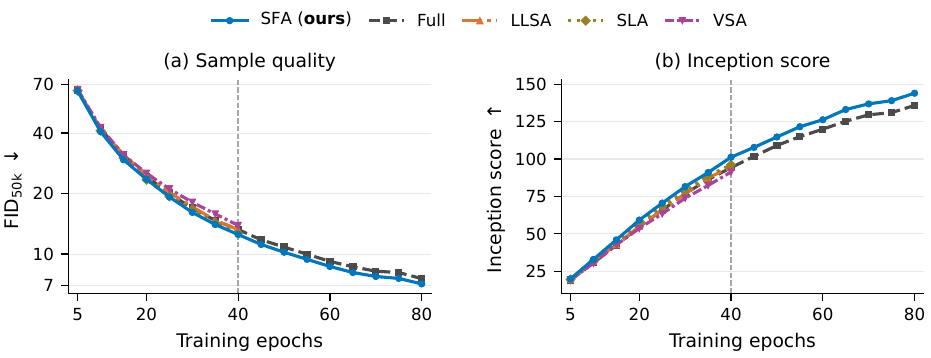}
    \caption{Training PixelFlow from scratch on ImageNet-$256$. SFA (\textbf{ours}) achieves the best FID and IS at the 40-epoch endpoint, marked by the vertical line, and improves both metrics over full attention at every recorded checkpoint through epoch 80. Each point uses 50,000 generated samples; LLSA, SLA, and VSA end at epoch 40.}
    \label{fig:pixelflow-training}
\end{figure}

At 40 epochs, SFA achieves the lowest FID of 12.51 and the highest IS of 101.20 among the five attention mechanisms. Its advantage over full attention persists through 80 epochs (Figure~\ref{fig:pixelflow-training}). SFA achieves these quality gains while improving both training and generation throughput. SLA also lowers FID relative to full attention, but with lower throughput in both training and generation. Appendix~\ref{app:pixelflow-full-results} (Tables~\ref{tab:pixelflow-runtime-full}--\ref{tab:pixelflow-extended}) provides the complete runtime and training results.

On FFHQ, SFA achieves the best generation quality among the four efficient-attention variants at resolutions 128 and 256 using $1\times1$ pixel patches. After the 20-epoch stage at each resolution, its FID$_{50\mathrm{k}}$ reaches 22.91 and 30.39, respectively (Figure~\ref{fig:overview-results}(D); Table~\ref{tab:ffhq-quality}). At resolution 512, we compare SFA and LLSA after two epochs of continuation from their respective resolution-256 checkpoints: SFA achieves FID$_{50\mathrm{k}}$ of 35.59, compared with 38.90 for LLSA. Appendix~\ref{app:ffhq-quality} gives the detailed results.

\subsection{Adapting pretrained generators}

We fine-tune the original PixelFlow model and a model with SFA in its highest-resolution stage from the same checkpoint under an identical schedule. After 3.5 epochs, full attention reaches FID$_{50\mathrm{k}}$ of 2.28 and IS of 273.25, compared with 2.29 and 284.71 for SFA, using the same 50,000-sample evaluation protocol (Table~\ref{tab:pixelflow-finetune}). SFA thus retains comparable FID and improves IS under the matched budget, preserving generation quality with the original projections and parameter count.

For Stable Diffusion 3.5 Medium~\citep{sd35medium,sd35article}, we retain the original effective patch size of $16\times16$ pixels and replace image self-attention alone (SA), which the first 13 of its 24 blocks carry, or together with image-to-image interactions in the joint attention of all 24 blocks (ALL). After one epoch of matched fine-tuning, both scopes recover the quality of the equally fine-tuned original model. ALL achieves a Dense Prompt Graph Benchmark (DPG-Bench; \citealp{dpgbench}) score of 85.51, matching full attention, and a human preference score (HPS) v2.1~\citep{wu2023hpsv2} of 30.262, compared with 30.261 for full attention.

The joint-attention replacement is specified in Appendix~\ref{app:sd35-joint-replacement}, with the complete direct-replacement and fine-tuning comparisons in Tables~\ref{tab:sd35-dpg-summary}--\ref{tab:sd35-hps}.

\subsection{Ablations}
\label{sec:ablation}
We evaluate four variants to examine count correction, normalization across scales, far-field context, and aggregation. Under the FFHQ-$128$ setup above, removing the $\log m_B$ correction raises FID$_{50\mathrm{k}}$ from 22.91 to 25.91, and normalizing each scale separately, as FMA does, raises it to 27.99. Retaining only the exact $12\times12$ local neighborhood raises it to 45.76, supporting the contribution of far-field summaries.

The first two changes also reduce quality under direct image self-attention replacement in Stable Diffusion 3.5 Medium without fine-tuning. In particular, separate softmax lowers overall HPS v2.1 from 29.596 for SFA to 11.176 and produces grid-like artifacts. On ImageNet-$256$, replacing mean pooling with learnable aggregation and removing count correction raises FID$_{50\mathrm{k}}$ from 40.66 to 41.40 and lowers IS from 32.90 to 31.50 after ten epochs. In our RTX 5090 implementation, this variant also takes $1.96$--$2.64\times$ SFA's forward attention time and $2.89$--$4.74\times$ its backward attention time across the tested sequence lengths of 4096--262,144 tokens. These results support the combination of mean pooling, count correction, and shared normalization. Appendix~\ref{app:ablation} gives the four variant definitions, detailed quality results, and efficiency analysis.

\begingroup
    \let\FloatBarrier\relax
    \section{Discussion and Limitations}
\label{sec:limitations}

The spatial hierarchy in SFA suggests a natural extension to video generation through spatiotemporal partitioning. Such a hierarchy could preserve fine interactions within nearby frames while summarizing increasingly distant regions across both space and time. Mean pooling, token-count correction, and shared normalization apply to these spatiotemporal blocks, with motion and the different rates of spatial and temporal variation guiding their shapes and resolutions. Adapting temporal coarsening to these correlations throughout denoising would be a promising direction for balancing generation quality and computational cost.

Beyond video generation, this principle could support action-conditioned visual world models, which predict how an environment evolves in response to actions \citep{bruce2024genie}. A hierarchy over past observations could retain detailed interactions with recent frames and conditioning actions while representing older visual context through coarser summaries, potentially making longer histories more affordable during prediction. In a causal formulation, each pooled summary and its token count would need to include only the history visible to its query. The resulting partition would thus need to respect both spatiotemporal structure and causal ordering.

Our study focuses on regular two-dimensional image grids, leaving irregular grids unexplored. Additionally, further acceleration is constrained by the remaining transformer components: our FFHQ experiments show that the computational bottleneck has shifted from attention to projections, feed-forward networks, and other operations. Optimizing the transformer block as a whole therefore remains a direction for future work.

\endgroup
\section{Conclusion}
\label{sec:conclusion}

By combining a scale-free spatial partition with mean aggregation, token-count correction, and a single softmax across scales, SFA gives an $O(N\log N)$ attention mechanism with a precise approximation relationship to full attention and a spatial inductive bias suited to image generation. Our analysis links approximation quality to within-block logit variation and explains why relative spatial resolution is appropriate for power-law interactions. Empirically, SFA achieves the best quality in the five-method PixelFlow comparison at 40 epochs, improves FFHQ generation, and preserves full-attention quality under matched fine-tuning budgets in PixelFlow and Stable Diffusion 3.5 Medium. Together with substantial attention and whole-model speedups, these results support its use in both training new generators and adapting pretrained ones.

\section*{Acknowledgments}
This work was supported by the National Key R\&D Program of China, Project Number 2025YFA1016700. This work was also supported by the National Natural Science Foundation of China (NSFC) under Grant No. 62576011.
This work was supported in part by the Beijing Major Science and Technology Project under Contract no. Z251100008125055. This work was also supported by Beijing Academy of Artificial Intelligence (BAAI). 

\label{main-text-end}

\subsection*{AI use statement}

In this work, we used generative AI tools for language polishing and identifying relevant literature.
We have not used generative AI tools for generating synthetic data sets, helping develop theoretical models or conceptual frameworks, formulating mathematical claims, providing critical ingredients for proving mathematical claims, assisting in the writing of proofs, proposing or refining hypotheses, designing or providing feedback on research methodology or experiments, implementing methods, assisting with translation, cleaning and reformatting datasets, supporting qualitative and thematic data analysis, or interpreting results. We have reviewed all AI-assisted work. We have reviewed the relevant literature found by the large language model. We take responsibility for the final content of this work, including text, claims or artifacts produced with the aid of generative AI.

\subsection*{Ethics statement}

This work introduces efficient attention mechanisms for diffusion transformers. Our experiments use existing public image datasets, including FFHQ, which contains photographs of real people. We acknowledge the privacy concerns and demographic biases associated with such datasets, and the importance of respecting their licenses and usage restrictions. No real images from the FFHQ dataset are displayed in this manuscript. All datasets and models utilized in this study are strictly used for non-commercial, academic research purposes in full compliance with their respective licenses. Additionally, reducing computational costs may improve access to image-generation models and reduce resource use, but may also lower the cost of producing deceptive or harmful synthetic content. We therefore encourage responsible use and consideration of privacy, bias, and misuse risks in downstream applications.

\subsection*{Reproducibility statement}

Section~\ref{sec:method} specifies SFA's multiscale partition, mean pooling, and count-corrected shared softmax. Appendix~\ref{app:theory} states the assumptions and provides proofs of the theoretical results, while Appendix~\ref{app:implementation} details the forward and backward Triton computations. To support reproduction of the experiments, Appendix~\ref{app:experiment-details} documents the hardware and software environments, baseline configurations, training and fine-tuning schedules, and sampling and evaluation protocols, with training hyperparameters in Table~\ref{tab:training-settings}.

\bibliography{references}
\bibliographystyle{sfa}

\appendix

\section{Theoretical Details}
\label{app:theory}

\paragraph{Organization of the analysis.}
We develop the approximation results by following how pooling changes an attention distribution. Appendix~\ref{app:spatial-correlation} first connects inverse-square image spectra to the logarithmic spatial model. Appendix~\ref{app:projection_proof} then proves SFA's reverse-KL optimality for arbitrary finite logits. The quantitative error analysis proceeds through two probability inequalities (Appendix~\ref{app:probability_lemmas}), an exact separation of block-mass and within-block errors (Appendix~\ref{app:error_decomposition}), and the scale-consistent bounds of Theorem~\ref{thm:scale_accuracy} (Appendix~\ref{app:scale_proof}). Appendix~\ref{app:feature_statistics} expresses the same error through query-projected feature statistics. Appendices~\ref{app:radial-mass}--\ref{app:logarithmic_geometry} then explain the associated spatial bias: mass across distance shells, noise reduction by averaging, the frequency response after rotary encoding, and the special geometry of the two-dimensional logarithmic model.

\paragraph{Notation and scope.}
We fix one attention head and one query $i$, using the token indices, head dimensions, queries, keys, and values from Section~\ref{sec:method}. Positional encoding is included in the queries and keys. The partition $\mathcal P_i$ consists of disjoint nonempty blocks covering the effective key set; $m_B=|B|$, and each exact near-field interaction is a singleton block. Unless specified otherwise, sums over $B$ range over $\mathcal P_i$, and the queries, keys, and values are held fixed. Logarithms are natural. For probability vectors $\vp$ and $\vq$ with coordinates $p_j$ and $q_j$, we use $\KL(\vp\|\vq)=\sum_jp_j\ln(p_j/q_j)$ and $\operatorname{TV}(\vp,\vq)=\frac12\sum_j|p_j-q_j|$ for Kullback--Leibler (KL) divergence and total variation (TV), respectively.

The operator $\E_B$ denotes uniform finite averaging over tokens in $B$, while $\E_{\vp}$ denotes averaging with probabilities $\vp$. For a scalar function $f$ and a vector function $\vg$ of the same sampled index, define $\Var_{\vp}(f)=\E_{\vp}(f-\E_{\vp}f)^2$ and $\Cov_{\vp}(f,\vg)=\E_{\vp}[(f-\E_{\vp}f)(\vg-\E_{\vp}\vg)]$. Subscript $B$ denotes the corresponding uniform block statistics. An unsubscripted $\E$ refers to the distribution specified for the random variable. These conventions also apply to block-indexed quantities when $\vp$ is a distribution over blocks.

\subsection{From inverse-square spectra to logarithmic correlation}
\label{app:spatial-correlation}

We begin with the spatial model that motivates distance-dependent pooling. Consider a centered, stationary, isotropic scalar random field $\rx(\vx)$ at positions $\vx\in\mathbb R^2$. Suppose its power spectral density is $S(\vomega)=A\|\vomega\|_2^{-2}$ for angular frequencies satisfying $0<k_{\min}\leq\|\vomega\|_2\leq k_{\max}<\infty$, and zero elsewhere, where $A>0$. The finite cutoffs make the covariance well defined while retaining the approximate inverse-square scaling of natural images \citep{ruderman1994statistics}.

Let $\vzero$ denote the origin and use the Fourier convention $C(\vx)=\E[\rx(\vzero)\rx(\vx)]=(2\pi)^{-2}\int e^{\mathrm{i}\vomega^\top\vx}S(\vomega)\,\diff\vomega$, where $\mathrm{i}^2=-1$. Isotropy makes this covariance a function of distance $r=\|\vx\|_2$. For $r>0$, rotate the coordinates so that $\vx=(r,0)^\top$ and write the frequency in polar form as $\vomega=k(\cos\theta,\sin\theta)^\top$, where $k>0$ and $\theta\in[0,2\pi)$. Then $\vomega^\top\vx=kr\cos\theta$ and the area element is $\diff\vomega=k\,\diff k\,\diff\theta$. Multiplying this Jacobian by the spectral factor $A/k^2$ leaves the radial measure $A\,\diff k/k$.

The angular integral is given by the integral representation of the order-zero Bessel function of the first kind,
\begin{equation*}
    J_0(t)=\frac1{2\pi}\int_0^{2\pi}e^{\mathrm{i}t\cos\theta}\,\diff\theta,
    \qquad t\in\mathbb R.
\end{equation*}
Substituting the polar coordinates and then using this identity at $t=kr$ gives
\begin{subequations}
    \label{eq:inverse-square-correlation}
    \begin{align}
        C(r)
        &=\frac{A}{(2\pi)^2}\int_{k_{\min}}^{k_{\max}}
        \left[\int_0^{2\pi}e^{\mathrm{i}kr\cos\theta}\,\diff\theta\right]\frac{\diff k}{k}\notag\\
        &=\frac{A}{2\pi}\int_{k_{\min}}^{k_{\max}}J_0(kr)\,\frac{\diff k}{k}.
        \label{eq:inverse-square-covariance}
    \end{align}

    To differentiate the radial covariance, set $t=kr$. Since $\diff k/k=\diff t/t$, Eq.~\eqref{eq:inverse-square-covariance} becomes
    \begin{equation*}
        C(r)=\frac{A}{2\pi}\int_{k_{\min}r}^{k_{\max}r}\frac{J_0(t)}{t}\,\diff t.
    \end{equation*}
    The integrand now has no dependence on $r$, and $r$ enters through both integration limits. The Leibniz rule gives the upper-limit contribution minus the lower-limit contribution, each multiplied by the derivative of its limit:
    \begin{align}
        C'(r)
        &=\frac{A}{2\pi}\left[
        k_{\max}\frac{J_0(k_{\max}r)}{k_{\max}r}
        -k_{\min}\frac{J_0(k_{\min}r)}{k_{\min}r}\right]\notag\\
        &=-\frac{A}{2\pi r}\big[J_0(k_{\min}r)-J_0(k_{\max}r)\big].
        \label{eq:inverse-square-covariance-derivative}
    \end{align}
\end{subequations}
In the intermediate range $k_{\min}r\ll1\ll k_{\max}r$, the low-frequency term approaches one and the high-frequency term decays in amplitude. Thus Eq.~\eqref{eq:inverse-square-covariance-derivative} gives $C'(r)\simeq-A/(2\pi r)$, which integrates to $C(r)\simeq C_0-A\ln(r/r_*)/(2\pi)$, with reference distance $r_*>0$ and additive constant $C_0$.

This calculation motivates the logarithmic spatial component of Eq.~\eqref{eq:log-distance}. Its coefficient in a learned attention logit also depends on feature projections, positional encoding, and dot-product scaling, so $\alpha_i$ remains a learned decay exponent. We next establish the approximation identity for arbitrary logits; the radial model will enter when we bound their variation within each block.

\subsection{Projection and reverse-KL optimality}
\label{app:projection_proof}

We use the same attention definitions as in Sections~\ref{sec:method} and~\ref{sec:theory}, restated here for the fixed query $i$. The full-attention logit is $z_{ij}=\vq_i^\top\vk_j/\sqrt d$. For a block $B$, let $\bar{\vk}_B=\E_B\vk_j$ and $\bar{\vv}_B=\E_B\vv_j$ be its mean key and value, and let $\bar z_{iB}=\E_Bz_{ij}=\vq_i^\top\bar{\vk}_B/\sqrt d$. Full attention assigns a weight $a_{ij}$ to each token, while SFA assigns a probability mass $\widehat p_{iB}$ to each block:
\begin{equation}
    \begin{aligned}
        Z_i&=\sum_{j=1}^N e^{z_{ij}},
        &a_{ij}&=\frac{e^{z_{ij}}}{Z_i},\\
        \widehat Z_i&=\sum_{B\in\mathcal P_i}m_Be^{\bar z_{iB}},
        &\widehat p_{iB}&=\frac{m_Be^{\bar z_{iB}}}{\widehat Z_i}.
    \end{aligned}
    \label{eq:appendix-attention-definitions}
\end{equation}
The corresponding outputs are $\vy_i=\sum_j a_{ij}\vv_j$ and $\widehat{\vy}_i=\sum_B\widehat p_{iB}\bar{\vv}_B$.

To compare the distributions on a common token set, expand each SFA block mass uniformly as $\widehat a_{ij}=\widehat p_{iB}/m_B$ for $j\in B$. Conversely, summing full-attention token weights gives the block mass $p_{iB}=\sum_{j\in B}a_{ij}$. Thus $\va_i=(a_{ij})_{j=1}^N$ and $\widehat{\va}_i=(\widehat a_{ij})_{j=1}^N$ are token probability vectors, whereas $\vp_i=(p_{iB})_{B\in\mathcal P_i}$ and $\widehat{\vp}_i=(\widehat p_{iB})_{B\in\mathcal P_i}$ are block probability vectors. As in Section~\ref{sec:theory}, $\mathcal U_i$ denotes the family of token distributions uniform within each block of $\mathcal P_i$.

Define the block-averaging matrix $\mPi_i\in\mathbb R^{N\times N}$ by $(\mPi_i\vg)_j=m_B^{-1}\sum_{t\in B}g_t$ for $j\in B$ and $\vg\in\mathbb R^N$. For the logit vector $\vz_i=(z_{ij})_{j=1}^N$, $\mPi_i\vz_i$ repeats $\bar z_{iB}$ at every token in $B$. This representation expresses SFA directly through averaged logits on the original token set.

\begin{proof}[Proof of Theorem~\ref{thm:kl_projection}]
    Substituting the SFA block mass from Eq.~\eqref{eq:appendix-attention-definitions} into $\widehat a_{ij}=\widehat p_{iB}/m_B$ gives
    \begin{equation}
        \widehat a_{ij}=\frac{e^{\bar z_{iB}}}{\widehat Z_i}
        =\softmax(\mPi_i\vz_i)_j,\qquad j\in B.
        \label{eq:expanded_attention}
    \end{equation}
    Mean pooling of the values preserves this expansion at the output: $\sum_B\widehat p_{iB}\bar{\vv}_B=\sum_j\widehat a_{ij}\vv_j$.

    Any distribution $\vb\in\mathcal U_i$ can likewise be written as $b_j=\pi_B/m_B$ for $j\in B$, where $\vpi=(\pi_B)_{B\in\mathcal P_i}$ is a probability vector over blocks, so $\sum_B\pi_B=1$. Because the blocks partition the token set, we can group the token-level KL sum by block. Substituting $b_j=\pi_B/m_B$ and $a_{ij}=e^{z_{ij}}/Z_i$, then expanding the logarithm, gives
    \begin{align}
        \KL(\vb\|\va_i)
        &=\sum_B\sum_{j\in B}\frac{\pi_B}{m_B}
        \ln\frac{\pi_B/m_B}{e^{z_{ij}}/Z_i}\notag\\
        &=\sum_B\sum_{j\in B}\frac{\pi_B}{m_B}
        \left(\ln\frac{\pi_B}{m_B}-z_{ij}+\ln Z_i\right)\notag\\
        &=\sum_B\left[
        \pi_B\ln\frac{\pi_B}{m_B}
        -\frac{\pi_B}{m_B}\sum_{j\in B}z_{ij}
        +\pi_B\ln Z_i\right]\notag\\
        &=\sum_B\pi_B\ln\frac{\pi_B}{m_B}
        -\sum_B\pi_B\bar z_{iB}+\ln Z_i.
        \label{eq:kl_token_block_expansion}
    \end{align}
    Here the terms constant within $B$ are repeated $m_B$ times, the averaged logit is $m_B^{-1}\sum_{j\in B}z_{ij}=\bar z_{iB}$, and $\sum_B\pi_B\ln Z_i=\ln Z_i$. Terms from blocks with $\pi_B=0$ are interpreted by continuity.

    To identify the block-level divergence in Eq.~\eqref{eq:kl_token_block_expansion}, take logarithms of the SFA mass in Eq.~\eqref{eq:appendix-attention-definitions}: $\ln\widehat p_{iB}=\ln m_B+\bar z_{iB}-\ln\widehat Z_i$. Expanding the KL between the two block distributions therefore yields
    \begin{align*}
        \KL(\vpi\|\widehat{\vp}_i)
        &=\sum_B\pi_B\left[\ln\pi_B-\ln m_B-\bar z_{iB}+\ln\widehat Z_i\right]\\
        &=\sum_B\pi_B\ln\frac{\pi_B}{m_B}
        -\sum_B\pi_B\bar z_{iB}+\ln\widehat Z_i,
    \end{align*}
    where the last term again uses $\sum_B\pi_B=1$. Comparing this expression with Eq.~\eqref{eq:kl_token_block_expansion} leaves the difference between the two log normalizers:
    \begin{equation}
        \KL(\vb\|\va_i)=\KL(\vpi\|\widehat{\vp}_i)+\ln Z_i-\ln\widehat Z_i.
        \label{eq:kl_pythagorean}
    \end{equation}
    Only the first term on the right-hand side of Eq.~\eqref{eq:kl_pythagorean} depends on $\vpi$. It is nonnegative and vanishes precisely at $\vpi=\widehat{\vp}_i$, proving uniqueness of SFA's expanded distribution and giving
    \begin{equation}
        \KL(\widehat{\va}_i\|\va_i)=\ln\frac{Z_i}{\widehat Z_i}.
        \label{eq:kl_partition_function}
    \end{equation}
    Since $\vb$ and $\widehat{\va}_i$ share uniform conditionals within each block, $\KL(\vpi\|\widehat{\vp}_i)=\KL(\vb\|\widehat{\va}_i)$. Therefore the approximation also satisfies
    \begin{equation}
        \KL(\vb\|\va_i)=\KL(\vb\|\widehat{\va}_i)+\KL(\widehat{\va}_i\|\va_i).
        \label{eq:kl_projection_identity}
    \end{equation}
\end{proof}

Two consequences explain how the approximation responds to the partition. Refinement enlarges the family of blockwise uniform token distributions, so the minimum reverse KL cannot increase. If the logits are constant within each block, then $\mPi_i\vz_i=\vz_i$ and Eq.~\eqref{eq:expanded_attention} gives exact recovery of full attention, including its output for arbitrary values.

\subsection{Two probability inequalities}
\label{app:probability_lemmas}

The identity in Eq.~\eqref{eq:kl_partition_function} reduces token-level KL error to a ratio of normalizers. Bounding this ratio requires control of centered exponential moments; bounding attention outputs additionally requires control of changes in normalized probabilities. The following two lemmas provide these steps.

\begin{lemma}[Hoeffding's lemma; \citealp{Hoeffding}]
    \label{prop:bounded_mgf}
    If a scalar random variable $\rx$ satisfies $\E\rx=0$ and lies in an interval of width $\Delta\ge0$, then
    \begin{equation}
        0\le\ln\E e^{\rx}\le\frac{\Delta^2}{8}.
        \label{eq:bounded_mgf}
    \end{equation}
\end{lemma}

\begin{proof}
    Define $g(t)=\ln\E e^{t\rx}$ for $t\in[0,1]$. Centering gives $g(0)=g'(0)=0$, and differentiating twice expresses $g''(t)$ as the variance of $\rx$ under weights proportional to $e^{t\rx}$. Reweighting preserves the support interval. Its midpoint is at most $\Delta/2$ from every supported value, so the second moment about that midpoint is at most $\Delta^2/4$. Variance is no larger than the second moment about any chosen center; hence $g''(t)\le\Delta^2/4$. Integrating twice gives $g(1)=\int_0^1(1-t)g''(t)\,\diff t\le\Delta^2/8$. Jensen's inequality gives $g(1)\ge\ln e^{\E\rx}=0$.
\end{proof}

\begin{lemma}
    \label{prop:softmax_tv}
    For finite vectors $\vz,\vu\in\mathbb R^n$, let $\vq=\softmax(\vz)$ and $\vp=\softmax(\vz+\vu)$. With $\operatorname{osc}(\vu)=\max_j u_j-\min_j u_j$, where $u_j=(\vu)_j$,
    \begin{equation}
        \operatorname{TV}(\vp,\vq)\le\tanh\!\left(\frac{\operatorname{osc}(\vu)}{4}\right).
        \label{eq:softmax_tv}
    \end{equation}
\end{lemma}

\begin{proof}
    Draw an index $\rj$ with probabilities $\vq$ and define $\rx=e^{u_{\rj}}$, $l=\min_j e^{u_j}$, $h=\max_j e^{u_j}$, and $\mu_{\rx}=\E_{\vq}\rx$. The relation $p_j/q_j=e^{u_j}/\mu_{\rx}$ gives $\operatorname{TV}(\vp,\vq)=\E_{\vq}|\rx-\mu_{\rx}|/(2\mu_{\rx})$. When $h>l$, convexity bounds $|x-\mu_{\rx}|$ on $[l,h]$ by the line segment joining its endpoint values. Taking the expectation of this bound for a variable with mean $\mu_{\rx}$ gives
    \begin{equation}
        \operatorname{TV}(\vp,\vq)
        \le\frac{(h-\mu_{\rx})(\mu_{\rx}-l)}{\mu_{\rx}(h-l)}
        \le\frac{\sqrt h-\sqrt l}{\sqrt h+\sqrt l}.
        \label{eq:tv_chord_bound}
    \end{equation}
    The first expression on the right is maximal at $\mu_{\rx}=\sqrt{lh}$. The final ratio equals $\tanh(\ln(h/l)/4)$, and $\ln(h/l)=\operatorname{osc}(\vu)$. If $h=l$, the perturbation is constant and the two probability vectors coincide.
\end{proof}

We will apply TV to values through their coordinate ranges. For scalar values $v_j$ of range $R=\max_jv_j-\min_jv_j$, subtracting the midpoint of this interval from every value leaves $\sum_j(p_j-q_j)v_j$ unchanged. The triangle inequality then gives $|\sum_j(p_j-q_j)v_j|\le(R/2)\sum_j|p_j-q_j|=R\operatorname{TV}(\vp,\vq)$. Applying this argument to each coordinate yields the vector bounds below.

\subsection{Exact decomposition of approximation error}
\label{app:error_decomposition}

Pooling can change both the total probability assigned to a block and the weighting of values inside it. To separate these effects, write the logit as its block mean plus a centered residual. For $j\in B$, define
\begin{equation}
    \epsilon_{ij}=z_{ij}-\bar z_{iB},\qquad
    J_{iB}=\ln\E_B e^{\epsilon_{ij}}.
    \label{eq:block_residuals}
\end{equation}
By construction, $\E_B\epsilon_{ij}=0$. The original unnormalized block mass is $W_{iB}=\sum_{j\in B}e^{z_{ij}}=m_Be^{\bar z_{iB}}\E_B e^{\epsilon_{ij}}$, whereas SFA uses $\widehat W_{iB}=m_Be^{\bar z_{iB}}$. Thus $e^{J_{iB}}=W_{iB}/\widehat W_{iB}$ is exactly the multiplicative mass correction omitted by pooling. Jensen's inequality gives $J_{iB}\ge0$.

Let $F_i=\sum_B\widehat p_{iB}e^{J_{iB}}$ be the average of these corrections under the SFA block distribution. Restoring the full mass of every block and renormalizing gives
\begin{equation}
    p_{iB}=\frac{\widehat p_{iB}e^{J_{iB}}}{F_i},\qquad
    F_i=\frac{Z_i}{\widehat Z_i}\ge1.
    \label{eq:exact_block_mass}
\end{equation}
Hence the full block distribution is an exponential reweighting of SFA's block distribution by $J_{iB}$. A correction identical across all blocks cancels in this normalization.

Within a block, full attention has conditional probabilities $s_{ij\mid B}=a_{ij}/p_{iB}=e^{\epsilon_{ij}}/(m_Be^{J_{iB}})$. Let $\vs_{i\mid B}=(s_{ij\mid B})_{j\in B}$ and let $\vu_B$ be the uniform distribution with entries $1/m_B$. The difference between their conditional value averages is
\begin{equation}
    \vc_{iB}
    =\sum_{j\in B}s_{ij\mid B}\vv_j-\bar{\vv}_B
    =\frac{\E_B[e^{\epsilon_{ij}}(\vv_j-\bar{\vv}_B)]}{e^{J_{iB}}}
    \in\mathbb R^{d_v}.
    \label{eq:conditional_value_correction}
\end{equation}
Full attention therefore outputs $\sum_Bp_{iB}(\bar{\vv}_B+\vc_{iB})$. Subtracting SFA's output gives the exact decomposition
\begin{equation}
    \vy_i-\widehat{\vy}_i
    =\sum_B(p_{iB}-\widehat p_{iB})\bar{\vv}_B+\sum_Bp_{iB}\vc_{iB}.
    \label{eq:exact_output_decomposition}
\end{equation}
The first term changes the mixture of block means; the second changes the value average inside each block. Singleton blocks have $\epsilon_{ij}=J_{iB}=0$ and $\vc_{iB}=\vzero$, so their within-block approximation is exact.

The divergence admits the same separation. Since $\KL(\vu_B\|\vs_{i\mid B})=J_{iB}$, the KL chain rule and Eq.~\eqref{eq:exact_block_mass} yield
\begin{align}
    \KL(\widehat{\va}_i\|\va_i)
    &=\KL(\widehat{\vp}_i\|\vp_i)+\sum_B\widehat p_{iB}J_{iB},\\
    \KL(\widehat{\vp}_i\|\vp_i)
    &=\ln\E_{\widehat{\vp}_i}e^{J_{iB}}-\E_{\widehat{\vp}_i}J_{iB}.
    \label{eq:kl_block_decomposition}
\end{align}
The token-level divergence includes the average error of replacing each conditional distribution by a uniform one. The block-level divergence depends only on how the corrections $J_{iB}$ vary across blocks. This distinction explains the different orders of the bounds in the next section.

\subsection{Proof of scale-consistent accuracy}
\label{app:scale_proof}

We now connect the preceding identities to relative spatial resolution. Recall $p_{iB}=\sum_{j\in B}a_{ij}$ and the value ranges from Section~\ref{sec:theory}. Equivalently, writing $v_{jc}=(\vv_j)_c$ for $c\in\{1,\ldots,d_v\}$, these ranges are $R_V=\max_c(\max_jv_{jc}-\min_jv_{jc})$ and $R_{V,B}=\max_c(\max_{j\in B}v_{jc}-\min_{j\in B}v_{jc})$. They measure the largest coordinate variation globally and within a block. The norm $\|\vg\|_\infty=\max_c|g_c|$ uses the same coordinates. A common shift of all values changes neither these ranges nor the attention-output error.

\begin{proof}[Proof of Theorem~\ref{thm:scale_accuracy}]
    First, relative geometry controls the range of the logits. Define $r^+_{iB}=\max_{j\in B}r_{ij}$ for a pooled block. The triangle inequality gives $r^+_{iB}\le r^-_{iB}+\ell_B\le(1+\eta)r^-_{iB}$. With $u_{ij}=0$ in the logarithmic model,
    \begin{equation}
        \Delta_{iB}:=\max_{j\in B}z_{ij}-\min_{j\in B}z_{ij}
        =\alpha_i\ln\frac{r^+_{iB}}{r^-_{iB}}\le\delta_i.
        \label{eq:geometric_logit_range}
    \end{equation}
    Subtracting the mean preserves this range. Singleton blocks have $\Delta_{iB}=0$, including the query's own position, and require no radial assumption.

    Second, the logit range controls the correction to each block mass. Lemma~\ref{prop:bounded_mgf} gives $0\le J_{iB}\le\delta_i^2/8$. The projection identity and the definition of $F_i$ then imply
    \begin{equation}
        \KL(\widehat{\va}_i\|\va_i)
        =\ln F_i=\ln\E_{\widehat{\vp}_i}e^{J_{iB}}
        \le\frac{\delta_i^2}{8}.
        \label{eq:total_kl_bound}
    \end{equation}
    At the block level, Eq.~\eqref{eq:exact_block_mass} is a softmax perturbation with logits $(\ln\widehat p_{iB})_B$ and perturbations $(J_{iB})_B$. The range of the perturbations $J_{iB}$ is at most $\delta_i^2/8$, so Lemma~\ref{prop:softmax_tv} yields
    \begin{equation}
        \operatorname{TV}(\widehat{\vp}_i,\vp_i)
        \le\tanh\!\left(\frac{\delta_i^2}{32}\right).
        \label{eq:scale_probability}
    \end{equation}

    Third, within each block, $\vs_{i\mid B}$ is the softmax of $(\epsilon_{ij})_{j\in B}$, while $\vu_B$ is the softmax of a zero vector. Applying the same lemma and the coordinate-range inequality gives
    \begin{equation}
        \|\vc_{iB}\|_\infty
        \le R_{V,B}\operatorname{TV}(\vs_{i\mid B},\vu_B)
        \le R_{V,B}\tanh\!\left(\frac{\delta_i}{4}\right).
        \label{eq:within_block_output_bound}
    \end{equation}
    The block means have coordinate ranges at most $R_V$. Applying the coordinate-range inequality to the first term of Eq.~\eqref{eq:exact_output_decomposition}, and the triangle inequality to its second term, now gives
    \begin{align}
        \|\vy_i-\widehat{\vy}_i\|_\infty
        &\le R_V\operatorname{TV}(\vp_i,\widehat{\vp}_i)
        +\sum_Bp_{iB}\|\vc_{iB}\|_\infty \notag\\
        &\le R_V\tanh\!\left(\frac{\delta_i^2}{32}\right)
        +\tanh\!\left(\frac{\delta_i}{4}\right)\sum_Bp_{iB}R_{V,B}.
        \label{eq:scale_output}
    \end{align}
    Together with Eq.~\eqref{eq:total_kl_bound}, this proves Eq.~\eqref{eq:scale_bounds}.
\end{proof}

The proof also controls unnormalized block masses and reveals a tighter KL order after summing probabilities within blocks.

\begin{proposition}
    \label{prop:scale_refined}
    Under the assumptions of Theorem~\ref{thm:scale_accuracy}, the masses $W_{iB}$ and $\widehat W_{iB}$ defined in Appendix~\ref{app:error_decomposition} satisfy
    \begin{equation}
        0\le\ln\frac{W_{iB}}{\widehat W_{iB}}=J_{iB}\le\frac{\delta_i^2}{8},
        \qquad \KL(\widehat{\vp}_i\|\vp_i)\le\frac{\delta_i^4}{512}.
        \label{eq:mass_and_fourth_order_kl}
    \end{equation}
\end{proposition}

\begin{proof}
    The first bound is the exponential-moment estimate used above. For the second, center the block correction by setting $\mu_J=\E_{\widehat{\vp}_i}J_{iB}$. Equation~\eqref{eq:kl_block_decomposition} becomes $\KL(\widehat{\vp}_i\|\vp_i)=\ln\E_{\widehat{\vp}_i}e^{J_{iB}-\mu_J}$. This centered variable has range at most $\delta_i^2/8$, so Lemma~\ref{prop:bounded_mgf} bounds its log moment by $(\delta_i^2/8)^2/8=\delta_i^4/512$.
\end{proof}

Thus the expanded token distribution has a second-order KL bound in $\delta_i$, while the block distribution has a fourth-order KL bound. Within-block value variation determines how this difference carries over to the output. With only a global range bound, Pinsker's inequality and the coordinate-range inequality give $\|\vy_i-\widehat{\vy}_i\|_\infty\le R_V\min\{1,\delta_i/4\}$. If, in addition, $R_{V,B}\le L_V\eta$ for a constant $L_V\ge0$ uniform over blocks, Eq.~\eqref{eq:scale_output} gives
\begin{equation}
    \|\vy_i-\widehat{\vy}_i\|_\infty
    \le R_V\tanh\!\left(\frac{\delta_i^2}{32}\right)
    +L_V\eta\tanh\!\left(\frac{\delta_i}{4}\right)
    =O(\eta^2).
    \label{eq:second_order_output}
\end{equation}
Here the limit is $\eta\to0$: $\delta_i=\alpha_i\ln(1+\eta)=O(\eta)$, and each displayed term is quadratic. The constants are independent of token count when $\alpha_i$, $R_V$, and $L_V$ are uniformly bounded.

Content-dependent logits enter through the same range calculation. For a pooled block with residual $u_{ij}$ in Eq.~\eqref{eq:log-distance}, define $\operatorname{osc}_{j\in B}u_{ij}=\max_{j\in B}u_{ij}-\min_{j\in B}u_{ij}$. Then $\Delta_{iB}\le\alpha_i\ln(1+\eta)+\operatorname{osc}_{j\in B}u_{ij}$. Replacing $\delta_i$ by a common upper bound on these ranges gives the same KL, TV, and output inequalities. This extension identifies within-block content variation as the additional quantity needed beyond relative geometry. The $O(\eta^2)$ output conclusion also holds when $\operatorname{osc}_{j\in B}u_{ij}=O(\eta)$ uniformly, together with the preceding within-block value condition.

\subsection{Projected feature statistics and output error}
\label{app:feature_statistics}

The preceding section bounds residual logits through a spatial model. We now express their effect directly in terms of the learned keys and values, using the exact decomposition from Appendix~\ref{app:error_decomposition}. This analysis applies to arbitrary finite logits.

For each block, define $\rho_{iB}=\max_{j\in B}|\epsilon_{ij}|$, $\sigma_{z,iB}^2=\E_B\epsilon_{ij}^2$, and $\sigma_{V,B}=\max_c\sqrt{\E_B(v_{jc}-\bar v_{Bc})^2}$, where $\bar v_{Bc}=(\bar{\vv}_B)_c$. These are the maximum absolute logit residual, its variance, and the largest coordinate standard deviation of the values. Let $\mSigma_{K,B}=\E_B[(\vk_j-\bar{\vk}_B)(\vk_j-\bar{\vk}_B)^\top]\in\mathbb R^{d\times d}$ and $\mC_{KV,B}=\E_B[(\vk_j-\bar{\vk}_B)(\vv_j-\bar{\vv}_B)^\top]\in\mathbb R^{d\times d_v}$ be the key covariance and key--value cross-covariance. Substituting $\epsilon_{ij}=\vq_i^\top(\vk_j-\bar{\vk}_B)/\sqrt d$ gives
\begin{equation}
    \sigma_{z,iB}^2=\frac{\vq_i^\top\mSigma_{K,B}\vq_i}{d},
    \qquad \Cov_B(\epsilon_{ij},\vv_j)=\frac{\mC_{KV,B}^{\top}\vq_i}{\sqrt d}.
    \label{eq:projected_feature_statistics}
\end{equation}
The query therefore selects which directions of key variation affect the approximation, and the cross-covariance determines how those variations couple to the values.

\paragraph{A bound from block statistics.}
To connect these quantities to the output, write $f_{iB}=e^{J_{iB}}=\E_Be^{\epsilon_{ij}}$ and $\vd_{iB}=f_{iB}\vc_{iB}=\E_B[e^{\epsilon_{ij}}(\vv_j-\bar{\vv}_B)]$. Centering removes the linear term from $f_{iB}-1$ and the constant term from $\vd_{iB}$. Taylor's theorem and coordinatewise Cauchy--Schwarz then give
\begin{equation}
    0\le f_{iB}-1\le\frac12e^{\rho_{iB}}\sigma_{z,iB}^2,
    \qquad \|\vd_{iB}\|_\infty\le e^{\rho_{iB}}\sigma_{z,iB}\sigma_{V,B}.
    \label{eq:moment_bounds}
\end{equation}
For the first inequality, use $f_{iB}-1=\E_B(e^{\epsilon_{ij}}-1-\epsilon_{ij})$ and $|e^x-1-x|\le e^\rho x^2/2$ for $|x|\le\rho$. For the second, write $\vd_{iB}=\E_B[(e^{\epsilon_{ij}}-1)(\vv_j-\bar{\vv}_B)]$, apply $|e^x-1|\le e^\rho|x|$, and bound each coordinate by Cauchy--Schwarz.

The full output can be written as $\vy_i=F_i^{-1}\sum_B\widehat p_{iB}(f_{iB}\bar{\vv}_B+\vd_{iB})$. Since $\sum_B\widehat p_{iB}(f_{iB}-1)=F_i-1$, subtracting $\widehat{\vy}_i$ yields the useful identity
\begin{equation}
    \vy_i-\widehat{\vy}_i
    =\frac1{F_i}\sum_B\widehat p_{iB}
    \big[(f_{iB}-1)(\bar{\vv}_B-\widehat{\vy}_i)+\vd_{iB}\big].
    \label{eq:weighted_output_identity}
\end{equation}
Here $F_i\ge1$, and $\widehat{\vy}_i$ is a convex combination of block means, so $\|\bar{\vv}_B-\widehat{\vy}_i\|_\infty\le R_V$. Substituting Eq.~\eqref{eq:moment_bounds} gives
\begin{equation}
    \|\vy_i-\widehat{\vy}_i\|_\infty
    \le\sum_B\widehat p_{iB}e^{\rho_{iB}}
    \left(\frac{R_V}{2}\sigma_{z,iB}^2+\sigma_{z,iB}\sigma_{V,B}\right).
    \label{eq:weighted_statistical_bound}
\end{equation}
Each block contributes according to its SFA probability mass, its projected logit variation, and its value variation. Singleton contributions vanish. This gives a content-dependent counterpart to the geometric bound without requiring the radial logit model.

\paragraph{Leading terms for small residuals.}
Consider a family of attention problems as $\eta\to0$. Assume $\rho_{iB}=O(\eta)$, $\max_{j\in B}\|\vv_j-\bar{\vv}_B\|_\infty=O(\eta)$, and $R_V=O(1)$, with constants uniform over blocks and tokens. All vector remainder estimates below use the infinity norm. These assumptions imply $\sigma_{z,iB}=O(\eta)$ and $\sigma_{V,B}=O(\eta)$, and Taylor expansion gives
\begin{equation}
    f_{iB}-1=\frac12\sigma_{z,iB}^2+O(\eta^3),
    \qquad \vd_{iB}=\Cov_B(\epsilon_{ij},\vv_j)+O(\eta^3).
    \label{eq:block_moment_expansion}
\end{equation}
The scalar remainder is bounded by $e^{\rho_{iB}}\rho_{iB}\sigma_{z,iB}^2/6$, and the vector remainder by $e^{\rho_{iB}}\rho_{iB}\sigma_{z,iB}\sigma_{V,B}/2$. Their orders are therefore uniform. Since $\sum_B\widehat p_{iB}=1$ and $\|\bar{\vv}_B-\widehat{\vy}_i\|_\infty\le R_V=O(1)$, probability weighting preserves the $O(\eta^3)$ remainders in Eq.~\eqref{eq:weighted_output_identity}. Its numerator is $O(\eta^2)$, while averaging the scalar expansion gives $F_i=1+O(\eta^2)$ and hence $F_i^{-1}=1+O(\eta^2)$. Replacing $F_i^{-1}$ by one therefore contributes only an $O(\eta^4)$ term. Finally, $\widehat{\vy}_i=\E_{\widehat{\vp}_i}\bar{\vv}_B$ identifies the variance-weighted term as a covariance:
\begin{equation*}
    \sum_B\widehat p_{iB}\sigma_{z,iB}^2
    (\bar{\vv}_B-\widehat{\vy}_i)
    =\Cov_{\widehat{\vp}_i}(\sigma_{z,iB}^2,\bar{\vv}_B).
\end{equation*}
Substituting these relations into Eq.~\eqref{eq:weighted_output_identity} yields
\begin{equation}
    \vy_i-\widehat{\vy}_i
    =\sum_B\widehat p_{iB}\Cov_B(\epsilon_{ij},\vv_j)
    +\frac12\Cov_{\widehat{\vp}_i}(\sigma_{z,iB}^2,\bar{\vv}_B)
    +O(\eta^3).
    \label{eq:second_order_expansion}
\end{equation}
The first term measures within-block coupling between logit and value residuals. The second measures whether blocks with larger logit variance also have systematically different mean values. Both are quadratic under the stated assumptions.

A second-order output bound also holds under a weaker condition on the values. Suppose $R_V=O(1)$ and $\rho_{iB}=O(\eta)$, and assume $\|\Cov_B(\epsilon_{ij},\vv_j)\|_\infty=O(\eta^2)$ uniformly. The remainder after the linear term in $\vd_{iB}$ is at most $R_Ve^{\rho_{iB}}\sigma_{z,iB}^2/2=O(\eta^2)$. Hence $\vd_{iB}=O(\eta^2)$ and $f_{iB}-1=O(\eta^2)$, so Eq.~\eqref{eq:weighted_output_identity} again gives $\|\vy_i-\widehat{\vy}_i\|_\infty=O(\eta^2)$. This conclusion controls the output order; the more detailed expansion in Eq.~\eqref{eq:second_order_expansion} uses the additional within-block value regularity stated above.

\subsection{Mass in logarithmic distance shells}
\label{app:radial-mass}

The approximation bounds describe the error introduced by pooling. To understand how much attention is represented at each spatial scale, fix a query in Eq.~\eqref{eq:log-distance} and set the content term to zero. We use a continuous approximation to the token grid, with constant density $\rho_0>0$, and consider complete annuli contained in the spatial domain at positive radius $r$. Let $M_i$ denote the unnormalized attention mass on radial intervals. The radial logit assigns unnormalized weight $e^{b_i}(r/r_*)^{-\alpha_i}$ to a location at distance $r$. Multiplying by the annulus area $2\pi r\,\diff r$ and the token density gives its unnormalized mass:
\begin{equation}
    \diff M_i(r)=2\pi\rho_0e^{b_i}r_*^{\alpha_i}r^{1-\alpha_i}\,\diff r\propto r^{2-\alpha_i}\,\diff\log r.
    \label{eq:annulus-mass}
\end{equation}
The last expression uses $\diff r=r\,\diff\log r$ and describes mass per logarithmic distance interval. In particular, let $M_i([r,2r])$ be the mass in the annulus between radii $r$ and $2r$. Integrating the radial density yields
\begin{equation}
    M_i([r,2r])=2\pi\rho_0e^{b_i}r_*^{\alpha_i}
    \begin{cases}
        \displaystyle r^{2-\alpha_i}\frac{2^{2-\alpha_i}-1}{2-\alpha_i}, & \alpha_i\ne2, \\[6pt]
        \log 2, & \alpha_i=2.
    \end{cases}
    \label{eq:doubling-shell-mass}
\end{equation}
At $\alpha_i=2$, the increase in spatial population exactly balances the decay of individual weights, giving equal mass to successive distance doublings. Larger shells carry more mass when $\alpha_i<2$ and less when $\alpha_i>2$. This continuous calculation explains the population factor that SFA retains when it represents a region by one pooled key and value. On a finite token grid, that factor is the actual discrete count $m_B$. The $\log m_B$ correction restores this count after pooling, leaving the block's mean logit to determine its content-dependent relevance.

\subsection{Mean pooling and conditional noise variance}
\label{app:noise_variance}

Once the probability mass assigned to each block is fixed, its allocation among the constituent values determines how much noise reaches the output. For one value coordinate, model the observation at token $j$ as $\rz_j=s_j+\rn_j$, where $s_j$ is a deterministic signal and the noises $\rn_j$ are independent, centered, and have common variance $\sigma^2>0$. Take a probability vector $\vw=(w_j)_{j=1}^N$ that is fixed or independent of these noises; in the latter case, all calculations below condition on the weights. Prescribe block masses $\pi_B\ge0$ with $\sum_B\pi_B=1$ and require $\sum_{j\in B}w_j=\pi_B$.

Write the deviation from uniform weighting as $d_j=w_j-\pi_B/m_B$. Since $\sum_{j\in B}d_j=0$, expanding the square gives $\sum_{j\in B}w_j^2=\pi_B^2/m_B+\sum_{j\in B}d_j^2$. Independence of the noises gives $\Var(\sum_jw_j\rn_j)=\sigma^2\sum_jw_j^2$, so summing the blockwise identity yields
\begin{equation}
    \Var\!\left(\sum_jw_j\rn_j\right)
    =\sigma^2\sum_B\frac{\pi_B^2}{m_B}
    +\sigma^2\sum_B\sum_{j\in B}d_j^2.
    \label{eq:noise_variance_decomposition}
\end{equation}
For the prescribed masses, the first term is fixed and the second vanishes exactly at $w_j=\pi_B/m_B$. Uniform weights therefore uniquely minimize the conditional noise variance. The same calculation applies to independent centered vector noises $\rvn_j\in\mathbb R^{d_v}$ with common covariance matrix $\mSigma=\E[\rvn_j\rvn_j^\top]$. Their output covariance is $\mSigma\sum_jw_j^2$, which uniform weighting minimizes in the positive-semidefinite order.

Uniformization also changes the signal estimate. To quantify this tradeoff at the same block masses, define the target $y^*=\sum_jw_js_j$ and uniformized weights $w_j^{\mathrm{U}}=\pi_B/m_B$ for $j\in B$. The estimators $\rz_w=\sum_jw_j\rz_j$ and $\rz_{\mathrm{U}}=\sum_jw_j^{\mathrm{U}}\rz_j$ have mean squared errors (MSE) $\operatorname{MSE}_w=\E(\rz_w-y^*)^2$ and $\operatorname{MSE}_{\mathrm{U}}=\E(\rz_{\mathrm{U}}-y^*)^2$. The first estimator is unbiased for this target, while the second has bias $b_{\mathrm{U}}=\sum_j(w_j^{\mathrm{U}}-w_j)s_j$. Decomposing MSE into squared bias and variance gives
\begin{equation}
    \operatorname{MSE}_{\mathrm{U}}-\operatorname{MSE}_w
    =b_{\mathrm{U}}^2-\sigma^2\sum_B\sum_{j\in B}(w_j-\pi_B/m_B)^2.
    \label{eq:noise_bias_variance}
\end{equation}
The variance reduction improves this signal estimate when it exceeds the squared bias. Spatially coherent signals make uniformization useful because their small within-block variation keeps that bias small.

Full attention and SFA can assign different masses to a block, so their comparison also includes the mass adjustment. Substituting the full-attention weights $a_{ij}$ and the block-uniform SFA weights $\widehat a_{ij}$ into the variance decomposition gives the conditional noise variance of SFA minus that of full attention:
\begin{equation}
    \sigma^2\left[
        \sum_B\frac{\widehat p_{iB}^{\,2}-p_{iB}^2}{m_B}
        -\sum_B\sum_{j\in B}\left(a_{ij}-\frac{p_{iB}}{m_B}\right)^2
    \right].
    \label{eq:sfa_noise_difference}
\end{equation}
The first term depends on the block-probability changes controlled by Eq.~\eqref{eq:scale_probability}; the second is the variance reduction obtained by making the weights uniform within each block. Under the stated conditional noise model, these identities isolate how signal coherence, within-block averaging, and mass changes interact. The generation experiments assess their net effect after the attention weights and features are learned. This provides a multiscale averaging mechanism alongside the geometry-adaptive harmonic shrinkage observed in learned diffusion denoisers \citep{kadkhodaie2024generalization}.

\subsection{Rotary positional encoding and the frequency response of pooling}
\label{app:rope_pooling}

On the key side, SFA averages features after positional encoding. Rotary position embeddings (RoPE) rotate queries and keys according to position~\citep{su2024roformer}, and linearity of the dot product ensures that averaging the rotated keys preserves the mean of the actual attention logits. To describe the spatial filtering induced by this operation, represent a pair of real key channels at position $\vx\in\mathbb R^2$ by the complex scalar $\chi(\vx)=e^{\mathrm{i}\vomega^\top\vx}\chi^0(\vx)$. Here $\vomega=(\omega_x,\omega_y)^\top$ is the rotary frequency vector, $\chi^0$ is the unrotated content, and $\mathrm{i}^2=-1$.

Consider an axis-aligned rectangular block centered at $\vc_B\in\mathbb R^2$, with $n_x,n_y$ unit-spaced positions along its axes. For constant content $\chi^0(\vx)=\chi_0\in\mathbb C$, its mean $\bar\chi_B$ is determined entirely by the average rotary phase:
\begin{equation}
    \bar\chi_B=e^{\mathrm{i}\vomega^\top\vc_B}H_B(\vomega)\chi_0,
    \qquad
    H_B(\vomega)=
    \frac{\sin(n_x\omega_x/2)}{n_x\sin(\omega_x/2)}
    \frac{\sin(n_y\omega_y/2)}{n_y\sin(\omega_y/2)}.
    \label{eq:rope_pool_frequency}
\end{equation}
To obtain this expression, consider one axis with $n$ positions $t=0,\ldots,n-1$, center $c=(n-1)/2$, and scalar frequency $\omega$. The geometric-series identity gives $n^{-1}\sum_{t=0}^{n-1}e^{\mathrm{i}\omega t}=e^{\mathrm{i}\omega c}\sin(n\omega/2)/(n\sin(\omega/2))$. Multiplying the responses along the two axes and translating to the block center gives the displayed product and its center phase. At denominator zeros, the expressions are defined by continuity.

For spatially varying content, apply the same calculation to a Fourier component $\chi^0(\vx)=\chi_0e^{\mathrm{i}\bm{\nu}^{\top}\vx}$, where $\bm{\nu}\in\mathbb R^2$ is its content frequency and $\chi_0$ its complex amplitude. Multiplication by the rotary phase adds the frequency vectors before averaging, giving
\begin{equation}
    \bar\chi_B=\chi_0
    e^{\mathrm{i}(\bm{\nu}+\vomega)^\top\vc_B}
    H_B(\bm{\nu}+\vomega).
    \label{eq:modulated-content-response}
\end{equation}
Superposition extends this response to a general key field. The window's magnitude response has its main lobe at zero frequency in the rotated representation, so its main lobe for the unrotated content is centered at $-\vomega$, with frequencies understood modulo $2\pi$ along each unit-grid axis. Its widths decrease in proportion to $1/n_x$ and $1/n_y$. Thus larger blocks apply narrower spatial filtering to the position-modulated keys. Values undergo the same averaging window without rotary modulation.

Distinct uses of high and low rotary frequencies have also been observed in trained language models \citep{barbero2025rope}. In SFA, the additional averaging window makes the retained frequency resolution depend on block size and hence on distance. The approximation bounds account for this operation through the residuals of the rotated keys, while the exact local neighborhood retains individual token interactions.

\subsection{Two-dimensional logarithmic geometry}
\label{app:logarithmic_geometry}

We next examine the effect of region shape using continuous spatial averages of the logarithmic radial model. Place the query at the origin and suppress its index, writing $\alpha=\alpha_i$ and $b=b_i$. At position $\vx$ with radius $r=\|\vx\|_2>0$, the logit is $z(\vx)=b-\alpha\ln(r/r_*)$. Since $\nabla^2\ln r=-r^{-2}+r^{-2}=0$ away from the query, $z$ is harmonic. Its mean over a disk excluding the query therefore equals its center value. For a square, isotropic second moments make the average quadratic Taylor term proportional to the Laplacian, so that term also vanishes; symmetry leaves the fourth-order term as the first possible discrepancy.

To compute this term, take a continuous square $B$ of side $s>0$ centered at $\vc_B$. Represent its center by the complex number $R_B=re^{\mathrm{i}\theta}$, where $r=\|\vc_B\|_2$ and $\theta$ is the polar angle. Independent offsets $\rx,\ry\sim\operatorname{Uniform}[-s/2,s/2]$ define $\ru=\rx+\mathrm{i}\ry$, and the block mean is $\bar z_B=\E[z(\vc_B+(\rx,\ry)^\top)]$. The condition $s/(\sqrt2r)<1$ places every offset within the convergence radius of
\begin{equation}
    \ln|R_B+\ru|=\ln r+\operatorname{Re}\sum_{n\ge1}
    \frac{(-1)^{n+1}}{n}\left(\frac{\ru}{R_B}\right)^n,
    \label{eq:logarithm_complex_series}
\end{equation}
where $\operatorname{Re}$ takes the real part and $n$ is a positive integer. Fourfold rotational symmetry gives $\E\ru^n=0$ unless $n$ is divisible by four, and direct evaluation gives $\E\ru^4=2s^4/80-6(s^2/12)^2=-s^4/60$. Substituting this moment into the logarithm series and multiplying by $-\alpha$ yields, as $s/r\to0$,
\begin{equation}
    \bar z_B-z(\vc_B)=-\frac{\alpha s^4}{240r^4}\cos(4\theta)
    +O_\alpha\!\left((s/r)^8\right).
    \label{eq:square_harmonic_average}
\end{equation}
The remainder begins at the next power allowed by the same rotational symmetry, and $O_\alpha$ allows its constant to depend on the fixed exponent $\alpha$. For a centered regular square grid, interpret the mean as a uniform average over its discrete points. The same cancellation holds, with the coefficient of $\cos(4\theta)$ replaced by $\alpha(2\mu_4-6\mu_2^2)/(4r^4)$, where $\mu_2$ and $\mu_4$ are the second and fourth moments of a coordinate offset from the block center.

Attention mass also depends on exponentiating the variation around the mean logit. Returning to the continuous square, define the exponential correction $J_B=\ln\E[\exp(z(\vc_B+(\rx,\ry)^\top)-\bar z_B)]$. Its leading term is
\begin{equation}
    J_B=\frac{\alpha^2s^2}{24r^2}+O_\alpha\!\left((s/r)^4\right).
    \label{eq:square_mass_correction}
\end{equation}
To see the source of this term, let $\nabla z$ denote the spatial gradient. The linear Taylor term has variance $(s^2/12)\|\nabla z(\vc_B)\|_2^2=\alpha^2s^2/(12r^2)$. With the block center and $\alpha$ fixed, rescaling the offsets to a fixed unit square makes the centered logit residual analytic in $t=s/r$ near zero. Expanding its exponential and logarithm gives a quadratic term in $J_B$ equal to half the variance of the linear Taylor term. Simultaneous reflection of both offsets preserves their joint distribution and is equivalent to replacing $t$ by $-t$, so $J_B$ is analytic and even in $t$. Its cubic term therefore vanishes, giving the $O_\alpha(t^4)$ remainder in Eq.~\eqref{eq:square_mass_correction}. Thus Eq.~\eqref{eq:square_harmonic_average} describes a fourth-order mean-logit discrepancy from the center, while Eq.~\eqref{eq:square_mass_correction} describes the second-order mass correction generated by within-block variation.

These local calculations are consistent across geometrically scaled blocks. To state the scaling law, let $D$ be the spatial dimension, with $D=2$ for images, and use the same radial logit at unit spatial density. For a bounded block $B\subset\mathbb R^D$ of positive volume at positive distance from the query, let $|B|$ denote its volume. Define $\bar z_B=|B|^{-1}\int_Bz(\vx)\,\diff\vx$, the exact mass $W_B=\int_Be^{z(\vx)}\,\diff\vx$, and the pooled mass $\widehat W_B=|B|e^{\bar z_B}$, so $J_B=\ln(W_B/\widehat W_B)$. For a dilation $B\mapsto\lambda B$ about the query with $\lambda>0$,
\begin{equation}
    W_{\lambda B}=\lambda^{D-\alpha}W_B,
    \qquad
    \widehat W_{\lambda B}=\lambda^{D-\alpha}\widehat W_B,
    \qquad J_{\lambda B}=J_B.
    \label{eq:powerlaw_homogeneity}
\end{equation}
The change of variables in the integral contributes a volume factor $\lambda^D$, while the radial weight contributes $\lambda^{-\alpha}$, proving the first equality. The second follows from $|\lambda B|=\lambda^D|B|$ and $\bar z_{\lambda B}=\bar z_B-\alpha\ln\lambda$. Blocks related by these query-centered dilations therefore share a common relative mass correction, which cancels when normalization is restricted to such a family. For a fixed discrete collection of token positions, dilation changes distances while leaving the count unchanged, so the exact and pooled sums both scale as $\lambda^{-\alpha}$. Their relative correction is again invariant. The continuous and discrete statements describe the same radial homogeneity with their respective spatial measures.

\section{Details of the Triton Kernel Implementation}
\label{app:implementation}

This section follows the computation of SFA and its gradients. After introducing the notation in Appendix~\ref{app:interaction-lists}, we describe the forward pass in Appendix~\ref{app:kernel-forward}: constructing block means, sharing their loads across queries, and accumulating one softmax across scales. Appendix~\ref{app:kernel-backward} follows the backward pass from probability recomputation to the final query, key, and value gradients. At each step, we explain how the hierarchy enables data reuse and reduces memory traffic. The description covers the default Triton implementation~\citep{tillet2019triton}, with fixed mean pooling and token-count correction.

\subsection{Notation}
\label{app:interaction-lists}

Let $n_b$ and $n_h$ denote batch size and head count. The query, key, value, and output tensors $\tQ,\tK,\tV,\tY$ have shape $[n_b,n_h,N,d]$, with $N=L^2$ and $d_v=d$. At a fixed batch and head, their token-$i$ slices are $\vq_i$, $\vk_i$, $\vv_i$, and $\widehat{\vy}_i$, respectively; batch and head indices are suppressed below. The supplied kernels support power-of-two grid sides $8\le L\le512$ and head dimensions $d\in\{64,72\}$.

A level-$0$ base block contains $4\times4=16$ tokens. There are $G=N/16$ base blocks and $n_{\mathrm{lev}}=\log_2L-3$ pooled levels, indexed by $\ell=0,\ldots,n_{\mathrm{lev}}-1$. A block at level $\ell$ contains $m_\ell=16\cdot4^\ell$ tokens. The full quadtree extends to the root at level $n_{\mathrm{lev}}+1$; let $B_\ell(i)$ be its level-$\ell$ block containing query $i$. Let $\mathcal N(B)$ be the set of same-level blocks in the $3\times3$ neighborhood of $B$, clipped to the image boundary. The exact tokens form the union of the blocks in $\mathcal N(B_0(i))$; the pooled interaction list is
\begin{equation}
    \mathcal I_\ell(i)
    =\operatorname{children}\!\left(\mathcal N(B_{\ell+1}(i))\right)
    \setminus\mathcal N(B_\ell(i)),
    \label{eq:interaction-list}
\end{equation}
where $\operatorname{children}$ maps a block to its immediate quadtree children and a set of blocks to the union of their child sets. This gives at most 144 exact tokens and 27 pooled representatives per level, with the complete partition described in Section~\ref{sec:partition}. The geometry is shared across batches and heads and cached for each grid size. At $L=8$, the exact neighborhood already covers the image.

The partition follows from the nesting of the neighborhood regions. For $\ell=0,\ldots,n_{\mathrm{lev}}$, define the token set $\mathcal R_\ell(i)=\bigcup_{B\in\mathcal N(B_\ell(i))}B$. For $\ell<n_{\mathrm{lev}}$, the parent of every block in $\mathcal N(B_\ell(i))$ lies in $\mathcal N(B_{\ell+1}(i))$, so $\mathcal R_\ell(i)\subseteq\mathcal R_{\ell+1}(i)$. Since the children partition each parent, Eq.~\eqref{eq:interaction-list} gives
\begin{equation*}
    \bigcup_{B\in\mathcal I_\ell(i)}B
    =\mathcal R_{\ell+1}(i)\setminus\mathcal R_\ell(i).
\end{equation*}
These differences are disjoint across levels, and blocks within each level are disjoint. Together with the exact region $\mathcal R_0(i)$, their union is $\mathcal R_{n_{\mathrm{lev}}}(i)$. At level $n_{\mathrm{lev}}$, the grid consists of $2\times2$ blocks, so every clipped $3\times3$ neighborhood covers all $N$ tokens. The exact singleton tokens and pooled blocks therefore form a complete, disjoint partition, including at image boundaries.

\Needspace{4\baselineskip}
\subsection{Forward pass}
\label{app:kernel-forward}

\begin{figure}[!t]
    \centering
    \includegraphics[width=\linewidth]{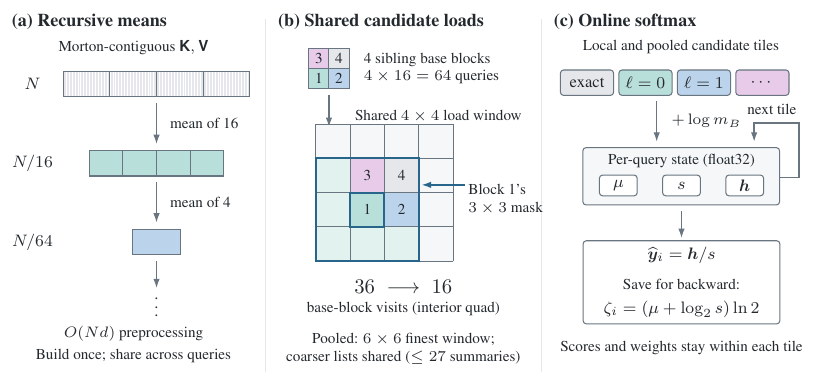}
    \caption{Forward computation of SFA (\textbf{ours}), with batch and head indices suppressed. \textbf{(a)} Morton-contiguous keys and values form a shared hierarchy: the first reduction averages 16 tokens, and each subsequent reduction averages four child summaries. \textbf{(b)} One program processes four sibling base blocks, or 64 queries. For an interior quad, their exact neighborhoods cover 16 distinct base blocks, compared with 36 block visits when processed separately; per-query masks preserve the original neighborhoods. \textbf{(c)} Each query maintains its own running maximum $\mu$, exponential sum $s$, and unnormalized output $\vh$, shared across exact tokens and all pooled scales. The final state gives $\widehat{\vy}_i=\vh/s$ and the log-sum-exp statistic $\zeta_i$ saved for backward.}
    \label{fig:kernel-forward}
\end{figure}

Figure~\ref{fig:kernel-forward} shows how the mean hierarchy and shared candidate loads feed one online softmax across spatial scales.

\paragraph{Construct shared block means.}
Morton order interleaves the bits of the two spatial coordinates, placing every aligned quadtree block in a contiguous token interval. After positional encoding, the first pooling step averages each group of 16 keys and values. Every coarser block then averages its four child summaries:
\begin{equation}
    \bar{\vk}_B=\frac14\sum_{C\in\operatorname{children}(B)}\bar{\vk}_C,
    \qquad
    \bar{\vv}_B=\frac14\sum_{C\in\operatorname{children}(B)}\bar{\vv}_C.
    \label{eq:recursive-means}
\end{equation}
The implementation uses contiguous reshaping and mean reductions, retaining the input data type. The number of summaries decreases geometrically, with $\sum_{\ell=0}^{n_{\mathrm{lev}}-1}N/(16\cdot4^\ell)<N/12$ summary vectors per key or value hierarchy. Construction therefore costs $O(Nd)$ per head. Each summary is computed once and reused by all queries that access it.

\paragraph{Share candidate loads across nearby queries.}
Each forward GPU program loads 64 queries from four sibling base blocks, called a quad. Their exact neighborhoods overlap: the union is a $4\times4$ window of base blocks. The program reads this window in 32-token tiles and masks each query back to its own $3\times3$ neighborhood. For an interior quad, this loads 16 base blocks instead of the $4\cdot9=36$ visits made by processing the four query blocks separately. The shared tiles retain regular matrix products while reducing repeated key and value loads.

The pooled traversal uses the same principle. At $\ell=0$, the quad shares a $6\times6$ candidate window, visited in two 32-column tiles, and each query masks out its exact neighborhood. At $\ell\ge1$, all four siblings have the same interaction list, so one list of at most 27 summaries supplies all 64 queries. The exact and finest pooled windows are generated from coordinates; coarser lists are read from the cached indices. These masks preserve each query's partition while avoiding a separately gathered candidate tensor for every query.

\paragraph{Accumulate one softmax across all scales.}
For a representative $B$, including singletons, define $\lambda_{iB}=\vq_i^\top\bar{\vk}_B/\sqrt d+\log m_B$ and its base-two equivalent $\beta_{iB}=\lambda_{iB}\log_2e$. At pooled level $\ell$, the count correction is simply $\log_2m_B=4+2\ell$. Following online tiled attention~\citep{dao2022flashattention,dao2023flashattention2}, each query maintains a running maximum $\mu$, exponential sum $s$, and unnormalized output vector $\vh\in\mathbb R^d$. A candidate tile $\mathcal T$ updates them as
\begin{align}
    \mu'&=\max\!\left(\mu,\max_{B\in\mathcal T}\beta_{iB}\right),\\
    s'&=2^{\mu-\mu'}s+\sum_{B\in\mathcal T}2^{\beta_{iB}-\mu'},\\
    \vh'&=2^{\mu-\mu'}\vh+\sum_{B\in\mathcal T}2^{\beta_{iB}-\mu'}\bar{\vv}_B.
    \label{eq:online-softmax}
\end{align}
The initial state is $(\mu,s,\vh)=(-\infty,0,\vzero)$, and empty masked tiles leave it unchanged. The state persists through the exact region and every pooled level, producing $\widehat{\vy}_i=\vh/s$. For training, the kernel also saves the natural-log log-sum-exp (LSE) statistic $\zeta_i=(\mu+\log_2s)\ln2=\ln\widehat Z_i$. Scores and probabilities are consumed within each tile, so the shared normalization requires no stored attention matrix.

Matrix operands use the input bfloat16 or float16 format, with float32 accumulation and normalization. In particular, probabilities are converted to the input format for the probability--value product. For $d=72$, matrix tiles use 128 feature lanes with zero padding, while stored tensors and dot-product scaling retain dimension 72. These choices combine tensor-core matrix operations with compact storage. The training wrapper retains the mean hierarchy and LSE values, together with references to the input and output tensors.

\subsection{Backward pass}
\label{app:kernel-backward}

\begin{figure}[!t]
    \centering
    \includegraphics[width=\linewidth]{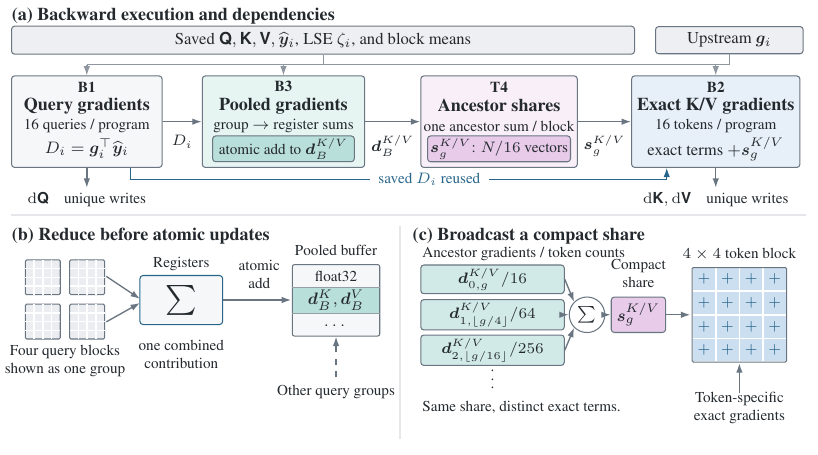}
    \caption{Backward computation of SFA (\textbf{ours}), with batch and head indices suppressed; $\diff\tQ,\diff\tK,\diff\tV$ denote input gradients, and superscript $K/V$ denotes either key or value quantities. The quantity $\vd_{\ell,t}^{K/V}$ is the pooled gradient of block $t$ at level $\ell$. \textbf{(a)} B1 computes query gradients and the statistics $D_i$; B3 accumulates pooled gradients; T4 combines ancestor contributions into base-block shares; and B2 adds these shares to the exact gradients during the final write. \textbf{(b)} Query-block contributions are summed in registers before atomic updates to shared pooled-gradient buffers. \textbf{(c)} For base block $g$, ancestor gradients weighted by reciprocal token counts give per-token shares $\vs_g^K,\vs_g^V$. These shares occupy $G=N/16$ rows per batch and head and are broadcast to the 16 tokens of each base block.}
    \label{fig:kernel-backward}
\end{figure}

The backward pass recomputes candidate scores using the saved inputs and means. Let $\mathcal L$ be the loss, $\vg_i=\partial\mathcal L/\partial\widehat{\vy}_i$, and $D_i=\vg_i^\top\widehat{\vy}_i$. The saved normalizer recovers each probability, which gives the loss derivative with respect to the natural-log score:
\begin{equation}
    \widehat p_{iB}=\exp(\lambda_{iB}-\zeta_i),
    \qquad
    \gamma_{iB}:=\frac{\partial\mathcal L}{\partial\lambda_{iB}}
    =\widehat p_{iB}\bigl(\vg_i^\top\bar{\vv}_B-D_i\bigr).
    \label{eq:representative-gradient}
\end{equation}
Treating the stored representatives as independent inputs to the attention computation, let $\vd_B^K,\vd_B^V\in\mathbb R^d$ denote their direct key and value gradients. The required accumulations are
\begin{equation}
    \begin{aligned}
        \frac{\partial\mathcal L}{\partial\vq_i}
        &=\frac1{\sqrt d}\sum_{B\in\mathcal P_i}\gamma_{iB}\bar{\vk}_B,\\
        \vd_B^K&=\frac1{\sqrt d}\sum_{i:B\in\mathcal P_i}\gamma_{iB}\vq_i,
        \qquad
        \vd_B^V=\sum_{i:B\in\mathcal P_i}\widehat p_{iB}\vg_i.
    \end{aligned}
    \label{eq:kernel-gradient-accumulation}
\end{equation}
The pooled gradients are subsequently propagated through the means. The fixed count bias requires no gradient; its effect is already included in $\widehat p_{iB}$. Figure~\ref{fig:kernel-backward} organizes the following four steps and shows how shared summaries reduce gradient traffic.

\paragraph{Compute query gradients (B1).}
One program owns the 16 queries of a base block. It computes and saves $D_i$, revisits the exact and pooled candidates, and accumulates each query gradient in float32 before storing it. Each query has a unique writer, while the saved $D_i$ values are reused by both key and value gradient computations.

\paragraph{Combine gradients of shared summaries (B3).}
Several query programs can update the same pooled representative. Each program first combines the contributions of a group of query blocks. At $\ell=0$, one program processes four sibling query blocks against a shared set of at most 36 candidates, retaining the per-query exclusion masks. At $\ell\ge1$, identical interaction lists allow one program to combine $c_\ell=\min(4^\ell,32)$ base query blocks. It loads the pooled keys and values once, processes 32 query rows at a time, and accumulates their contributions in registers. Atomic additions then write the combined gradients to initially zero float32 buffers. Relative to separate base-block programs, grouping reduces shared-candidate loads and gradient updates by a factor of $c_\ell$ at these coarser levels.

\paragraph{Propagate through the means (T4).}
All 16 tokens in a base block receive the same gradient share from every pooled ancestor. Let $g\in\{0,\ldots,G-1\}$ be its Morton index, and let $\vd_{\ell,t}^K,\vd_{\ell,t}^V$ denote the direct gradients of level-$\ell$ block $t$. Its ancestor index is $\lfloor g/4^\ell\rfloor$, so the per-token pooled contributions are
\begin{align}
    \vs_g^K&=\sum_{\ell=0}^{n_{\mathrm{lev}}-1}
    \frac{\vd_{\ell,\lfloor g/4^\ell\rfloor}^K}{16\cdot4^\ell},\\
    \vs_g^V&=\sum_{\ell=0}^{n_{\mathrm{lev}}-1}
    \frac{\vd_{\ell,\lfloor g/4^\ell\rfloor}^V}{16\cdot4^\ell}.
    \label{eq:mean-gradient}
\end{align}
The denominator is the number of original tokens in the corresponding mean. A single kernel sums these ancestor contributions for both keys and values, accumulating in float32 and storing one vector per base block in the input format. This completes the chain rule through the hierarchy in compact arrays of shape $[n_b,n_h,G,d]$.

\paragraph{Write the original key and value gradients (B2).}
Finally, one program owns the 16 original keys and values in each base block. Symmetry of the exact neighborhood identifies all queries that attend to those tokens directly. The program accumulates these exact contributions, adds $\vs_g^K$ and $\vs_g^V$ to each token's corresponding gradients, and performs the final stores. Thus the full-resolution gradients have unique writers, and the pooled contribution needs neither a full-resolution intermediate buffer nor a separate addition pass.

The saved mean hierarchy, pooled-gradient buffers, and base-block shares all contain $O(n_bn_hNd)$ floating-point entries. Integer geometry is cached separately and shared across batches and heads.

\section{Experimental Details and Additional Results}
\label{app:experiment-details}

\FloatBarrier

\subsection{Hardware and evaluation settings}

All throughput measurements use one NVIDIA RTX 5090 GPU, with the attention-kernel benchmark allocated 32 CPU cores and 64\,GB of host memory. The software stack comprises PyTorch 2.12.0+cu132, CUDA 13.2 (\texttt{nvcc} 13.2.78), and NVIDIA driver 595.71.05. Kernel inputs use bfloat16 precision, and training-throughput measurements for PixelFlow and FFHQ enable gradient checkpointing. Every reported training time covers one complete optimizer step, and every generation rate measures the sampling pipeline end to end. ImageNet training, Stable Diffusion 3.5 Medium fine-tuning, and every generation-quality evaluation run on eight NVIDIA A800 GPUs with PyTorch 2.13.0+cu129 and NVIDIA driver 535.261.03, whereas the FFHQ models are trained on the same RTX 5090.

Alongside the timing measurements, we evaluate PixelFlow quality with the evaluation suite released with the ablated diffusion model (ADM; \citealp{dhariwal2021adm}), using 50,000 generated samples and Euler sampling to compute FID~\citep{heusel2017fid} and IS~\citep{salimans2016improved}. In this comparison, efficient-attention operators replace the 4096-token stage, leaving full attention in the other three stages. Table~\ref{tab:training-settings} collects the PixelFlow and FFHQ training and sampling settings, and Appendix~\ref{app:powerlaw-details} specifies the PixelFlow-XL architecture used for the separate radial-attention study. The five-method ImageNet comparison uses 40 training epochs; full attention and SFA then continue through epoch 80 to examine whether their quality difference persists beyond this common budget.

\begin{table}[!t]
    \centering
    \caption{Training and sampling settings for PixelFlow and FFHQ, including exponential moving average (EMA) decay. PixelFlow expands its guidance scale across the four stages, so the listed value applies to the highest-resolution stage. The FFHQ models are unconditional and therefore use no guidance. The Stable Diffusion 3.5 Medium protocol is described in Appendix~\ref{app:sd35}.}
    \label{tab:training-settings}
    \setlength{\tabcolsep}{5pt}
    \newcolumntype{Y}{>{\raggedright\arraybackslash}X}
    \begin{tabularx}{\linewidth}{lYYY}
        \toprule
        Setting & PixelFlow\newline from scratch & PixelFlow\newline fine-tuning & FFHQ \\
        \midrule
        Dataset & ImageNet & ImageNet & FFHQ \\
        Image resolution & $256\times256$ & $256\times256$ & 128, 256, 512 \\
        Model variant & PixelFlow-XL, 677M & PixelFlow-XL, 677M & DiT-S, 35M \\
        Replaced stage & 4096 tokens & 4096 tokens & All attention layers \\
        Initialization & Random & Released checkpoint & Table~\ref{tab:ffhq-schedule} \\
        Training hardware & 8 NVIDIA A800 & 8 NVIDIA A800 & 1 NVIDIA\newline RTX 5090 \\
        Global batch size & 64 & 64 & 16, 4, 1 \\
        Training epochs & 40; 80 for Full/SFA (\textbf{ours}) & 5 scheduled & 20 at 128/256;\newline 2 at 512 \\
        Evaluated epochs & Every 5 & 3.5 & Final \\
        Optimizer & \multicolumn{3}{l}{AdamW, $\beta=(0.9,0.999)$, $\epsilon=10^{-8}$, gradient-norm clipping 1.0} \\
        Learning rate & $10^{-4}$ & $10^{-5}\rightarrow10^{-6}$ & $10^{-4}$ \\
        Learning-rate schedule & Constant & 2 epochs constant, 3 epochs cosine & Constant \\
        Weight decay & 0 & 0 & $10^{-2}$ \\
        EMA decay & 0.9999 & 0.9999, scaled with the learning rate & 0.9999 with warmup \\
        Training precision & bfloat16 & bfloat16 & bfloat16 \\
        Sampling method & Euler & Euler & Euler \\
        Sampling steps & 30 per stage & 30 per stage & 20 \\
        Guidance scale & 2.4 & 2.4 & None \\
        Evaluation sample count & 50,000 & 50,000 & 50,000 \\
        \bottomrule
    \end{tabularx}
\end{table}

The FFHQ models follow the resolution cascade of LLSA~\citep{zhou2026llsa}. A full-attention DiT-S is trained at $32\times32$ from random initialization, and each higher-resolution stage continues from the preceding stage's checkpoint, with the training noise rescaled as the token grid grows. The four efficient-attention operators replace full attention from $128\times128$ onward, starting from the same $32\times32$ model and training for 20 epochs at each of the resolutions 128 and 256. At $512\times512$, we compare SFA and LLSA, each continuing for two epochs from its own $256\times256$ checkpoint. Table~\ref{tab:ffhq-schedule} lists this schedule.

\begin{table}[!t]
    \centering
    \caption{Resolution cascade used for the FFHQ models, where one epoch covers the 70,000 FFHQ images. Each stage continues from the checkpoint of the preceding stage, and the compared attention operator replaces full attention from $128\times128$ onward. The rescaling factor multiplies the noise sampled during training at that resolution.}
    \label{tab:ffhq-schedule}
    \setlength{\tabcolsep}{6pt}
    \begin{tabular}{lcccc}
        \toprule
        Stage resolution & $32\times32$ & $128\times128$ & $256\times256$ & $512\times512$ \\
        \midrule
        Attention & Full & \multicolumn{2}{c}{SFA (\textbf{ours}), LLSA, SLA, VSA} & SFA (\textbf{ours}), LLSA \\
        Initialized from & Random & $32\times32$ & $128\times128$ & $256\times256$ \\
        Training epochs & 40 & 20 & 20 & 2 \\
        Image batch size & 64 & 16 & 4 & 1 \\
        Optimizer steps & 43,760 & 87,500 & 350,000 & 140,000 \\
        Noise rescaling & 1 & 2 & 4 & 8 \\
        \bottomrule
    \end{tabular}
\end{table}

\subsection{Baseline configurations and kernel performance}
\label{app:kernel-performance}

We use LLSA as the reference for the selected-block budget, with its two-level configuration, block size 16, and selection counts $k_1=k_2=8$ at the two hierarchy levels. SLA uses query and key/value blocks of 64 tokens and the softmax feature map in its linear branch, and VSA uses block shape $(1,8,8)$, whose unit temporal dimension corresponds to image inputs. Accounting for these different block sizes, the budget-matched SLA and VSA kernel settings select 4, 5, 8, and 20 key/value blocks for spatial grid sides $L=64,128,256,512$, respectively, matching LLSA up to block rounding.

At $L=64$, we round the selected-block budget down to four blocks in the isolated kernel benchmark and up to five blocks in generation-quality experiments. The higher sparsity in the kernel benchmark favors SLA and VSA's throughput, and the denser attention in generation experiments favors their quality, making both comparisons more demanding for SFA. Model-level throughput uses the same five-block configuration as generation quality. At $L=128,256,512$, rounding up and down yields identical counts of 5, 8, and 20 blocks, respectively, so the throughput and generation-quality settings coincide.

In addition to the budget-matched configurations, we evaluate fixed-rate alternatives that keep 5\% of the key blocks for SLA and 10\% for VSA at every grid size. We reuse the Triton implementations released by the authors of each baseline wherever possible. Those kernels require a power-of-two head dimension, so at the head dimension 72 of PixelFlow we zero-pad their features to 128 and keep the original softmax scale, either by rescaling the query (LLSA, VSA) or by passing the scale to the kernel (SLA), which leaves the attention they compute unchanged. SFA handles head dimension $72$ inside its kernel, using similar methods. SFA and LLSA add no parameters to the backbone, whereas SLA and VSA each introduce a small module in every replaced layer. FMA has not yet been open-sourced, so we examine several key design differences between SFA and FMA primarily through the ablation experiments in Appendix~\ref{app:ablation}. With these configurations, Table~\ref{tab:kernel-full} provides the numerical results underlying Figure~\ref{fig:overview-results}(A,B) and adds the fixed-rate SLA and VSA comparisons at all four grid sizes.

\begin{table}[!htbp]
    \centering
    \caption{Complete kernel benchmark, expressed as speedup over the FlashAttention backend of PyTorch scaled dot-product attention (SDPA) for forward and backward computation separately. Inputs have shape $[n_b,n_h,N,d]$, with batch size $n_b=4$, head count $n_h=6$, query/key dimension $d=64$, value dimension $d_v=d$, and $N=L^2$ tokens on an $L\times L$ grid, using bfloat16 precision. Full attention is the unit reference, and bold marks the best speedup.}
    \label{tab:kernel-full}
    \setlength{\tabcolsep}{3.6pt}
    \begin{tabular}{lrrrrrrrr}
        \toprule
        & \multicolumn{4}{c}{Forward speedup ($\times$)} & \multicolumn{4}{c}{Backward speedup ($\times$)} \\
        \cmidrule(lr){2-5}\cmidrule(lr){6-9}
        Grid side $L$ & 64 & 128 & 256 & 512 & 64 & 128 & 256 & 512 \\
        \midrule
        Full & 1.00 & 1.00 & 1.00 & 1.00 & 1.00 & 1.00 & 1.00 & 1.00 \\
        SFA (\textbf{ours}) & \textbf{2.99} & \textbf{25.2} & \textbf{95.6} & \textbf{357.3} & \textbf{2.50} & \textbf{22.7} & \textbf{75.3} & \textbf{265.6} \\
        LLSA & 1.21 & 11.9 & 35.8 & 76.0 & 1.95 & 10.0 & 29.7 & 59.7 \\
        SLA, budget-matched & 1.20 & 7.72 & 24.1 & 65.2 & 2.20 & 5.23 & 9.08 & 11.5 \\
        SLA, fixed 0.05 sparsity & 1.22 & 6.38 & 11.7 & 15.7 & 2.23 & 4.82 & 7.39 & 8.79 \\
        VSA, budget-matched & 0.75 & 12.6 & 39.2 & 60.6 & 1.72 & 17.7 & 43.7 & 80.9 \\
        VSA, fixed 0.10 sparsity & 0.76 & 6.59 & 7.60 & 7.98 & 1.92 & 5.89 & 6.48 & 6.81 \\
        \bottomrule
    \end{tabular}
\end{table}

\FloatBarrier

\subsection{FFHQ: Training and inference throughput and generation quality}
\label{app:end-to-end}
\label{app:ffhq-quality}

\begin{figure}[H]
    \centering
    \includegraphics[width=\linewidth]{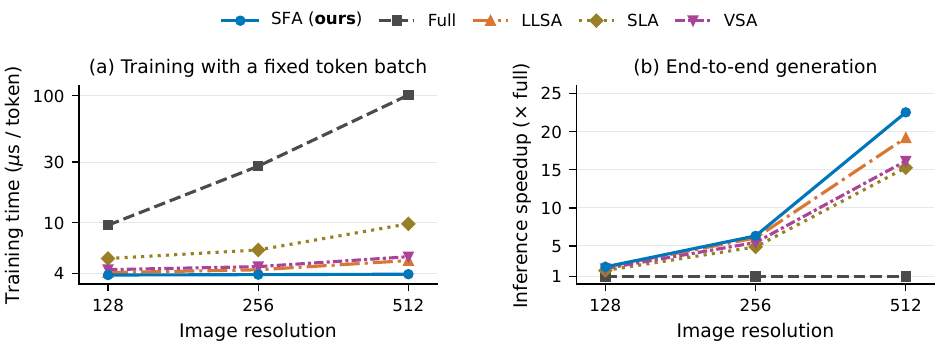}
    \caption{End-to-end scaling of pixel-space DiT-S on FFHQ. The left panel shows training time per token at a fixed batch of 262,144 tokens, using image batch sizes 16, 4, and 1 at resolutions 128, 256, and 512. The right panel shows generation speedup over full attention with image batch sizes 20, 4, and 1. SFA (\textbf{ours}) maintains nearly constant training time per token as resolution increases.}
    \label{fig:ffhq-scaling}
\end{figure}

To examine how the kernel gains carry through to a complete generator, Table~\ref{tab:ffhq-runtime-full} reports runtime and memory use at all three FFHQ resolutions. Training image batch sizes of 16, 4, and 1 keep the token count per step fixed at 262,144, and generation uses image batch sizes of 20, 4, and 1. This fixed training token batch makes the per-token scaling directly comparable: SFA takes 3.88, 3.91, and 3.94\,$\mu$s per token, whereas full-attention training time per token rises from 9.57 to 101.31\,$\mu$s over the same resolution range. Figure~\ref{fig:ffhq-scaling} summarizes this difference alongside the generation speedup.

Across this fixed token batch, step time grows by only $1.47\%$, while $\log_2N$ rises from 14 to 18, or $28.6\%$. SFA retains $O(N\log N)$ attention work, but its logarithmic contribution is small in the measured runtime. The remaining operations scale linearly with token count at fixed model size and dominate this regime, limiting further end-to-end gains from attention-only acceleration.

\Needspace{4\baselineskip}
\paragraph{Generation quality.}
Table~\ref{tab:ffhq-quality} and Figure~\ref{fig:overview-results}(D) compare four efficient-attention variants in pixel-space DiT-S models trained on FFHQ at resolutions 128 and 256, with 20 epochs at each resolution in the cascade of Table~\ref{tab:ffhq-schedule}. SFA achieves the lowest FID$_{50\mathrm{k}}$ at both resolutions, reaching 22.91 at resolution 128 and 30.39 at resolution 256. At resolution 256, the corresponding values for LLSA, SLA, and VSA are 32.03, 33.83, and 33.41, respectively.

\begin{table}[!htbp]
    \centering
    \caption{Complete FFHQ end-to-end runtime measurements on one RTX 5090, with gradient checkpointing enabled for training. Speedups are relative to full attention at the same resolution, and generation rates and speedups retain the precision of the measurements. Bold marks the lowest time and memory or the highest throughput and speedup, including ties.}
    \label{tab:ffhq-runtime-full}
    \setlength{\tabcolsep}{4pt}
    \begin{tabular}{clrrrrrr}
        \toprule
        & & \multicolumn{3}{c}{Training} & \multicolumn{3}{c}{Generation} \\
        \cmidrule(lr){3-5}\cmidrule(lr){6-8}
        Resolution & Method & ms/step & Speedup & GiB & images/s & Speedup & GiB \\
        \midrule
        128 & Full & 2510 & 1.00 & \textbf{10.57} & 1.38 & 1.00 & \textbf{3.44} \\
        & SFA (\textbf{ours}) & \textbf{1017} & \textbf{2.47} & 10.79 & \textbf{3.10} & \textbf{2.25} & \textbf{3.44} \\
        & LLSA & 1062 & 2.36 & 10.60 & 3.06 & 2.22 & \textbf{3.44} \\
        & SLA & 1371 & 1.83 & 12.27 & 2.45 & 1.78 & 4.15 \\
        & VSA & 1121 & 2.24 & 11.41 & 2.81 & 2.04 & 3.75 \\
        \midrule
        256 & Full & 7308 & 1.00 & \textbf{10.58} & 0.12 & 1.00 & \textbf{2.84} \\
        & SFA (\textbf{ours}) & \textbf{1024} & \textbf{7.13} & 10.80 & \textbf{0.77} & \textbf{6.32} & \textbf{2.84} \\
        & LLSA & 1118 & 6.53 & 10.60 & 0.74 & 6.05 & \textbf{2.84} \\
        & SLA & 1602 & 4.56 & 12.30 & 0.59 & 4.87 & 3.43 \\
        & VSA & 1184 & 6.17 & 11.60 & 0.67 & 5.45 & 3.21 \\
        \midrule
        512 & Full & 26557 & 1.00 & \textbf{10.60} & 0.0085 & 1.00 & \textbf{2.86} \\
        & SFA (\textbf{ours}) & \textbf{1032} & \textbf{25.75} & 10.83 & \textbf{0.19} & \textbf{22.54} & 2.87 \\
        & LLSA & 1322 & 20.08 & 10.63 & 0.16 & 19.20 & \textbf{2.86} \\
        & SLA & 2572 & 10.32 & 12.39 & 0.13 & 15.27 & 3.52 \\
        & VSA & 1417 & 18.75 & 12.32 & 0.14 & 16.15 & 3.74 \\
        \bottomrule
    \end{tabular}
\end{table}

\begin{table}[!t]
    \centering
    \caption{FID$_{50\mathrm{k}}$ for pixel-space DiT-S trained on FFHQ. The $128\times128$ and $256\times256$ results follow the 20-epoch stage at each resolution in Table~\ref{tab:ffhq-schedule} and use 50,000 generated samples. Lower values are better, and bold marks the lowest FID among the reported methods at each resolution.}
    \label{tab:ffhq-quality}
    \begin{tabular}{lrr}
        \toprule
        Method & $128\times128$ & $256\times256$ \\
        \midrule
        SFA (\textbf{ours}) & \textbf{22.91} & \textbf{30.39} \\
        LLSA & 23.24 & 32.03 \\
        SLA & 24.99 & 33.83 \\
        VSA & 23.17 & 33.41 \\
        \bottomrule
    \end{tabular}
\end{table}

At $512\times512$, SFA and LLSA continue from their respective $256\times256$ checkpoints for two epochs. SFA achieves FID$_{50\mathrm{k}}$ of 35.59, compared with 38.90 for LLSA, an improvement of 3.31. Together with the four-method comparisons at resolutions 128 and 256, this result shows that SFA's generation-quality advantage accompanies the throughput gains in Table~\ref{tab:ffhq-runtime-full}.

\Needspace{5\baselineskip}
\subsection{PixelFlow: Training and inference throughput and generation quality}
\label{app:pixelflow-full-results}

Table~\ref{tab:pixelflow-runtime-full} reports training and generation throughput and memory use for the complete PixelFlow model, with attention replaced in its 4096-token stage. These whole-model measurements include the cost of the stages that retain full attention.

\begin{table}[!htb]
    \centering
    \caption{Complete PixelFlow end-to-end measurements using an image batch size of 4 and gradient checkpointing for training, and an image batch size of 16 for generation. Each efficient operator replaces the 4096-token stage. Speedups are relative to Full, computed from the listed step times and generation rates. Bold marks the lowest time and memory or the highest throughput and speedup, including ties.}
    \label{tab:pixelflow-runtime-full}
    \setlength{\tabcolsep}{4pt}
    \begin{tabular}{lrrrrrr}
        \toprule
        & \multicolumn{3}{c}{Training} & \multicolumn{3}{c}{Generation} \\
        \cmidrule(lr){2-4}\cmidrule(lr){5-7}
        Method & ms/step & Speedup ($\times$) & GiB & images/s & Speedup ($\times$) & GiB \\
        \midrule
        Full & 308.6 & 1.00 & \textbf{13.41} & 0.178 & 1.00 & \textbf{7.24} \\
        SFA (\textbf{ours}) & \textbf{286.8} & \textbf{1.08} & 13.42 & \textbf{0.194} & \textbf{1.09} & \textbf{7.24} \\
        LLSA & 300.5 & 1.03 & 13.45 & 0.179 & 1.01 & 8.22 \\
        SLA & 316.7 & 0.97 & 13.53 & 0.171 & 0.96 & 9.81 \\
        VSA & 313.0 & 0.99 & 14.18 & 0.175 & 0.98 & 12.67 \\
        \bottomrule
    \end{tabular}
\end{table}

\begin{table}[!htb]
    \centering
    \caption{FID$_{50\mathrm{k}}$ during PixelFlow training from scratch on ImageNet-$256$ through epoch 40 of the five-method comparison. Lower values indicate better sample quality, and bold marks the lowest FID among the five methods at each epoch.}
    \label{tab:pixelflow-fid-full}
    \begin{tabular}{rrrrrr}
        \toprule
        Epoch & Full & SFA (\textbf{ours}) & LLSA & SLA & VSA \\
        \midrule
        5 & 65.06 & \textbf{64.77} & 65.92 & 64.95 & 65.90 \\
        10 & 41.88 & \textbf{40.66} & 41.66 & 41.10 & 42.62 \\
        15 & 31.14 & \textbf{29.45} & 31.19 & 30.06 & 31.13 \\
        20 & 24.35 & 23.58 & 24.73 & \textbf{23.43} & 25.27 \\
        25 & 19.95 & \textbf{19.25} & 20.25 & 19.38 & 21.11 \\
        30 & 16.93 & \textbf{16.12} & 17.19 & 16.59 & 18.10 \\
        35 & 14.70 & \textbf{14.03} & 14.72 & 14.31 & 15.84 \\
        40 & 13.25 & \textbf{12.51} & 13.22 & 12.68 & 13.89 \\
        \bottomrule
    \end{tabular}
\end{table}

\paragraph{Generation quality.}
The corresponding generation-quality trajectories appear in Tables~\ref{tab:pixelflow-fid-full} and~\ref{tab:pixelflow-is-full}, which trace FID and IS throughout the five-method ImageNet-$256$ comparison through epoch 40. SFA achieves the lowest FID at seven of the eight checkpoints and the highest IS at every checkpoint. SLA briefly leads in FID, reaching 23.43 at epoch 20 compared with 23.58 for SFA, and remains a strong quality baseline at epoch 40 with FID 12.68 versus 12.51. SLA's quality gains over full attention come with lower model throughput: Table~\ref{tab:pixelflow-runtime-full} reports 316.7 versus 308.6\,ms per training step and 0.171 versus 0.178 images/s during generation.

\begin{table}[!htb]
    \centering
    \caption{Inception Score during PixelFlow training from scratch on ImageNet-$256$ through epoch 40 of the five-method comparison, evaluated on 50,000 generated samples. Higher values are better, and bold marks the highest IS among the five methods at each epoch.}
    \label{tab:pixelflow-is-full}
    \begin{tabular}{rrrrrr}
        \toprule
        Epoch & Full & SFA (\textbf{ours}) & LLSA & SLA & VSA \\
        \midrule
        5 & 19.02 & \textbf{19.88} & 19.45 & 19.43 & 19.07 \\
        10 & 30.58 & \textbf{32.90} & 31.81 & 31.38 & 30.40 \\
        15 & 42.43 & \textbf{45.94} & 42.74 & 44.04 & 42.63 \\
        20 & 54.67 & \textbf{59.15} & 54.97 & 56.72 & 53.60 \\
        25 & 65.87 & \textbf{70.53} & 65.89 & 68.25 & 63.38 \\
        30 & 76.78 & \textbf{81.53} & 76.63 & 78.45 & 73.82 \\
        35 & 86.24 & \textbf{90.93} & 87.33 & 88.83 & 82.29 \\
        40 & 94.23 & \textbf{101.20} & 94.72 & 96.13 & 91.33 \\
        \bottomrule
    \end{tabular}
\end{table}

Beyond the common five-method budget, Table~\ref{tab:pixelflow-extended} follows full attention and SFA at epochs 45--80 to examine the effect of longer training. SFA maintains lower FID and higher IS at every recorded checkpoint, showing that its advantage over full attention persists through this extended period.

\begin{table}[!htb]
    \centering
    \caption{Extended ImageNet-$256$ training of full attention and SFA (\textbf{ours}) beyond the 40-epoch five-method comparison, with both metrics evaluated on 50,000 generated samples. Bold marks the lower FID and higher IS within each two-method comparison.}
    \label{tab:pixelflow-extended}
    \begin{tabular}{rrrrr}
        \toprule
        & \multicolumn{2}{c}{Full attention} & \multicolumn{2}{c}{SFA (\textbf{ours})} \\
        \cmidrule(lr){2-3}\cmidrule(lr){4-5}
        Epoch & FID$_{50\mathrm{k}}$ $\downarrow$ & IS $\uparrow$ & FID$_{50\mathrm{k}}$ $\downarrow$ & IS $\uparrow$ \\
        \midrule
        45 & 11.78 & 101.76 & \textbf{11.17} & \textbf{107.74} \\
        50 & 10.84 & 109.05 & \textbf{10.22} & \textbf{114.73} \\
        55 & 9.99 & 114.92 & \textbf{9.42} & \textbf{121.56} \\
        60 & 9.20 & 119.94 & \textbf{8.68} & \textbf{126.25} \\
        65 & 8.65 & 125.24 & \textbf{8.10} & \textbf{133.08} \\
        70 & 8.22 & 129.53 & \textbf{7.74} & \textbf{136.86} \\
        75 & 8.10 & 131.03 & \textbf{7.57} & \textbf{139.00} \\
        80 & 7.55 & 135.80 & \textbf{7.11} & \textbf{143.97} \\
        \bottomrule
    \end{tabular}
\end{table}

\subsection{Fine-tuning a pretrained PixelFlow model}

\begin{table}[!htbp]
    \centering
    \caption{Pretrained and adapted PixelFlow checkpoints, evaluated with 50,000 generated samples. The adapted models share the same initialization, learning-rate schedule, and 3.5-epoch budget. Among the fine-tuned models, bold marks the lower FID and higher IS.}
    \label{tab:pixelflow-finetune}
    \begin{tabular}{lrrr}
        \toprule
        Model & Fine-tuning epochs & FID$_{50\mathrm{k}}$ $\downarrow$ & IS $\uparrow$ \\
        \midrule
        Pretrained full attention & 0 & 2.15 & 277.84 \\
        \midrule
        Fine-tuned full attention & 3.5 & \textbf{2.28} & 273.25 \\
        Fine-tuned SFA (\textbf{ours}) & 3.5 & 2.29 & \textbf{284.71} \\
        \bottomrule
    \end{tabular}
\end{table}

To evaluate adaptation of pretrained weights, we start from the original checkpoint with FID$_{50\mathrm{k}}=2.15$ and IS $=277.84$ under our Euler sampling protocol. Evaluating the same checkpoint with the original work's dopri5 solver gives FID 2.00; its higher sampling cost leads us to use Euler sampling throughout the paper unless stated otherwise. Both full attention and SFA are fine-tuned from these weights at a learning rate of $10^{-5}$ for two epochs, followed by a three-epoch cosine decay to $10^{-6}$. Table~\ref{tab:pixelflow-finetune} compares the resulting models at epoch 3.5 under this shared budget, using 50,000 generated samples for each evaluation. SFA reaches FID 2.29 and IS 284.71, retaining FID comparable to full attention at 2.28 and improving on its IS of 273.25.

\FloatBarrier

\Needspace{5\baselineskip}
\subsection{Fine-tuning Stable Diffusion 3.5 Medium}
\label{app:sd35}

We extend the evaluation to pretrained text-to-image generation with Stable Diffusion 3.5 Medium~\citep{sd35medium,sd35article}, retaining its original image tokenization. Its variational autoencoder (VAE) compresses each spatial dimension by a factor of 8, after which the transformer embeds $2\times2$ latent patches, producing an effective patch size of $16\times16$ pixels. Image resolutions of $1024\times1024$ and $2048\times2048$ consequently give $64\times64$ and $128\times128$ image-token grids with 4096 and 16,384 image tokens, respectively. Joint attention combines these image tokens with the text tokens.

\phantomsection
\label{app:sd35-joint-replacement}
To distinguish the effect of attention replacement from subsequent adaptation, we compare six attention and fine-tuning settings. Of the 24 transformer blocks, the first 13 contain a separate image self-attention module alongside the joint attention. Full retains the original attention, and the self-attention-only setting (SA) replaces those 13 modules with SFA. ALL extends the replacement to the image-to-image interactions inside the joint attention of all 24 blocks, preserving the original computation for text-to-text, text-to-image, and image-to-text interactions. Image and text contributions share one softmax normalizer for each query. Because ALL covers more of the quadratic attention work, it reaches a larger speedup than SA at both resolutions (Table~\ref{tab:sd35-runtime}). The suffix +ft identifies a fine-tuned checkpoint. The quality tables group Full, SA, and ALL together, followed by Full+ft, SA+ft, and ALL+ft, so that each group compares methods within a shared adaptation budget.

\paragraph{Fine-tuning protocol.}
For the adaptation comparison, we fine-tune Full, SA, and ALL for one epoch on the first 30 shards of the Flux-generated dataset (about 12,000 steps at a global batch size of 16), using eight NVIDIA A800 GPUs and a learning rate of $10^{-5}$. Training uses AdamW with weight decay $10^{-4}$, gradient-norm clipping 1.0, bfloat16 parameters and optimizer states, and gradient checkpointing. All models start from the same pretrained checkpoint, and we evaluate their raw checkpoint weights.

\Needspace{22\baselineskip}
\paragraph{Runtime.}
For these same replacement scopes, Table~\ref{tab:sd35-runtime} compares training time, generation throughput, and memory use on one NVIDIA RTX 5090 GPU at image resolutions $1024\times1024$ and $2048\times2048$. Training uses image batch sizes 2 and 1, respectively, with gradient checkpointing enabled at both resolutions; generation uses batch sizes 4 and 1. At $2048\times2048$, ALL reduces the training step from 2214.7 to 1054.7\,ms and increases generation throughput from 0.02414 to 0.04767 images/s, with inference memory unchanged at 24.75\,GiB. Generation follows the released sampling defaults of 40 Euler steps at guidance scale 4.5, and each rate covers one complete pipeline call.

\begin{table}[H]
    \centering
    \caption{Stable Diffusion 3.5 Medium runtime on one RTX 5090. Image resolutions $1024\times1024$ and $2048\times2048$ give 4096 and 16,384 image tokens with effective $16\times16$ pixel patches. Training batch sizes are 2 and 1, and generation batch sizes are 4 and 1, respectively. Speedups are relative to Full at the same resolution and can be reproduced from the listed times and rates. Bold marks the lowest time and memory or the highest throughput and speedup at each resolution, including ties.}
    \label{tab:sd35-runtime}
    \setlength{\tabcolsep}{4pt}
    \begin{tabular}{llrrrrrr}
        \toprule
        & & \multicolumn{3}{c}{Training} & \multicolumn{3}{c}{Generation} \\
        \cmidrule(lr){3-5}\cmidrule(lr){6-8}
        Image resolution & Setting & ms/step & Speedup ($\times$) & GiB & images/s & Speedup ($\times$) & GiB \\
        \midrule
        $1024\times1024$ & Full & 686.5 & 1.00 & \textbf{26.69} & 0.1549 & 1.00 & \textbf{23.27} \\
        & SA & 644.4 & 1.07 & 26.72 & 0.1629 & 1.05 & \textbf{23.27} \\
        & ALL & \textbf{567.7} & \textbf{1.21} & 26.75 & \textbf{0.1852} & \textbf{1.20} & \textbf{23.27} \\
        \midrule
        $2048\times2048$ & Full & 2214.7 & 1.00 & \textbf{27.80} & 0.02414 & 1.00 & \textbf{24.75} \\
        & SA & 1797.4 & 1.23 & 27.85 & 0.02913 & 1.21 & \textbf{24.75} \\
        & ALL & \textbf{1054.7} & \textbf{2.10} & 27.93 & \textbf{0.04767} & \textbf{1.97} & \textbf{24.75} \\
        \bottomrule
    \end{tabular}
\end{table}

\Needspace{4\baselineskip}
\paragraph{DPG-Bench.}
To assess generation quality, we evaluate prompt following with the Dense Prompt Graph Benchmark (DPG-Bench; \citealp{dpgbench}), which scores alignment with the entities, attributes, and relationships in detailed prompts. Every benchmark image below is generated at $1024\times1024$ with the same 40 Euler steps and guidance scale 4.5. Following the DPG-Bench protocol, we generate four images for each of 1,065 prompts and score them as a $2\times2$ grid. Table~\ref{tab:sd35-dpg-summary} presents category and overall scores, and Table~\ref{tab:sd35-dpg-details} provides the subcategory breakdown under Attribute, Entity, Relation, and Other, with higher scores indicating better alignment. Direct SA replacement gives an overall score of 84.35, slightly below Full at 84.62, while ALL gives a comparable score of 84.74. Under the shared fine-tuning schedule, SA+ft reaches 85.68 and ALL+ft reaches 85.51, matching or exceeding the equally fine-tuned Full+ft at 85.51.

\begin{table}[H]
    \centering
    \setlength{\tabcolsep}{4pt}
    \caption{DPG-Bench category and overall scores for Stable Diffusion 3.5 Medium. The upper group compares Full, SA, and ALL without fine-tuning; the lower group compares them after the same one-epoch fine-tuning schedule. Bold marks the highest value within each group and metric, including ties. The subcategory breakdown follows in Table~\ref{tab:sd35-dpg-details}.}
    \label{tab:sd35-dpg-summary}
    \begin{tabular}{lrrrrrr}
        \toprule
        Setting & Attribute & Global & Relation & Entity & Other & \textbf{Overall} \\
        \midrule
        Full & \textbf{88.73} & 84.50 & 93.17 & 89.94 & 82.40 & 84.62 \\
        SA & 88.53 & 84.50 & 93.14 & 89.77 & \textbf{84.40} & 84.35 \\
        ALL & 88.51 & \textbf{86.32} & \textbf{93.60} & \textbf{90.23} & 79.60 & \textbf{84.74} \\
        \midrule
        Full+ft & \textbf{89.34} & 86.02 & 92.98 & 90.87 & \textbf{86.00} & 85.51 \\
        SA+ft & 89.06 & 84.50 & 93.33 & 90.85 & 84.40 & \textbf{85.68} \\
        ALL+ft & 89.16 & \textbf{86.32} & \textbf{93.41} & \textbf{90.98} & 82.80 & 85.51 \\
        \bottomrule
    \end{tabular}
\end{table}

\Needspace{23\baselineskip}
\begin{table}[H]
    \centering
    \setlength{\tabcolsep}{4pt}
    \caption{DPG-Bench subcategory scores for the same six settings as Table~\ref{tab:sd35-dpg-summary}. Bold marks the highest value in each row separately among the three settings without fine-tuning and among the three fine-tuned settings, including ties.}
    \label{tab:sd35-dpg-details}
    \begin{tabular}{llrrrrrr}
        \toprule
        & & \multicolumn{3}{c}{No fine-tuning} & \multicolumn{3}{c}{After fine-tuning} \\
        \cmidrule(lr){3-5}\cmidrule(lr){6-8}
        Category & Subcategory & Full & SA & ALL & Full+ft & SA+ft & ALL+ft \\
        \midrule
        \multirow{5}{*}{Attribute} & Color & \textbf{92.82} & 92.72 & 92.17 & \textbf{94.18} & 93.63 & 93.18 \\
        & Shape & 81.66 & \textbf{82.53} & 80.79 & 81.66 & \textbf{82.53} & 78.60 \\
        & Size & 72.73 & 70.66 & \textbf{73.55} & 68.18 & 68.60 & \textbf{71.49} \\
        & Texture & 88.14 & \textbf{88.69} & 87.78 & 88.69 & 88.69 & \textbf{88.81} \\
        & Other & 86.82 & 85.27 & \textbf{87.71} & 87.49 & 86.82 & \textbf{88.37} \\
        \midrule
        \multirow{3}{*}{Entity} & Part & 85.16 & 84.38 & \textbf{86.13} & 83.59 & \textbf{83.98} & 82.62 \\
        & State & 83.84 & 83.32 & \textbf{84.37} & 81.94 & 82.37 & \textbf{83.21} \\
        & Whole & 91.67 & 91.65 & \textbf{91.84} & \textbf{93.44} & 93.29 & \textbf{93.44} \\
        \midrule
        \multirow{2}{*}{Relation} & Non-spatial & 86.79 & 86.79 & \textbf{89.31} & 88.05 & 87.42 & \textbf{89.94} \\
        & Spatial & 93.59 & 93.55 & \textbf{93.88} & 93.30 & \textbf{93.71} & 93.63 \\
        \midrule
        \multirow{2}{*}{Other} & Count & 79.50 & \textbf{81.00} & 76.50 & \textbf{83.00} & 81.50 & 80.00 \\
        & Text & 94.00 & \textbf{98.00} & 92.00 & \textbf{98.00} & 96.00 & 94.00 \\
        \bottomrule
    \end{tabular}
\end{table}

\Needspace{4\baselineskip}
\paragraph{Human preference.}
Complementing the prompt-alignment results, Table~\ref{tab:sd35-hps} reports predicted human preference using human preference score (HPS) v2.1~\citep{wu2023hpsv2} across anime, concept art, paintings, and photographs. Each style contributes 800 prompts with one image each, and the overall score averages the four style means. Direct SA replacement slightly lowers the overall score from 29.967 to 29.596, and ALL produces a larger decrease to 27.927 despite its comparable DPG-Bench score. After fine-tuning, SA+ft reaches 30.242 and ALL+ft reaches 30.262, closely matching Full+ft at 30.261. The two benchmarks therefore show that both replacement scopes recover the quality of the identically fine-tuned original model.

\begin{table}[H]
    \centering
    \caption{HPS v2.1 scores for Stable Diffusion 3.5 Medium, with higher values indicating stronger predicted human preference. Within each adaptation group, either without fine-tuning or after the shared one-epoch fine-tuning schedule, bold marks the highest value, including ties.}
    \label{tab:sd35-hps}
    \setlength{\tabcolsep}{4pt}
    \begin{tabular}{lrrrrr}
        \toprule
        Setting & Anime & Concept art & Paintings & Photo & \textbf{Overall} \\
        \midrule
        Full & \textbf{31.399} & \textbf{30.526} & \textbf{30.478} & \textbf{27.466} & \textbf{29.967} \\
        SA & 31.162 & 30.128 & 29.954 & 27.141 & 29.596 \\
        ALL & 29.426 & 27.976 & 27.070 & 27.237 & 27.927 \\
        \midrule
        Full+ft & 31.066 & \textbf{30.499} & 30.512 & 28.968 & 30.261 \\
        SA+ft & \textbf{31.072} & 30.450 & 30.489 & 28.955 & 30.242 \\
        ALL+ft & 30.975 & 30.438 & \textbf{30.590} & \textbf{29.044} & \textbf{30.262} \\
        \bottomrule
    \end{tabular}
\end{table}

\FloatBarrier

\subsection{Radial attention statistics}
\label{app:powerlaw-details}

To relate the generation experiments to learned spatial structure, we measure attention in a pretrained PixelFlow-XL with 28 layers, 16 heads, and head dimension 72. Its final stage operates at $256\times256$ resolution, where $4\times4$ patches form a $64\times64$ token grid. We select one image from each of the 1000 ImageNet training classes, then resize each image by its shorter side, center-crop it to $256\times256$, and normalize it to $[-1,1]$.

To follow the final stage's training interpolation, let $\vf$ denote the three-channel image flattened into a vector and $\bar{\vf}$ the vector obtained by twofold bilinear downsampling followed by nearest-neighbor upsampling. We introduce a noise vector $\rvepsilon$ with the same dimension as $\vf$ and independent standard normal entries. At normalized within-stage time $\tau=0.95$, the resulting noisy image vector supplied to the model is
\begin{equation}
    \rvf_\tau=\tau\vf+(1-\tau)(0.6\bar{\vf}+0.4\rvepsilon).
\end{equation}
Using this input with its corresponding timestep and class label, we retain the original scaled dot-product attention computation for model outputs and additionally compute the attention probabilities from the queries and keys. These probabilities are collected for every head in each of the 28 layers.

For a fixed image, layer, and head, the softmax attention matrix $\mA=(a_{ij})_{i,j=1}^{N}$ contains probabilities $a_{ij}$ assigned by query $i$ to key $j$, with $N=4096$. To isolate interactions between distinct spatial positions, we exclude the diagonal and renormalize each row, obtaining
\begin{equation}
    \widetilde a_{ij}=\frac{a_{ij}}{1-a_{ii}},\qquad j\ne i,\qquad \sum_{j\ne i}\widetilde a_{ij}=1.
\end{equation}
We convert these query-wise probabilities into radial statistics by averaging over images and token pairs at equal Euclidean distance. The pairs are grouped by the 1576 exact squared-distance values of the $64\times64$ lattice, with the diagonal kept separate, and their attention mass is accumulated. For plotting, we combine positive distances by quantiles of token-pair counts, requesting 100 bins. The plotted value for each bin is its attention mass divided by the number of token pairs, and the horizontal coordinate is the pair-count-weighted mean distance. This weighting accounts for the discrete multiplicities of lattice distances, with a uniform reference of $1/(4096-1)$ for the renormalized rows.

This aggregation produces a curve $s_{\ell h}(r)$ for each layer--head pair, where $\ell\in\{1,\ldots,28\}$ indexes the layer, $h\in\{1,\ldots,16\}$ indexes the head, and $r$ is spatial distance. Let $\mathcal H(r)$ collect the pairs with a positive value in the corresponding distance bin. Their geometric mean summarizes the spatial pattern across layers and heads and is plotted in Figure~\ref{fig:main-overview}(C) against distance $r$ in pixels on a base-2 logarithmic scale:
\begin{equation}
    s_{\mathrm{geo}}(r)=\exp\left[ \frac{1}{|\mathcal H(r)|}\sum_{(\ell,h)\in\mathcal H(r)} \log s_{\ell h}(r) \right].
\end{equation}
For an individual or aggregate radial curve $s(r)$, we fit $\log s=b_{\mathrm{fit}}-\alpha\log r$, with intercept $b_{\mathrm{fit}}$ and decay exponent $\alpha$, over 8--160 pixels. The exponent follows the convention $s(r)\propto r^{-\alpha}$. To compare the radial decay forms, we also fit $s=a_{\mathrm{fit}}\log r+c_{\mathrm{fit}}$, where $a_{\mathrm{fit}}$ and $c_{\mathrm{fit}}$ are fitted coefficients, and evaluate both coefficients of determination $R^2$ from residuals on the original $s$ scale. Table~\ref{tab:powerlaw-fits} reports this comparison together with a random-initialization measurement using the same architecture, images, and interpolation. The randomly initialized model's geometric-mean curve stays close to the uniform reference, in contrast to the pronounced spatial decay learned by the pretrained model.

\begin{table}[!htbp]
    \centering
    \caption{Power-law and logarithmic fits to radial attention over 2--40 patches, with decay exponent defined by $s(r)\propto r^{-\alpha}$. The final row summarizes the individual fitted curves through their median parameters and fit scores.}
    \label{tab:powerlaw-fits}
    \begin{tabular}{llrrr}
        \toprule
        Weights & Aggregation & $\alpha$ & Power-law $R^2$ & Logarithmic $R^2$ \\
        \midrule
        Pretrained & Geometric mean & 1.414 & 0.974 & 0.589 \\
        Pretrained & Pointwise median & 1.306 & 0.986 & 0.635 \\
        Random & Geometric mean & $-0.009$ & 0.325 & 0.328 \\
        Pretrained & Median over 448 individual fits & 1.437 & 0.929 & 0.673 \\
        \bottomrule
    \end{tabular}
\end{table}

The aggregate trend is also reflected in individual layer--head curves: 321 of 448 favor the power-law fit over logarithmic decay, and 261 have power-law $R^2>0.9$. For the 16 heads, the median decay exponents across layers range from 0.993 to 1.692. To assess sensitivity to the distance grouping, we also use logarithmically spaced and linearly spaced bins, obtaining geometric-mean exponents of 1.205 and 1.324, respectively. Their power-law $R^2$ values are 0.829 and 0.977, compared with 0.622 and 0.748 for logarithmic decay, so the spatial trend persists across these binning choices.

These measurements of the radial structure of normalized attention support SFA's scale-free prior and connect the learned attention pattern to the image-statistics motivation. The relationship between image covariance and learned logits provides the corresponding theoretical interpretation.

\FloatBarrier

\subsection{Ablations}
\label{app:ablation}

We examine count correction, normalization across scales, far-field context, and block aggregation through four variants of SFA. We first report their generation quality and then analyze how these design choices affect computational efficiency.

\textbf{Variant 1} removes the $\log|B|$ correction while retaining mean pooling and shared softmax. \textbf{Variant 2} retains mean pooling but applies softmax separately at each scale and sums the resulting outputs. \textbf{Variant 3} retains only SFA's exact local neighborhood: the $3\times3$ neighborhood of $4\times4$ base blocks, covering $12\times12$ tokens away from image boundaries. It applies softmax over these local tokens and omits all pooled far-field representatives, with the neighborhood clipped at image boundaries. \textbf{Variant 4} replaces mean pooling with learnable aggregation and omits the $\log|B|$ correction, as FMA does \citep{kang2025fma}, while retaining shared softmax.

\begin{figure}[!htb]
    \centering
    \includegraphics[width=\linewidth]{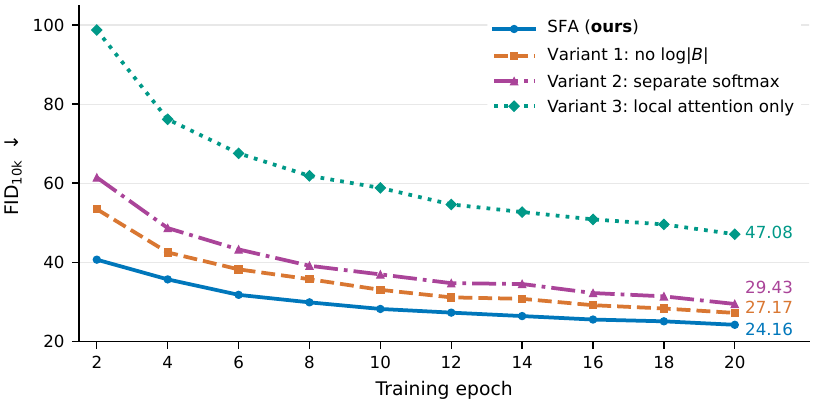}
    \caption{FID$_{10\mathrm{k}}$ during the FFHQ-$128$ stage. Each point evaluates 10,000 generated samples at a two-epoch interval. SFA (\textbf{ours}) achieves lower FID than all three ablations at every recorded checkpoint; Variant 3 retains only the exact local neighborhood.}
    \label{fig:ffhq-ablation-training}
\end{figure}

\paragraph{FFHQ training.}
We compare SFA with Variants 1--3 at FFHQ-$128$, following the cascade in Table~\ref{tab:ffhq-schedule} and the settings in Table~\ref{tab:training-settings}. The four operators therefore start from the same $32\times32$ model and share the same 20-epoch budget. Figure~\ref{fig:ffhq-ablation-training} traces FID$_{10\mathrm{k}}$ every two epochs through epoch 20, and Table~\ref{tab:ablation-ffhq} reports the complete trajectory together with the final FID$_{50\mathrm{k}}$. SFA achieves the lowest FID$_{10\mathrm{k}}$ at every recorded checkpoint. At epoch 20, removing count correction increases FID$_{50\mathrm{k}}$ from 22.91 to 25.91, while separate softmax increases it to 27.99. Retaining only the exact local neighborhood raises it to 45.76, showing the substantial contribution of pooled far-field information beyond the local attention branch.

\begin{table}[!htbp]
    \centering
    \caption{Ablations of count correction, shared softmax, and far-field summaries on FFHQ-$128$. The first ten data rows report FID$_{10\mathrm{k}}$; the final row reports FID$_{50\mathrm{k}}$ at epoch 20. Lower values indicate better generation quality, and bold marks the lowest FID in each row.}
    \label{tab:ablation-ffhq}
    \begin{tabular}{lrrrr}
        \toprule
        Epochs & SFA (\textbf{ours}) & Variant 1 & Variant 2 & Variant 3 \\
        \midrule
        2 & \textbf{40.64} & 53.41 & 61.43 & 98.73 \\
        4 & \textbf{35.65} & 42.47 & 48.63 & 76.14 \\
        6 & \textbf{31.74} & 38.19 & 43.23 & 67.54 \\
        8 & \textbf{29.84} & 35.70 & 39.10 & 61.84 \\
        10 & \textbf{28.17} & 33.01 & 36.91 & 58.79 \\
        12 & \textbf{27.24} & 31.10 & 34.69 & 54.61 \\
        14 & \textbf{26.37} & 30.73 & 34.48 & 52.69 \\
        16 & \textbf{25.49} & 29.12 & 32.19 & 50.83 \\
        18 & \textbf{25.04} & 28.28 & 31.36 & 49.55 \\
        20 & \textbf{24.16} & 27.17 & 29.43 & 47.08 \\
        \midrule
        20 (FID$_{50\mathrm{k}}$) & \textbf{22.91} & 25.91 & 27.99 & 45.76 \\
        \bottomrule
    \end{tabular}
\end{table}

\Needspace{4\baselineskip}
\paragraph{Direct replacement in Stable Diffusion 3.5 Medium.}
To assess pretrained-weight reuse, we also compare SFA with Variants 1 and 2 by directly replacing only image self-attention in Stable Diffusion 3.5 Medium, preserving joint attention and applying no fine-tuning. This follows the SA replacement setting in Appendix~\ref{app:sd35}. Table~\ref{tab:ablation-sd35} reports HPS v2.1 for anime, concept art, paintings, and photographs, together with the overall score, under the same evaluation protocol as Table~\ref{tab:sd35-hps}.

\begin{table}[!htb]
    \centering
    \caption{HPS v2.1 ablation for direct self-attention replacement in Stable Diffusion 3.5 Medium without fine-tuning. Higher values indicate stronger predicted human preference; bold marks the highest score in each column.}
    \label{tab:ablation-sd35}
    \setlength{\tabcolsep}{5pt}
    \begin{tabular}{lrrrrr}
        \toprule
        Attention & Anime & Concept art & Paintings & Photo & Overall \\
        \midrule
        SFA (\textbf{ours}) & \textbf{31.162} & \textbf{30.128} & \textbf{29.954} & \textbf{27.141} & \textbf{29.596} \\
        Variant 1 & 30.144 & 29.001 & 28.534 & 26.032 & 28.428 \\
        Variant 2 & 11.484 & 11.401 & 11.104 & 10.715 & 11.176 \\
        \bottomrule
    \end{tabular}
\end{table}

\begin{figure}[H]
    \centering
    \begin{subfigure}[t]{0.32\linewidth}
        \centering
        \includegraphics[width=\linewidth]{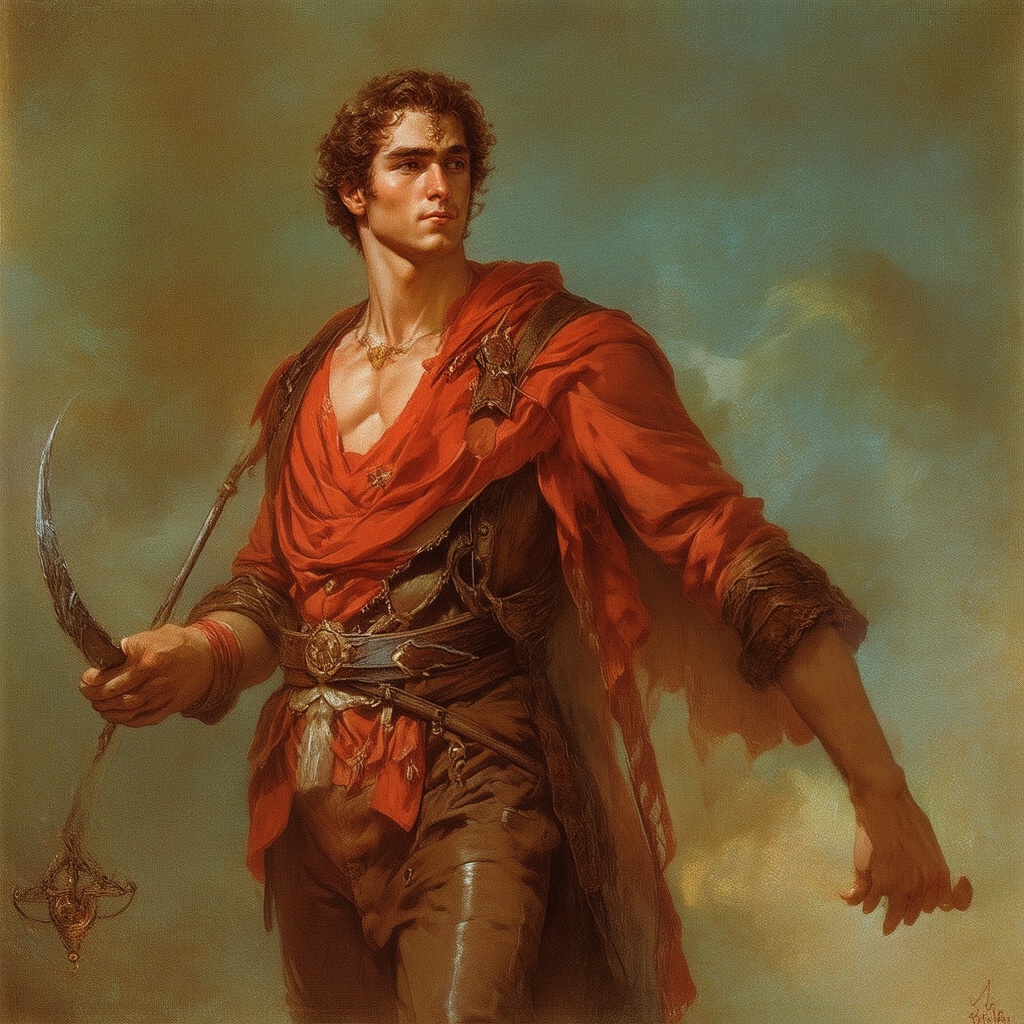}
        \caption{SFA (\textbf{ours})}
        \label{fig:ablation-sd35-sfa}
    \end{subfigure}\hfill
    \begin{subfigure}[t]{0.32\linewidth}
        \centering
        \includegraphics[width=\linewidth]{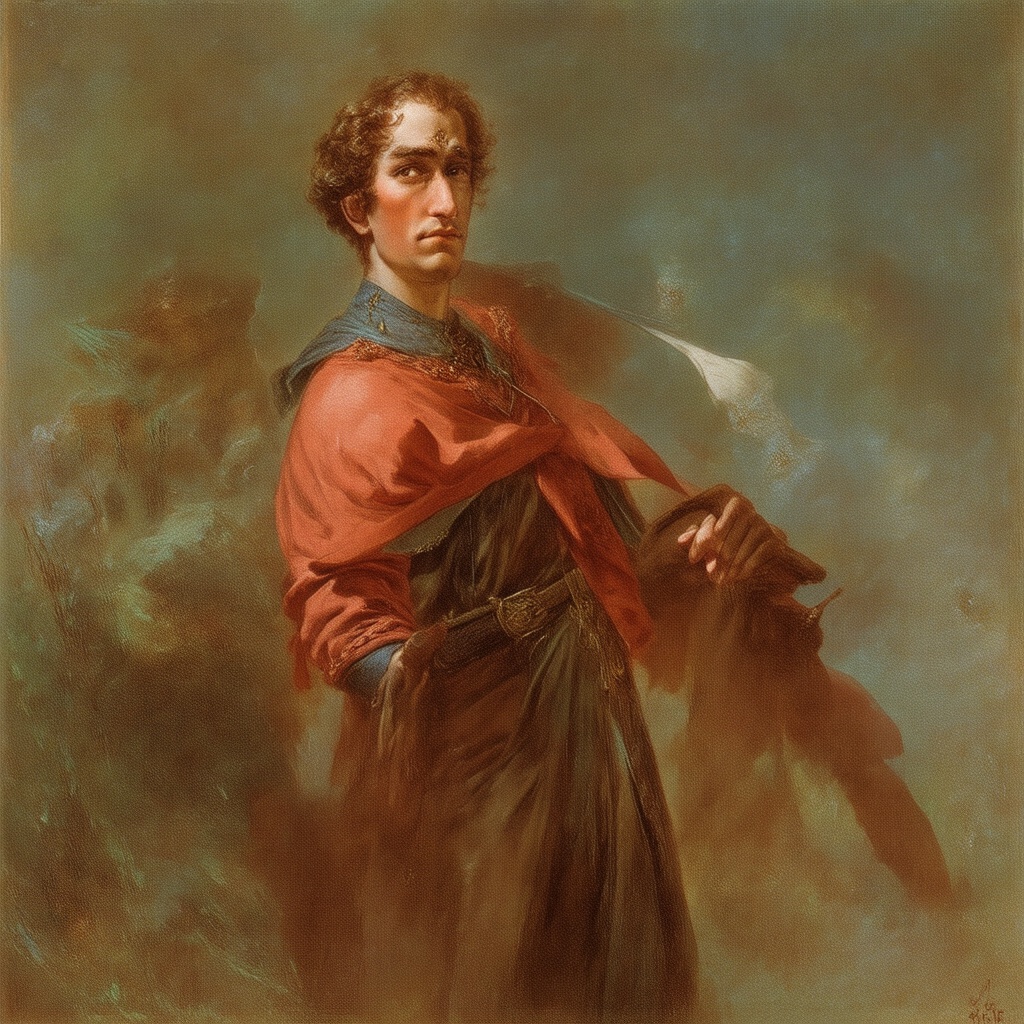}
        \caption{Variant 1: no $\log|B|$}
        \label{fig:ablation-sd35-nologb}
    \end{subfigure}\hfill
    \begin{subfigure}[t]{0.32\linewidth}
        \centering
        \includegraphics[width=\linewidth]{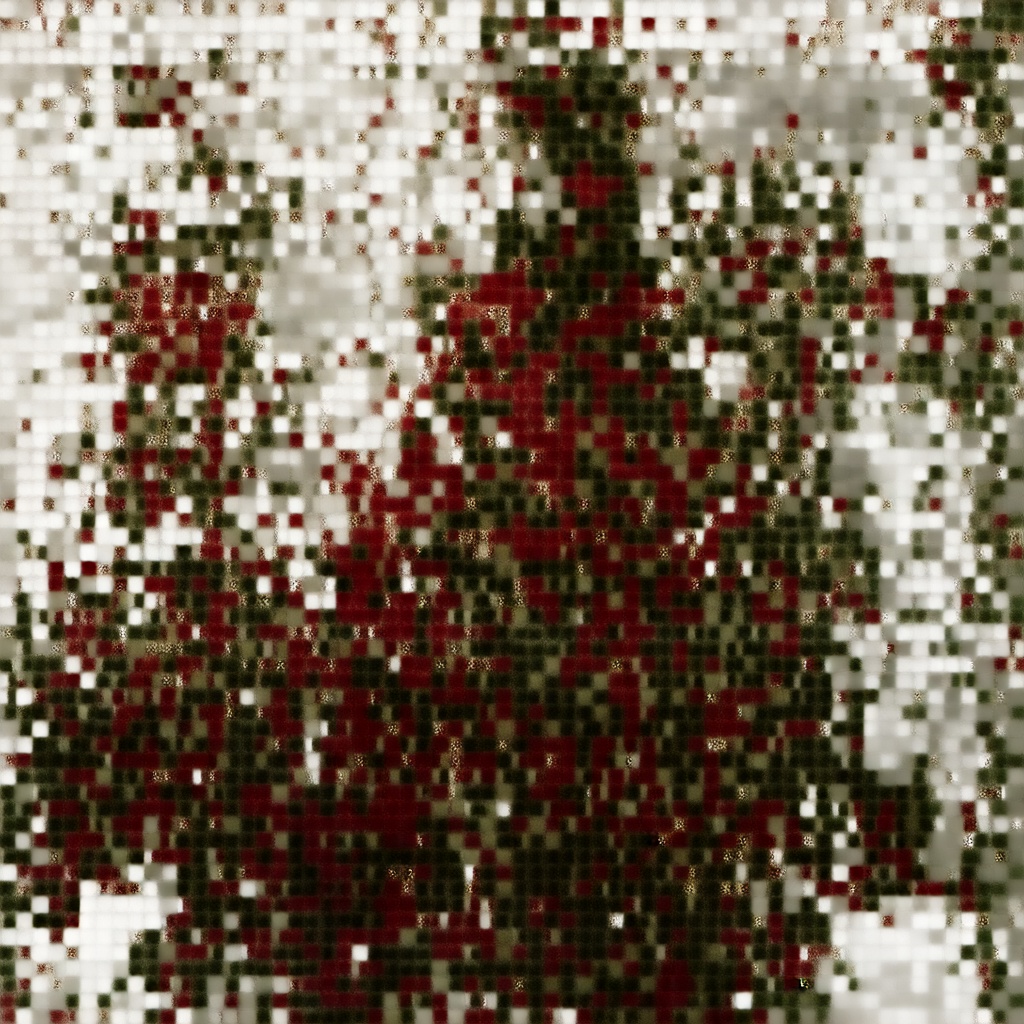}
        \caption{Variant 2: separate softmax}
        \label{fig:ablation-sd35-separate}
    \end{subfigure}
    \caption{Qualitative ablation of direct self-attention replacement in Stable Diffusion 3.5 Medium. All three $1024\times1024$ images are generated without fine-tuning using the prompt ``Male character illustration by Gaston Bussiere.'' Panels compare (a) SFA (\textbf{ours}), (b) removal of count correction, and (c) separate softmax at each scale.}
    \label{fig:ablation-sd35}
\end{figure}

SFA achieves the highest score in all four categories. Removing count correction lowers overall HPS from 29.596 to 28.428, while separate softmax reduces it to 11.176. The qualitative comparison in Figure~\ref{fig:ablation-sd35} illustrates these differences under a common prompt: SFA produces a detailed character illustration, Variant 1 loses definition in the clothing and background, and Variant 2 is dominated by grid-like artifacts that obscure the subject. These results support the importance of count correction and shared normalization for preserving generation quality when reusing pretrained weights.

\Needspace{4\baselineskip}
\paragraph{Learnable aggregation in PixelFlow.}
Variant 4 introduces aggregation parameters that must be learned, so we evaluate it through training from scratch on ImageNet-$256$ with PixelFlow. After ten epochs under the setup in Appendix~\ref{app:pixelflow-full-results}, SFA achieves FID$_{50\mathrm{k}}$ of 40.66 and IS of 32.90, compared with 41.40 and 31.50, respectively, for Variant 4.

\paragraph{Interpretation.}
Variant 1 supports weighting pooled blocks by their original token counts, and Variant 2 shows the benefit of normalizing all scales together. Their Stable Diffusion 3.5 Medium results also connect these choices to pretrained-weight reuse. Variant 3 demonstrates that pooled far-field information improves generation beyond the exact local branch. Finally, the comparison with Variant 4 favors SFA's mean pooling with count correction over learnable aggregation without that correction at the same ten-epoch budget.

\paragraph{Efficiency.}
Variants 1 and 3 run at approximately the same speed as SFA in our experiments, despite substantially lower generation quality. Variant 2 introduces a separate softmax denominator and output-dependent gradient correction at every scale. Each query therefore requires scale-specific normalization and backward statistics, giving up the shared normalization and gradient correction used by SFA across all interactions. Variant 4 loses two useful properties of mean pooling. Independently learned aggregation weights at different scales generally prevent a parent representative from being constructed from its four existing child summaries, removing the inexpensive recursive construction available to SFA. During backward, spatially varying weights require distinct gradient contributions for individual token positions, replacing SFA's uniform gradient sharing within each base block; learning those weights adds parameter-gradient computations. These structural differences increase aggregation and differentiation work and remove the computational reuse described in Appendix~\ref{app:implementation}. Our current implementation of Variant 4 shows a substantial runtime penalty on an RTX 5090 with batch size 4, six heads, head dimension 64, and bfloat16 inputs. For $N=4096,16384,65536,262144$ tokens, its forward runtimes are $2.64\times$, $2.27\times$, $1.96\times$, and $1.98\times$ those of SFA; its backward runtimes are $4.74\times$, $2.89\times$, $3.10\times$, and $3.14\times$, respectively. This slowdown substantially reduces or even eliminates the speed advantage over other efficient-attention mechanisms. Timings are medians after warmup. Forward is measured without gradient recording and includes aggregation; backward is measured separately on a retained graph and includes gradients through aggregation and its learned weights.

\FloatBarrier

\subsection{Qualitative samples}
\label{app:qualitative}

We provide additional generated images below for FFHQ, ImageNet, and Stable Diffusion 3.5 Medium, complementing the examples in Figure~\ref{fig:overview-results}(E).

\paragraph{FFHQ.}
Figure~\ref{fig:ffhq-samples} presents additional samples from pixel-space face generation on FFHQ at $128\times128$, $256\times256$, and $512\times512$ resolution, extending the examples in Figure~\ref{fig:overview-results}(E). These samples are generated by a pixel-space DiT-S with SFA in all attention layers. The three groups display the resolutions from highest to lowest, with two, four, and eight samples, respectively.

\begingroup
\makeatletter
\setlength{\@fptop}{0pt}
\makeatother
\setlength{\parskip}{0pt}
\newlength{\ffhqSampleWidth}
\setlength{\ffhqSampleWidth}{\dimexpr(\linewidth-6pt)/14\relax}
\newcommand{\ffhqSample}[2]{%
    \hbox{\includegraphics[width=#1]{figures/moreexample/FFHQ/#2.png}}%
}

\begin{figure}[!ht]
    \centering
    {\small
        \makebox[8\ffhqSampleWidth]{$512\times512$}\hspace{3pt}%
        \makebox[4\ffhqSampleWidth]{$256\times256$}\hspace{3pt}%
        \makebox[2\ffhqSampleWidth]{$128\times128$}\par
    }
    \vspace{4pt}
    \hbox{%
        \vbox{%
            \offinterlineskip
            \hbox{%
                \ffhqSample{4\ffhqSampleWidth}{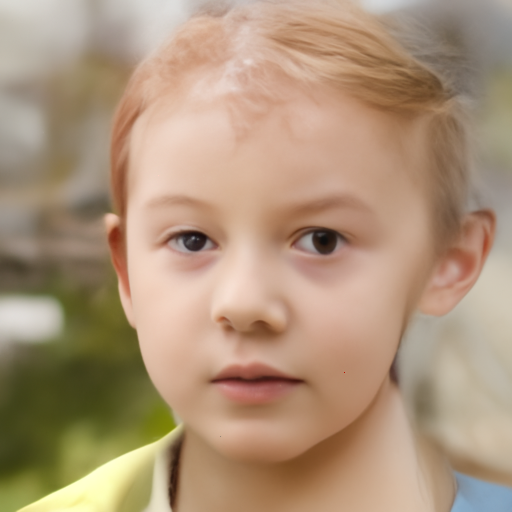}%
                \ffhqSample{4\ffhqSampleWidth}{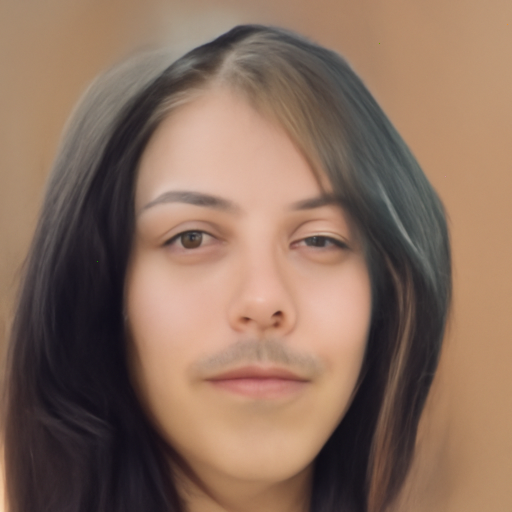}%
            }%
        }%
        \hskip3pt%
        \vbox{%
            \offinterlineskip
            \hbox{%
                \ffhqSample{2\ffhqSampleWidth}{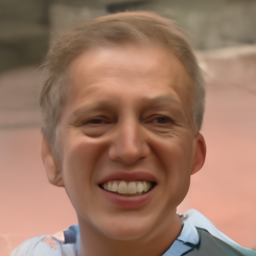}%
                \ffhqSample{2\ffhqSampleWidth}{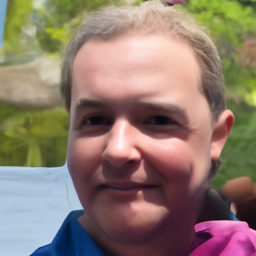}%
            }%
            \hbox{%
                \ffhqSample{2\ffhqSampleWidth}{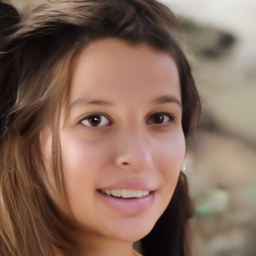}%
                \ffhqSample{2\ffhqSampleWidth}{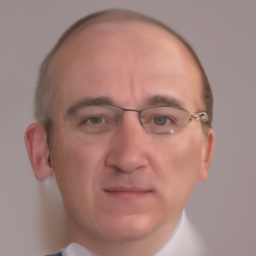}%
            }%
        }%
        \hskip3pt%
        \vbox{%
            \offinterlineskip
            \hbox{%
                \ffhqSample{\ffhqSampleWidth}{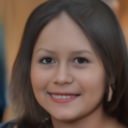}%
                \ffhqSample{\ffhqSampleWidth}{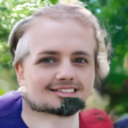}%
            }%
            \hbox{%
                \ffhqSample{\ffhqSampleWidth}{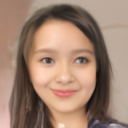}%
                \ffhqSample{\ffhqSampleWidth}{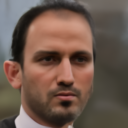}%
            }%
            \hbox{%
                \ffhqSample{\ffhqSampleWidth}{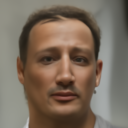}%
                \ffhqSample{\ffhqSampleWidth}{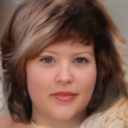}%
            }%
            \hbox{%
                \ffhqSample{\ffhqSampleWidth}{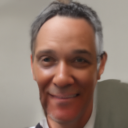}%
                \ffhqSample{\ffhqSampleWidth}{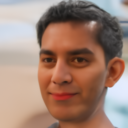}%
            }%
        }%
    }
    \caption{Pixel-space face samples on FFHQ at $512\times512$ (left), $256\times256$ (middle), and $128\times128$ (right). The three groups contain two, four, and eight images, respectively, with displayed side lengths proportional to image resolution.}
    \label{fig:ffhq-samples}
\end{figure}

\FloatBarrier
\endgroup

\paragraph{ImageNet.}
Figure~\ref{fig:imagenet-samples} presents five groups of class-conditional ImageNet samples generated with PixelFlow. The left panel compares Full, LLSA, SLA, VSA, and SFA after 40 epochs of training from scratch; the right panel compares the original pretrained weights with the full-attention and SFA models after fine-tuning. This organization complements the training comparison in Table~\ref{tab:pixelflow-main} and the adaptation comparison in Table~\ref{tab:pixelflow-finetune}.

\begingroup
\setlength{\parskip}{0pt}
\newlength{\imagenetSampleWidth}
\setlength{\imagenetSampleWidth}{0.116\linewidth}
\newlength{\imagenetScratchWidth}
\setlength{\imagenetScratchWidth}{5\imagenetSampleWidth}
\addtolength{\imagenetScratchWidth}{8pt}
\newlength{\imagenetFinetuneWidth}
\setlength{\imagenetFinetuneWidth}{3\imagenetSampleWidth}
\addtolength{\imagenetFinetuneWidth}{4pt}
\newcommand{\imagenetSample}[2]{%
    \includegraphics[width=\imagenetSampleWidth]{figures/moreexample/imagenet/#1/#2.png}%
}
\newcommand{\imagenetScratchRow}[1]{%
    \imagenetSample{fromscratch}{full_#1} &
    \imagenetSample{fromscratch}{llsa_#1} &
    \imagenetSample{fromscratch}{sla_#1} &
    \imagenetSample{fromscratch}{vsa_#1} &
    \imagenetSample{fromscratch}{sfa_#1}%
}
\newcommand{\imagenetFinetuneRow}[1]{%
    \imagenetSample{ft}{full_#1} &
    \imagenetSample{ft}{fullft_#1} &
    \imagenetSample{ft}{sfaft_#1}%
}

\begin{figure}[!t]
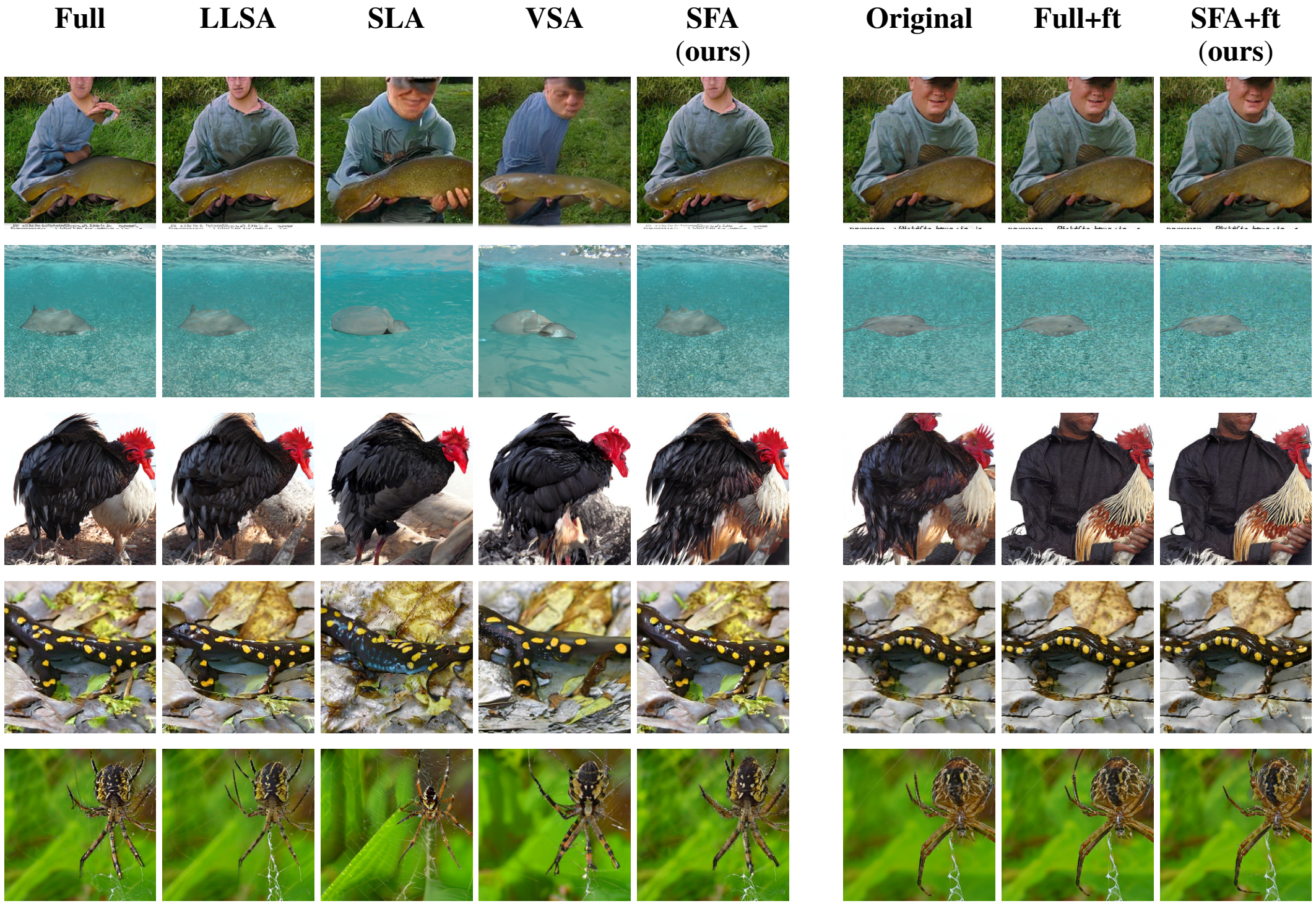

    \centering
    \begin{subfigure}[t]{\imagenetScratchWidth}
        \centering
        \small
        \setlength{\tabcolsep}{1pt}
        \begin{tabular}{@{}ccccc@{}}
            \textbf{Full} & \textbf{LLSA} & \textbf{SLA} & \textbf{VSA} & \textbf{SFA} \\
            & & & & (\textbf{ours}) \\[2pt]
            \imagenetScratchRow{1} \\[2pt]
            \imagenetScratchRow{7} \\[2pt]
            \imagenetScratchRow{8} \\[2pt]
            \imagenetScratchRow{29} \\[2pt]
            \imagenetScratchRow{75}
        \end{tabular}
        \caption{From scratch (epoch 40).}
        \label{fig:imagenet-samples-scratch}
    \end{subfigure}\hfill
    \begin{subfigure}[t]{\imagenetFinetuneWidth}
        \centering
        \small
        \setlength{\tabcolsep}{1pt}
        \begin{tabular}{@{}ccc@{}}
            \textbf{Original} & \textbf{Full+ft} & \textbf{SFA+ft} \\
            & & (\textbf{ours}) \\[2pt]
            \imagenetFinetuneRow{1} \\[2pt]
            \imagenetFinetuneRow{7} \\[2pt]
            \imagenetFinetuneRow{8} \\[2pt]
            \imagenetFinetuneRow{29} \\[2pt]
            \imagenetFinetuneRow{75}
        \end{tabular}
        \caption{Fine-tuning.}
        \label{fig:imagenet-samples-finetune}
    \end{subfigure}
    \caption{PixelFlow samples on ImageNet at $256\times256$ resolution. (a) Five attention methods after 40 epochs of training from scratch. (b) The original pretrained full-attention model (Original), the full-attention model after fine-tuning (Full+ft), and the model fine-tuned with SFA (\textbf{ours}) in its highest-resolution stage, labeled SFA+ft (\textbf{ours}). The five sample groups are aligned by row across both panels. In panel (a), SFA (\textbf{ours}) produces the most coherent human anatomy in the scene with a person holding a fish (first row). In the ray scene (second row), SFA (\textbf{ours}) renders the finest ripple patterns, while SLA and VSA produce less realistic water textures.}
    \label{fig:imagenet-samples}
\end{figure}

\endgroup

\Needspace{5\baselineskip}
\paragraph{Stable Diffusion 3.5 Medium.}
We present text-to-image samples generated with Stable Diffusion 3.5 Medium. The $1024\times1024$ example in Figure~\ref{fig:overview-results}(E) uses the prompt ``Black and white portrait of Thabo Mbeki with highly detailed ink lines and a cyberpunk flair, created for the Inktober challenge as part of the Cyberpunk 2020 manual coloring pages''.

Figures~\ref{fig:sd35-beforeft-hps} and~\ref{fig:sd35-beforeft-geneval} compare ten groups of samples before fine-tuning. From left to right, each row shows samples from the original full-attention model (Full), the model with SFA replacing image self-attention (SA), and the model with SFA replacing both image self-attention and image-to-image interactions in joint attention (ALL). The three images in each row share the same prompt and are generated at $1024\times1024$ resolution. This comparison illustrates the immediate effect of attention replacement under the settings in Appendix~\ref{app:sd35}.

Figures~\ref{fig:sd35-afterft-hps} and~\ref{fig:sd35-afterft-dpg} extend the comparison to ten further examples after the shared one-epoch fine-tuning schedule, with columns for Full+ft, SA+ft, and ALL+ft. These samples are generated with the raw checkpoints evaluated in Tables~\ref{tab:sd35-hps} and~\ref{tab:sd35-dpg-summary}, with prompts matched across models and the same $1024\times1024$ image resolution.

\begingroup
\makeatletter
\setlength{\@fptop}{0pt}
\makeatother
\setlength{\parskip}{0pt}

\newcommand{\sdBeforeHeader}{%
    \parbox[t]{0.17\linewidth}{\textbf{Prompt}}\hfill
    \parbox[t]{0.255\linewidth}{\centering\textbf{Full}\\\small Original attention}\hfill
    \parbox[t]{0.255\linewidth}{\centering\textbf{SA}\\\small Only self-attention\\replaced}\hfill
    \parbox[t]{0.255\linewidth}{\centering\textbf{ALL}\\\small All attention\\replaced}\par
    \vspace{4pt}
}
\newcommand{\sdBeforeRow}[3]{%
    \begin{minipage}[c]{0.17\linewidth}
        \raggedright
        \small #3
    \end{minipage}\hfill
    \begin{minipage}[c]{0.255\linewidth}
        \includegraphics[width=\linewidth]{figures/moreexample/sd35medium/beforeft/#1_full.#2}
    \end{minipage}\hfill
    \begin{minipage}[c]{0.255\linewidth}
        \includegraphics[width=\linewidth]{figures/moreexample/sd35medium/beforeft/#1_sa.#2}
    \end{minipage}\hfill
    \begin{minipage}[c]{0.255\linewidth}
        \includegraphics[width=\linewidth]{figures/moreexample/sd35medium/beforeft/#1_all.#2}
    \end{minipage}\par
    \vspace{5pt}
}

\begin{figure}[p]
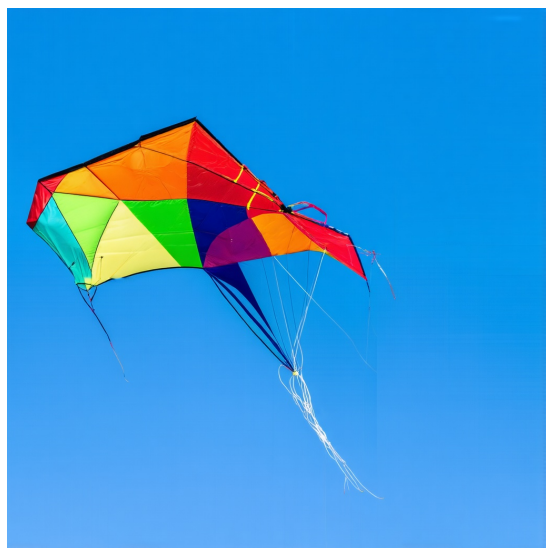

    \centering
    \sdBeforeHeader
    \sdBeforeRow{hps/concept_art_00195}{jpg}{Close-up portrait of a goblin emperor with a bone headdress, depicted in vibrant colors with intricate details.}
    \sdBeforeRow{hps/concept_art_00260}{jpg}{An astronaut in white futuristic cybernetic armor running on the surface of the moon, featured in an artwork illustration on Artstation.}
    \sdBeforeRow{hps/paintings_00400}{jpg}{A birch forest in autumn with falling leaves that resemble flying butterflies and dancing elves.}
    \sdBeforeRow{hps/photo_00246}{jpg}{a guy in the desert  sitting on his motorcycle}
    \sdBeforeRow{hps/photo_00276}{jpg}{A colorful kite is ready for launch on a blue sky day}
    \caption{Samples from Stable Diffusion 3.5 Medium before fine-tuning. Each row compares Full, SA, and ALL using the same prompt. ALL replaces both self-attention and joint-attention operators.}
    \label{fig:sd35-beforeft-hps}
\end{figure}

\begin{figure}[p]
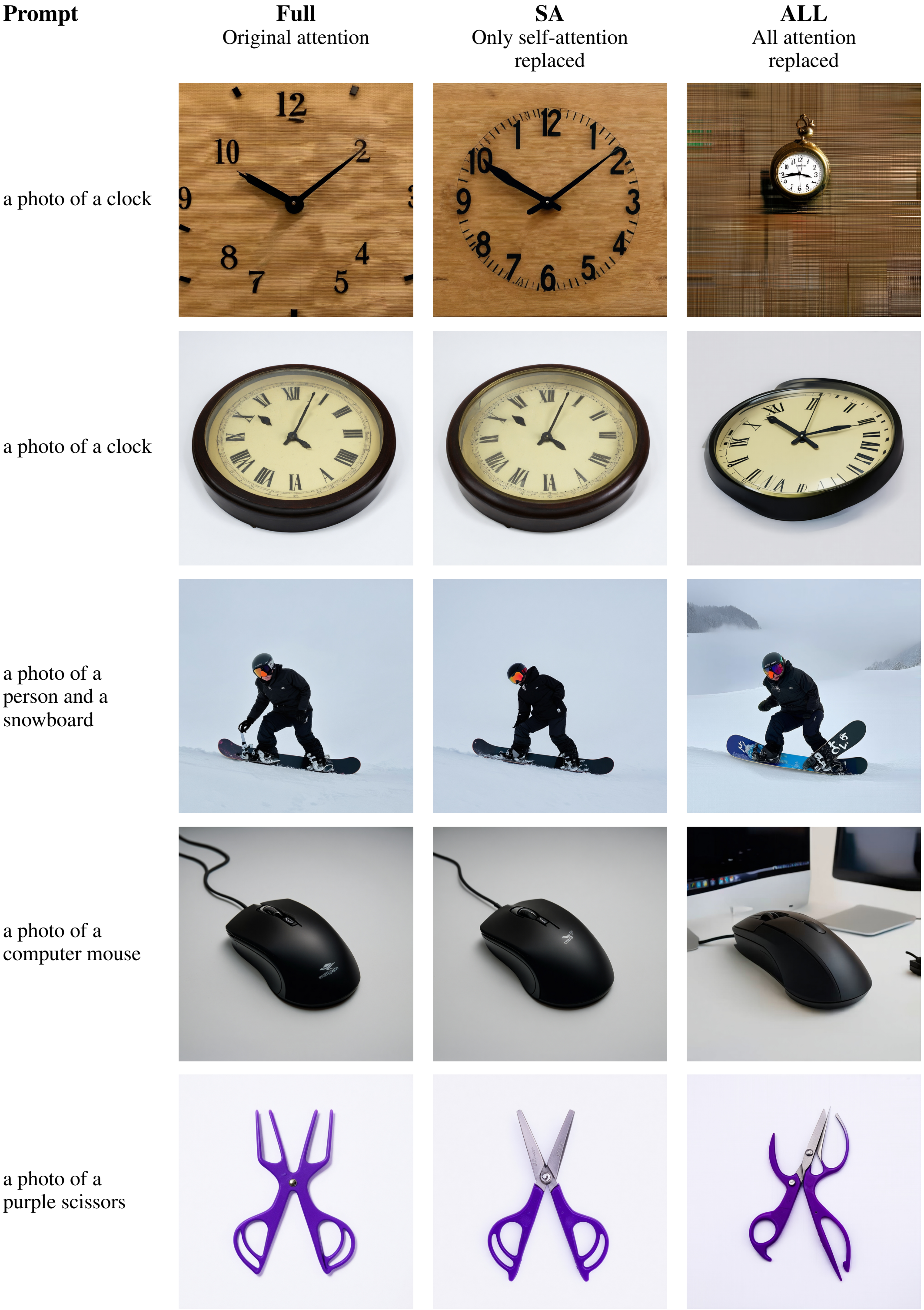

    \centering
    \sdBeforeHeader
    \sdBeforeRow{pdf_assets/geneval/00003_00002}{jpg}{a photo of a clock}
    \sdBeforeRow{pdf_assets/geneval/00003_00000}{jpg}{a photo of a clock}
    \sdBeforeRow{pdf_assets/geneval/00135_00000}{jpg}{a photo of a person and a snowboard}
    \sdBeforeRow{pdf_assets/geneval/00030_00001}{jpg}{a photo of a computer mouse}
    \sdBeforeRow{pdf_assets/geneval/00315_00000}{jpg}{a photo of a purple scissors}
    \caption{Additional samples from Stable Diffusion 3.5 Medium before fine-tuning. Columns show Full, SA, and ALL, with matching sampling settings within each row. ALL replaces both self-attention and joint-attention operators. The first two rows are distinct samples from the same clock prompt. SA depicts the clock numerals more accurately and renders the snowboarder's goggle lenses with more realistic colors (third row). It also renders the computer mouse and scissors more realistically than Full (last two rows).}
    \label{fig:sd35-beforeft-geneval}
\end{figure}

\FloatBarrier
\endgroup

\begingroup
\makeatletter
\setlength{\@fptop}{0pt}
\makeatother
\setlength{\parskip}{0pt}

\newcommand{\sdAfterPromptWidth}{0.17\linewidth}
\newcommand{\sdAfterImageWidth}{0.255\linewidth}
\newcommand{\sdAfterPromptFont}{\small}
\newcommand{\sdAfterHeader}{%
    \parbox[t]{\sdAfterPromptWidth}{\textbf{Prompt}}\hfill
    \parbox[t]{\sdAfterImageWidth}{\centering\textbf{Full+ft}\\\small Original attention}\hfill
    \parbox[t]{\sdAfterImageWidth}{\centering\textbf{SA+ft}\\\small Only self-attention\\replaced}\hfill
    \parbox[t]{\sdAfterImageWidth}{\centering\textbf{ALL+ft}\\\small All attention\\replaced}\par
    \vspace{4pt}
}
\newcommand{\sdAfterRow}[3]{%
    \begin{minipage}[c]{\sdAfterPromptWidth}
        \raggedright
        \sdAfterPromptFont #3
    \end{minipage}\hfill
    \begin{minipage}[c]{\sdAfterImageWidth}
        \includegraphics[width=\linewidth]{figures/moreexample/sd35medium/afterft/#1_full.#2}
    \end{minipage}\hfill
    \begin{minipage}[c]{\sdAfterImageWidth}
        \includegraphics[width=\linewidth]{figures/moreexample/sd35medium/afterft/#1_sa.#2}
    \end{minipage}\hfill
    \begin{minipage}[c]{\sdAfterImageWidth}
        \includegraphics[width=\linewidth]{figures/moreexample/sd35medium/afterft/#1_all.#2}
    \end{minipage}\par
    \vspace{5pt}
}

\begin{figure}[p]
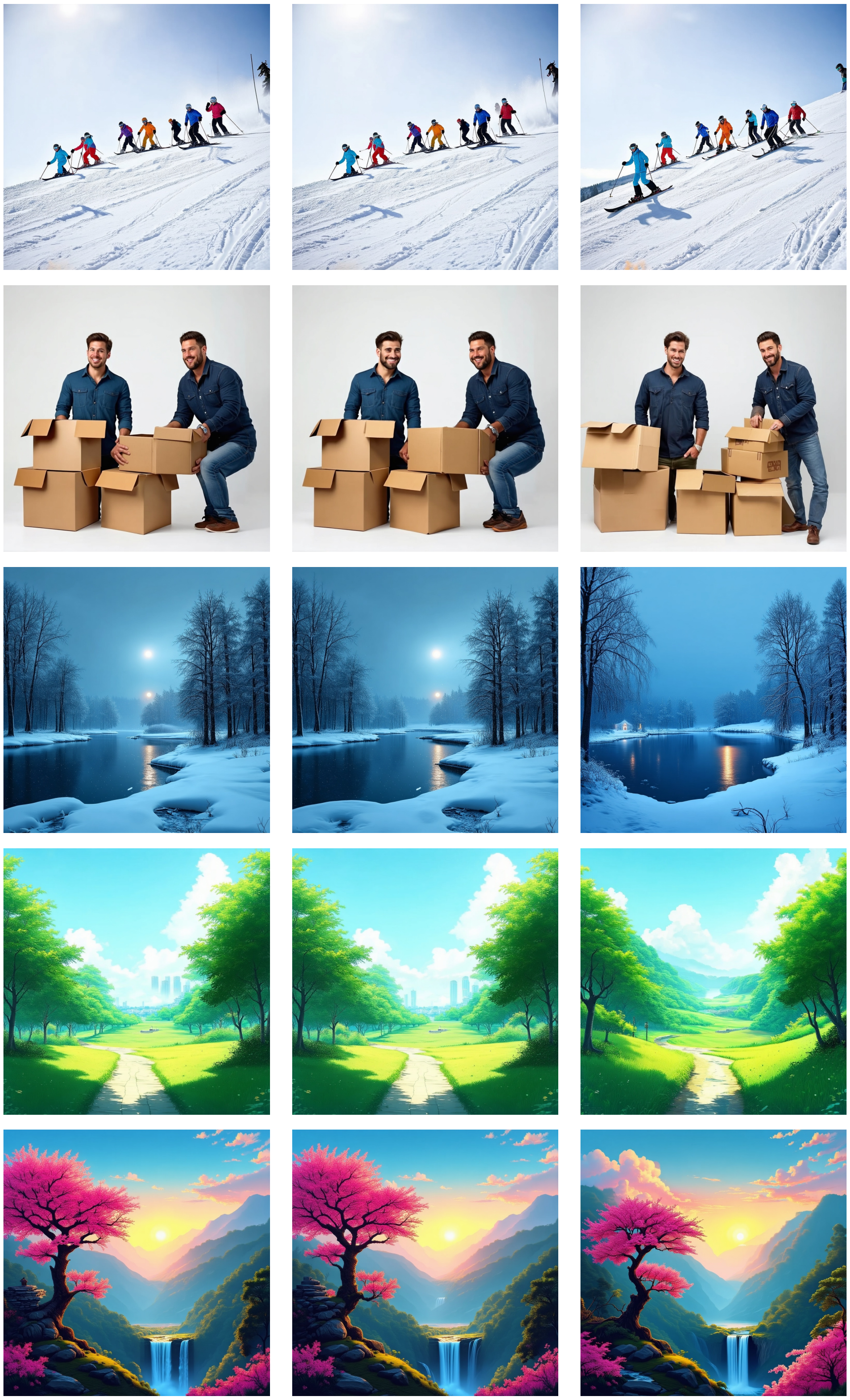

    \centering
    \sdAfterHeader
    \sdAfterRow{hps/photo_00799}{jpg}{a bunch of people on skiing on a hill}
    \sdAfterRow{hps/photo_00530}{jpg}{a couple of men that are next to some boxes}
    \sdAfterRow{hps/paintings_00164}{jpg}{Swedish lake at night with heavy snowfall depicted in hyper-realistic and detailed art.}
    \sdAfterRow{hps/paintings_00547}{jpg}{A green landscape by Makoto Shinkai.}
    \sdAfterRow{hps/concept_art_00362}{jpg}{A stylized digital art image of a cherry tree overlooking a valley with a waterfall during sunset.}
    \caption{Samples from Stable Diffusion 3.5 Medium after one epoch of fine-tuning. Each row compares Full+ft, SA+ft, and ALL+ft using the same prompt. ALL+ft replaces both self-attention and joint-attention operators. ALL+ft avoids the unnatural overlap between skiers seen in Full+ft (first row) and renders the men handling boxes with more plausible anatomy (second row).}
    \label{fig:sd35-afterft-hps}
\end{figure}

\begin{figure}[p]
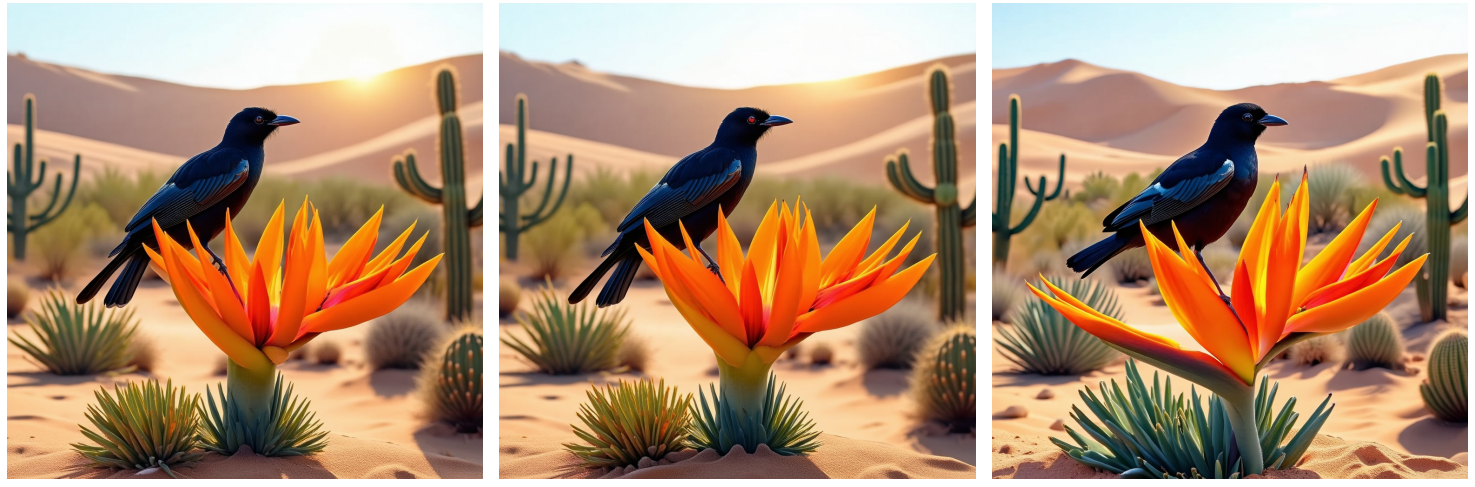

    \centering
    \renewcommand{\sdAfterPromptWidth}{0.30\linewidth}
    \renewcommand{\sdAfterImageWidth}{0.225\linewidth}
    \renewcommand{\sdAfterPromptFont}{\fontsize{8.5}{9.5}\selectfont}
    \sdAfterHeader
    \sdAfterRow{pdf_assets/dpg/drawtext37_3}{jpg}{a simple white background highlighting a hand-drawn black circle that encompasses the playful, cursive text `infinity makes me happy'. The words are crafted to resemble a quick, personal brush script, giving it a casual and intimate feel. Just outside the circle, faint sketch marks are visible, implying a human touch in the creation of this design.}
    \sdAfterRow{pdf_assets/dpg/midjourney13_3}{jpg}{A vivid and eclectic depiction of a fisherman's village drawn by Ralph Steadman, composed of dark velvet hues, contrasting with splashes of bright yellow and teal. In the foreground, a tall coconut tree sways, its silhouette framed against the backdrop of a small, bustling island. A sturdy jetty extends into the water, where several fishing boats rock gently with the waves.}
    \sdAfterRow{pdf_assets/dpg/partiprompts42_3}{jpg}{A chilled, transparent bottle of light beer with condensation beads forming on its surface, placed on a wooden table. A bright yellow slice of lemon is wedged onto the rim of the bottle, adding a citrus accent to the beverage. The bottle label is partially obscured by the moisture, but hints of green and gold are visible on the design.}
    \sdAfterRow{pdf_assets/dpg/stanford7_3}{jpg}{A man in a casual gray shirt and faded blue jeans is captured mid-air above a sleek black skateboard, executing a skillful jump. To the side of him lies a large black skateboard ramp with visible scuff marks from frequent use. In the expansive blue sky overhead, a few wispy clouds drift lazily by, adding a sense of height to his aerial feat.}
    \sdAfterRow{pdf_assets/dpg/whoops7_3}{jpg}{A striking black bird with glossy feathers sits atop the vibrant orange petals of a Bird of Paradise flower. The unique flower is positioned in the midst of an arid desert landscape, with various cacti and sparse vegetation dotting the sandy ground. In the background, the sun casts a warm glow on the distant rolling dunes.}
    \caption{Additional samples from Stable Diffusion 3.5 Medium after one epoch of fine-tuning. Each row compares Full+ft, SA+ft, and ALL+ft using the same prompt. ALL+ft replaces both self-attention and joint-attention operators.}
    \label{fig:sd35-afterft-dpg}
\end{figure}

\FloatBarrier
\endgroup

\section{Extended Related Work}
\label{app:extended-related-work}

The attention comparisons in Section~\ref{sec:related_work} draw on broader work in computational efficiency, spatial hierarchies, and image statistics.

\subsection{Linear and hardware-efficient attention}
\label{app:linear-hardware-attention}

Attention efficiency depends on both the representation of interactions and their implementation. FlashAttention \citep{dao2022flashattention} reduces memory traffic through tiled exact attention, and FlashAttention-2 \citep{dao2023flashattention2} improves work partitioning and parallelism. However, their dense arithmetic remains quadratic in token count. Linear attention addresses this arithmetic cost through kernel feature maps and associative multiplication \citep{katharopoulos2020linear}, with Performer approximating the softmax kernel through positive random features \citep{choromanski2021performer}. Kernel approximations and memory-efficient implementations therefore address distinct components of attention cost. SFA combines an $O(N\log N)$ attention architecture that delivers strong generative performance with a computationally efficient Triton implementation.

In diffusion models, Sana demonstrates the use of linear attention in a high-resolution diffusion architecture \citep{xie2025sana}. Native Sparse Attention (NSA) combines compressed representations, selected blocks, and a local window in language models \citep{yuan2025nsa}. These approaches allocate computation through a kernel representation or several attention branches; SFA instead assigns spatial resolution through a complete, query-specific partition of the token grid.

\subsection{Spatial hierarchies}

Spatial hierarchies provide a structured allocation of computation by representing interactions at different resolutions. For example, Swin Transformer \citep{liu2021swin} uses shifted local windows, and H-Transformer-1D \citep{zhu2021htransformer} constructs a one-dimensional hierarchical matrix approximation. For long-range interactions, Focal attention \citep{yang2021focal} combines fine local tokens and coarse distant learnable summaries in one softmax, retaining overlapping representations of regions across scales. QuadTree attention \citep{tang2022quadtree} instead makes refinement content dependent, building pooled query and key pyramids and selecting relevant coarse regions before evaluating finer interactions. This content-dependent refinement contrasts with SFA's fixed distance-dependent allocation, and the methods also differ in how they aggregate contributions across levels.

The methods mentioned above and in Section~\ref{sec:related_work} all differ from SFA. SFA introduces no additional parameters and supports both effective reuse of pretrained weights and high-quality generation when trained from scratch. Our comparison with LLSA shows that SFA's fixed spatial hierarchy achieves higher kernel throughput and better generation quality than hierarchical dynamic selection. The four ablations in Section~\ref{sec:ablation} and Appendix~\ref{app:ablation} further demonstrate the necessity of various components of our approach. Variants 1 and 2 remove count correction and normalize scales separately, respectively; both reduce quality in FFHQ-$128$ training and in direct self-attention replacement for Stable Diffusion 3.5 Medium without fine-tuning. Variant 3 retains only the exact local neighborhood, causing a substantial loss of FFHQ generation quality and demonstrating the contribution of pooled far-field context. Variant 4 replaces mean pooling with learnable aggregation and removes count correction, yielding worse FID and IS than SFA under the same ten-epoch PixelFlow training budget. Together, these comparisons support the combination of global coverage, mean pooling, count correction, and shared normalization.

These choices also support the kernel efficiencies detailed in Appendix~\ref{app:implementation}. Shared softmax reuses normalization and gradient statistics across scales, while mean pooling enables recursive preprocessing and uniform gradient propagation within each base block. Variants 2 and 4 lose these respective forms of computational reuse, with their efficiency consequences analyzed in Appendix~\ref{app:ablation}. In the FFHQ measurements, SFA reduces attention's share of runtime sufficiently for other transformer components to dominate, yielding nearly constant token throughput at a fixed token batch across resolutions. Further end-to-end acceleration therefore depends on optimizing those components.

\subsection{Natural-image statistics and positional frequencies}

The spatial prior underlying SFA has roots in classical studies that connect natural-image statistics to efficient spatial representations \citep{field1987relations,simoncelli2001natural} and document approximate scale invariance \citep{ruderman1994statistics}. In two dimensions, an inverse-square spectrum distributes comparable energy across spatial octaves, motivating attention resolution relative to distance. A related connection between frequency and image structure appears in diffusion models, where \citet{kadkhodaie2024generalization} relate generalization to denoising in harmonic bases adapted to image geometry. Within attention, positional encoding adds another source of frequency dependence by determining how spatial phase enters the computation. Rotary position embeddings (RoPE) express relative positions through rotations \citep{su2024roformer} and have effective two-dimensional constructions for vision transformers \citep{heo2024ropevit}. Studies of trained language models further reveal distinct uses of high and low rotary frequencies \citep{barbero2025rope}. These observations are relevant to SFA because it averages keys after positional encoding, so its coarse blocks aggregate both content and positional phase. The resulting connection between spatial averaging and frequency response is analyzed in Appendix~\ref{app:rope_pooling}.

\typeout{get arXiv to do 4 passes: Label(s) may have changed. Rerun}
\end{document}